\newlength\bibsep 
\documentclass[preprint,5p,nonatbib]{elsarticle}

\makeatletter 
\let\c@author\relax

\makeatother

\usepackage[%
  backend=biber,
  url=true,            
  doi=true,             
  style=authoryear,     
  citestyle=authoryear-comp, 
  maxnames=2,           
  minnames=1,           
  maxbibnames=99,       
  giveninits=true,     
  uniquename=mininit,   
  uniquelist=false,     
  dashed=false,          
  sorting=ynt,          
]{biblatex}

\AtBeginBibliography{\small}

\usepackage{graphicx}
\usepackage{multicol,multirow}
\usepackage{amsmath,amssymb,amsfonts}
\usepackage[T1]{fontenc}
\usepackage{xcolor}
\usepackage{doi}
\usepackage{hyperref}
\hypersetup{
    breaklinks = true,
    colorlinks = true,
}

\usepackage{bm}

\def\vc{{\bm{c}}}
\def\vd{{\bm{d}}}

\def\vx{{\bm{x}}}

\usepackage{booktabs}       
\usepackage{nicefrac}       
\usepackage{microtype}      
\usepackage[inline]{enumitem} 
\usepackage[super]{nth} 
\usepackage{adjustbox} 
\usepackage{wrapfig} 

\usepackage[capitalise,nameinlink]{cleveref} 

\usepackage{pifont} 
\newcommand{\goodI}{\textcolor{green!30!black}{\ding{51}}}
\newcommand{\goodII}{\textcolor{green!55!black}{\ding{51}\ding{51}}}
\newcommand{\goodIII}{\textcolor{green!80!black}{\ding{51}\ding{51}\ding{51}}}
\newcommand{\badI}{\textcolor{red!30!black}{\ding{55}}}
\newcommand{\badII}{\textcolor{red!55!black}{\ding{55}\ding{55}}}
\newcommand{\badIII}{\textcolor{black}{N/A}}

\newcommand{\dInc}{\texttt{Inc}}
\newcommand{\dIncLow}{\texttt{Inc\textsubscript{low}}}
\newcommand{\dIncHigh}{\texttt{Inc\textsubscript{high}}}
\newcommand{\dIncVar}{\texttt{Inc\textsubscript{var}}}
\newcommand{\dTra}{\texttt{Tra}}
\newcommand{\dTraExt}{\texttt{Tra\textsubscript{ext}}}
\newcommand{\dTraInt}{\texttt{Tra\textsubscript{int}}}
\newcommand{\dTraLong}{\texttt{Tra\textsubscript{long}}}
\newcommand{\dIso}{\texttt{Iso}}

\newcommand{\mTf}[1]{\textit{TF\textsubscript{{#1}}}}
\newcommand{\mUnet}[1]{\textit{U-Net\textsubscript{{#1}}}}
\newcommand{\mACDM}[1]{\textit{ACDM\textsubscript{{#1}}}}
\newcommand{\mResNet}[1]{\textit{ResNet\textsubscript{{#1}}}}
\newcommand{\mFNO}[1]{\textit{FNO\textsubscript{{#1}}}}
\newcommand{\mRefiner}[1]{\textit{Refiner\textsubscript{{#1}}}}
\newcommand{\mDFP}[1]{\textit{DFP\textsubscript{{#1}}}}

\newcommand{\best}[1]{\textcolor{blue!50!black}{\underline{{#1}}}}
\newcommand{\bestSec}[1]{\textcolor{blue!50!black}{{#1}}}
\newcommand{\diverge}[1]{\textcolor{black!60!}{{#1}}}
\newcommand{\miss}{---}

\makeatletter
\def\ps@pprintTitle{%
    \let\@oddhead\@empty
    \let\@evenhead\@empty
    \def\@oddfoot{\hfill \thepage \hfill}%
    \let\@evenfoot\@oddfoot
    }
\makeatother

\begin{document}


\begin{frontmatter}

\title{Benchmarking Autoregressive Conditional Diffusion Models for \\ Turbulent Flow Simulation}


\author[tum]{Georg Kohl}
\emailauthor{georg.kohl@tum.de}{Georg Kohl}
\author[tum]{Li-Wei Chen}
\emailauthor{jilinchl@163.com}{Li-Wei Chen}
\author[tum]{Nils Thuerey}
\emailauthor{nils.thuerey@tum.de}{Nils Thuerey}

\affiliation[tum]{organization={Technical University of Munich},
             addressline={Boltzmannstraße 3},
             city={85748 Garching},
             country={Germany}}

\begin{abstract}
    Simulating turbulent flows is crucial for a wide range of applications, and machine learning-based solvers are gaining increasing relevance. However, achieving temporal stability when generalizing to longer rollout horizons remains a persistent challenge for learned PDE solvers. In this work, we analyze if fully data-driven fluid solvers that utilize an autoregressive rollout based on conditional diffusion models are a viable option to address this challenge. We investigate accuracy, posterior sampling, spectral behavior, and temporal stability, while requiring that methods generalize to flow parameters beyond the training regime. To quantitatively and qualitatively benchmark the performance of various flow prediction approaches, three challenging 2D scenarios including incompressible and transonic flows, as well as isotropic turbulence are employed. We find that even simple diffusion-based approaches can outperform multiple established flow prediction methods in terms of accuracy and temporal stability, while being on par with state-of-the-art stabilization techniques like unrolling at training time. Such traditional architectures are superior in terms of inference speed, however, the probabilistic nature of diffusion approaches allows for inferring multiple predictions that align with the statistics of the underlying physics. Overall, our benchmark contains three carefully chosen data sets that are suitable for probabilistic evaluation alongside various established flow prediction architectures.
\end{abstract}



\begin{keyword}

flow prediction \sep turbulent flow \sep PDEs \sep diffusion models \sep numerical simulation

\end{keyword}

\end{frontmatter}

\section{Introduction}
Simulations of partial differential equations (PDEs), particularly those involving turbulent fluid flows, constitute a crucial research area with applications ranging from medicine \parencite{olufsen2000_Numerical} to climate research \parencite{wyngaard1992_Atmospheric}, as well as numerous engineering fields \parencite{moin1998_Direct,verma2018_Efficient}. Historically, such flows have been simulated via iterative numerical solvers for the Navier-Stokes equations. Recently, there has been a growing interest in combining or replacing traditional solvers with deep learning. These approaches have shown considerable promise in terms of enhancing the accuracy and efficiency of fluid simulations \parencite{geneva2022_Transformers,han2021_Predicting,stachenfeld2022_Learned}.
However, despite the significant progress made in this field, a major remaining challenge is the ability to predict rollouts that maintain both stability and accuracy over longer temporal horizons \parencite{kochkov2021_Machine,um2020_Solverintheloop}. Fluid simulations are inherently complex and dynamic, and therefore, it is highly challenging to accurately capture the intricate physical phenomena that occur over extended periods of time. Additionally, due to their chaotic nature, even small ambiguities of the spatially averaged states used for simulations can lead to fundamentally different solutions over time \parencite{pope2000_Turbulent}. However, most learned methods and traditional numerical solvers process simulation trajectories deterministically, and thus only provide a single answer. 

We explore these issues by investigating the usefulness of the recently emerging con\-di\-tio\-nal diffusion models \parencite{ho2020_Denoising,song2021_Scorebased} for turbulent flows, which serve as representatives for more general PDE-based simulations. Specifically, we are interested in the probabilistic prediction of fluid flow trajectories from an initial condition. We aim for answering the question: \textit{Are autoregressive diffusion models competitive tools for modeling fluid simulations compared to other established architectures?} For this purpose, we analyze model accuracy, temporal rollout stability, spectral behavior, as well as computational costs for different approaches. Furthermore, our focus on fluid flows makes it possible to analyze the generated posterior samples with the statistical temporal metrics established by turbulence research \parencite{dryden1943_Review}. Unlike application areas like imaging or speech, where the exact distribution of possible solutions is typically unknown, these turbulence metrics make it possible to reliably evaluate the quality of different samples generated by a probabilistic model. To summarize, the central aims and outcomes of this work are as follows:

\begin{enumerate}[label=(\textbf{\textit{\roman*}})]
\item We combine a conditional diffusion approach with an autoregressive rollout to produce a probabilistic surrogate flow simulator, that represents state of the art diffusion models applied to flow prediction.
\item We broadly compare this approach on flow prediction problems with increasing difficulty against a range of common architectures in terms of accuracy, posterior sampling, temporal stability, and statistical correspondence to the underlying physical behavior.
\item We show that even simple, diffusion-based flow predictors are remarkably competitive with state-of-the-art stabilization techniques like unrolling at training time, while providing diverse, physically plausible posterior samples.
\item We provide a benchmark featuring various established flow prediction architectures and three carefully curated data sets. Our source code and data sets are available at \url{https://github.com/tum-pbs/autoreg-pde-diffusion}.
\end{enumerate}

\section{Related Work}
In the following, we discuss related literature on techniques for incorporating machine learning into fluid solvers, on diffusion models, and works that utilize diffusion models for fluids and temporal predictions.

\subsection{Fluid Solvers utilizing Machine Learning}
A variety of works have used machine learning as means to improve numerical solvers. Several approaches focus on learning computational stencils \parencite{bar-sinai2019_Learning,kochkov2021_Machine} or additive corrections \parencite{deavilabelbute-peres2020_Combining,list2022_Learned,um2020_Solverintheloop,melchers2023_Comparison} to increase simulation accuracy. In addition, differentiable solvers have been applied to solve inverse problems such as fluid control \parencite{holl2020_Learning}. An overview is provided, e.g., by \textcite{thuerey2022_Physicsbased}. When the solver is not integrated into the computational graph, typically a data-driven surrogate model is trained to replace the solver.
Convolutional neural networks (CNNs) for such flow prediction problems are very popular, and often employ an encoder-processor-decoder architecture. For the latent space processor, multilayer perceptrons \parencite{kim2019_Deep,wu2022_Learning}, long short-term memory networks \parencite{wiewel2019_Latent} as well as echo state networks \parencite{racca2023_Predicting, gupta2023_Modelfree} were proposed. 

As particularly successful latent architectures, transformers \parencite{vaswani2017_Attention} have also been combined with CNN-based encoders as a reduced-order model \parencite{hemmasian2023_Reducedorder}, for example to simulate incompressible flows via Koopman-based latent dynamics \parencite{geneva2022_Transformers}. Alternatives do not rely on an autoregressive latent model, e.g., by using spatio-temporal 3D convolutions \parencite{deo2023_Combined}, dilated convolutions \parencite{stachenfeld2022_Learned}, Bayesian neural networks for uncertainty quantification \parencite{geneva2019_Quantifying}, or problem-specific multi-scale architectures \parencite{wang2020_Physicsinformed}.
Furthermore, various works utilize message passing architectures \parencite{brandstetter2022_Message,pfaff2021_Learning}, and adding noise to training inputs was likewise proposed to improve temporal prediction stability for graph networks \parencite{sanchez-gonzalez2020_Learning}. Transformer-based latent models have also been combined with a graph network encoder and decoder in previous work \parencite{han2021_Predicting}, and attention was similarly used for particle-based simulations \parencite{chen2024_DualFluidNet}. Compared to prior PDE benchmarks \parencite{gupta2023_Multispatiotemporalscale,takamoto2022_PDEBench}, our work is especially suitable for evaluating probabilistic posterior samples, as it contains chaotic and increasingly underdetermined tasks where different trajectories are physically plausible solutions.

\subsection{Diffusion Models}
Diffusion models \parencite{hyvarinen2005_Estimation, sohl-dickstein2015_Deep} became popular after diffusion probabilistic models and denoising score matching were combined for high-quality unconditional image generation \parencite{ho2020_Denoising}. This approach has since been improved in many aspects, e.g., with meaningful latent representations \parencite{song2021_Denoising} or better sampling \parencite{nichol2021_Improved}. In addition, generative hybrid approaches were proposed, for instance diffusion autoencoders \parencite{preechakul2022_Diffusion} or score-based latent models \parencite{vahdat2021_Scorebased}. Diffusion models for image generation are typically conditioned on simple class labels \parencite{dhariwal2021_Diffusion} or textual inputs \parencite{saharia2022_Photorealistic}. Pre-trained diffusion models were employed for inverse image problems \parencite{kawar2022_Denoising}, and different conditioning approaches were compared for score-based models on similar tasks \parencite{batzolis2021_Conditional}. Furthermore, an unconditional diffusion model has been combined with inverse problem solving for medical applications \parencite{song2022_Solving}. For an in-depth review of diffusion approaches we refer to \parencite{yang2024_Diffusion}.

\begin{figure*}[th]
    \centering
    \includegraphics[width=0.95\textwidth]{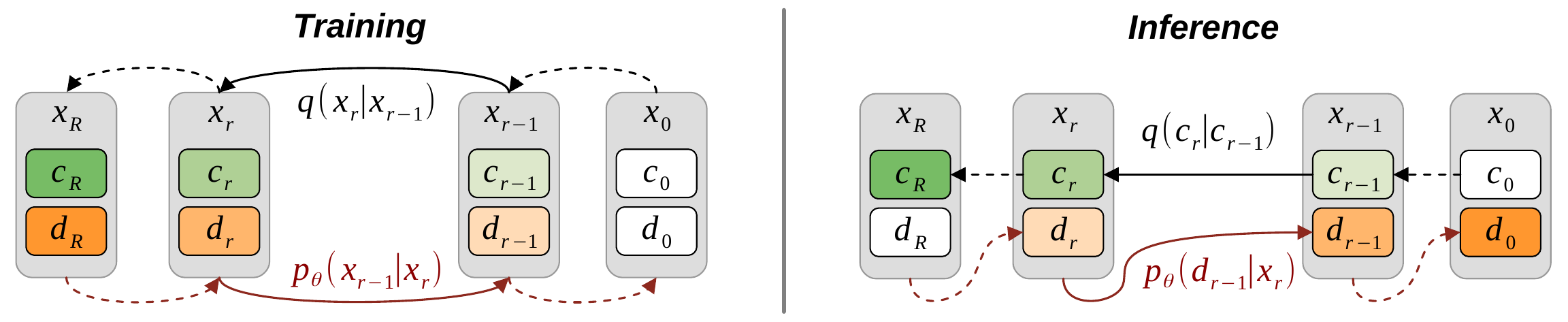}
    \caption{Diffusion conditioning approach with the forward (black) and reverse process (red) during training and inference. White backgrounds for \( \vc \) or \( \vd \) indicate given information, i.e., inputs for each phase. In the context of the autoregressive surrogate simulator, \( \vc_0 \) contains information about the simulated process like Reynolds or Mach number, as well as the initial or previous simulation state. \( \vd_0 \) contains the next target simulation state during training, and is the resulting prediction of the next state during inference.}
    \label{fig: diffusion conditioning}
\end{figure*}

\subsection{Diffusion Models for Fluids and Temporal Prediction}
Selected works have applied diffusion models to temporal prediction tasks like unconditional or text-based video generation, as well as video prediction \parencite[e.g.][]{harvey2022_Flexible,ho2022_Video,hoppe2022_Diffusion}. These methods typically directly include time as a third dimension or re-use the batch dimension \parencite{blattmann2023_Align}. As a result, autoregressive rollouts are only used to create longer output sequences compared to the training domain, with the drawback that predictions quickly accumulate errors or lose temporal coherence. Very few works exist that apply diffusion methods to transient physical processes. To solve inverse physics problems score matching was utilized \parencite{holzschuh2023_Solving}, and physics-informed diffusion models for a frame-by-frame super-resolution task for physics simulations exist \parencite{shu2023_Physicsinformed}. Early steps towards turbulent flows in 3D were taken, via a purely generative diffusion setup based on boundary geometry information \parencite{lienen2024_Zero}. Instead of an autoregressive approach, using physical time as a conditioning for diffusion-based fluid field prediction has been investigated, but the authors report stability issues and unphysical predictions as a result \parencite{yang2023_Denoising}. A multi-step refinement process similar to diffusion models was proposed to improve PDE predictions \parencite{lippe2023_PDERefiner}, which is also analyzed below. While it is possible to achieve stability improvements with this approach, it provided little variance in posterior samples and was highly sensitive to hyperparameters in our experiments. Predictor-interpolator schemes inspired by diffusion models were also introduced \parencite{cachay2023_DYffusion}. This method combines a predictor, that equates the diffusion time step with the physical time step of the simulation, with a probabilistic, Bayesian interpolator model. As a result, compared to the direct application of diffusion models analyzed here, this allows for larger time steps and potential performance improvements, while having drawbacks in terms of posterior coverage and temporal coherence.

\section{Background on Conditional Diffusion Models} \label{sec: diffusion background}
A denoising diffusion probabilistic model (DDPM) is a generative model based on a parameterized Markov chain, and contains a fixed forward and a learned reverse process \parencite{ho2020_Denoising} over \( R \) steps. For any \( r \in {1,2,\ldots,R} \), the forward process
\begin{equation}
    \label{eq: forward process}
    q(\vx_{r} | \vx_{r-1}) =
    \mathcal{N}(\vx_r; \sqrt{1-\beta_r} \vx_{r-1}, \beta_r \mathbf{I})
\end{equation}
incrementally adds Gaussian noise to the original data \( \vx_0 \) according to a variance schedule \( \beta_1, \ldots, \beta_R \) resulting in the latent variable \( \vx_R \), that corresponds to pure Gaussian noise. The reverse process 
\begin{equation}
    \label{eq: reverse process}
    p_\theta(\vx_{r-1} | \vx_{r}) =
    \mathcal{N}(\vx_{r-1}; \boldsymbol{\mu}_\theta(\vx_r,r), \boldsymbol{\Sigma}_\theta(\vx_r,r))
\end{equation}
contains learned transitions, i.e., \( \boldsymbol{\mu}_\theta \) and \( \boldsymbol{\Sigma}_\theta \) are computed by a neural network parameterized by \( \theta \) given \( \vx_r \) and \( r \). The network is trained via the variational lower bound (ELBO) using reparameterization. During inference the initial latent variable \( \vx_R \sim \mathcal{N}(\boldsymbol{0}, \mathbf{I}) \) as well as the inter\-mediate diffusion steps are sampled, leading to a probabilistic generation of \( \vx_0 \) with a distribution that is similar to the distribution of the training data. Note that the latent space of a DDPM by construction has the same dimensionality as the input space, in contrast to, e.g., variational autoencoders (VAEs) \parencite{kingma2014_Autoencoding}. Thereby, it avoids problems with the generation of high frequency details due to compressed representations. Compared to generative adversarial networks (GANs), diffusion models typically do not suffer from mode collapse or convergence issues \parencite{metz2017_Unrolled}.

To condition the DDPM on information like the initial state and characteristic dimensionless quantities for flow prediction, we employ a concatenation-based conditioning approach \parencite{batzolis2021_Conditional}: Each element \( \vx_0 = (\vd_0, \vc_0 )\) of the diffusion process now consists of a data component \( \vd_0 \) that is only available during training and a conditioning component \( \vc_0 \) that is always given. Correspondingly, the task at inference time is the conditional prediction \( P(\vd_0 | \vc_0) \), i.e., computing the prediction target \( \vd_0 \) given the conditioning \( \vc_0 \). During training, the basic DDPM algorithm remains unchanged as
\begin{equation}
    \label{eq: diffusion training}
    \vx_r = (\vc_r, \vd_r) \;\; \text{with} \;\;
    \vc_r \sim q( \:\cdot\: | \vc_{r-1}); \; \vd_r \sim q( \:\cdot\: | \vd_{r-1})
\end{equation}
is still produced by the incremental addition of noise during the forward process. During inference \( \vd_R \sim \mathcal{N}(\boldsymbol{0}, \mathbf{I}) \) is sampled and processed in the reverse process, while \( \vc_0 \) is known and any \( \vc_r \) thus can be obtained from \cref{eq: forward process}, i.e.,
\begin{equation}
    \label{eq: diffusion inference}
    \vx_r = (\vc_r, \vd_r) \;\; \text{where} \;\;
    \vc_r \sim q( \:\cdot\: | \vc_{r-1}); \; \vd_r \sim p_\theta( \:\cdot\: | \vx_{r+1}).
\end{equation}
Here, \( q(\vc_r | \vc_{r-1}) \) denotes the forward process for \( \vc \), and \( \vd_r \sim p_\theta( \:\cdot\: | \vx_{r+1}) \) is realized by discarding the prediction of \( \vc_{r} \) when evaluating \( p_\theta \). A visualization of this conditioning technique is shown in \cref{fig: diffusion conditioning}. We found the addition of noise to the conditioning, instead of simply using \( \vc_{0} \) over the entire diffusion process, to be crucial for the temporal stability of the simulation rollout as detailed in \cref{sec: results}.

\section{Flow Prediction Architectures}
Our problem setting is the following: a temporal trajectory \( s^1, s^2, \ldots, s^T \) of states should be predicted given an initial state \( s^0 \). Each \(s^t\) consists of dense simulation fields like velocity or pressure, combined with simulation parameters like the Reynolds or Mach number, as constant, spatially expanded channels (see right of \cref{fig: diffusion rollout}). Numerical solvers \( f \) iteratively predict \( s^t = f(s^{t-1}) \), and here we similarly investigate neural surrogate models \( f_\theta \) with parameters \(\theta\) to autoregressively predict \( s^t = f_\theta(s^{t-1}) \), or \( s^t \sim f_\theta(s^{t-1}) \) for probabilistic methods. In the following, it is important to distinguish this \emph{simulation rollout} via $f_\theta$ from the \textit{diffusion rollout} $p_\theta$. The former corresponds to physical time and consists of \emph{simulation-} or \emph{time steps} denoted by \( t \in {0,1,\ldots,T} \) superscripts. The latter refers to \emph{diffusion steps} inside the Markov chain, denoted by \( r \in {0,1,\ldots,R} \) subscripts as above. To ensure a fair comparison, all models were parameterized with a similar parameter count, and suitable key hyperparameters were determined with a broad search for each architecture. Implementation details for each of the following architectures can be found in \cref{app: implementation}.

\subsection{Autoregressive Conditional Diffusion Models}
To extend the conditional DDPM described in \cref{sec: diffusion background} to temporal tasks, we build on autoregressive single-step prediction, with \( k \) previous steps: \( s^t \sim f_\theta(\:\cdot\: | s^{t-k}, \ldots, s^{t-1}) \).

\paragraph{Training}
\(f_\theta\) is trained in the following manner: Given a data set with different physical simulation trajectories, a random simulation state \( s^t \in s^0, s^1, \ldots, s^T \) is selected from a sequence as the prediction target \( \vd_0 \). The corresponding conditioning consists of \( k \) previous simulation states \( \vc_0 = (s^{t-k}, \ldots, s^{t-1})\). Next, a random diffusion time step \( r \) is sampled, leading to \( \vx_r \) via the forward process. The network learns to predict the added noise level via the ELBO as in the original DDPM \parencite{ho2020_Denoising}.

\begin{figure}[ht]
    \centering
    \includegraphics[width=0.48\textwidth]{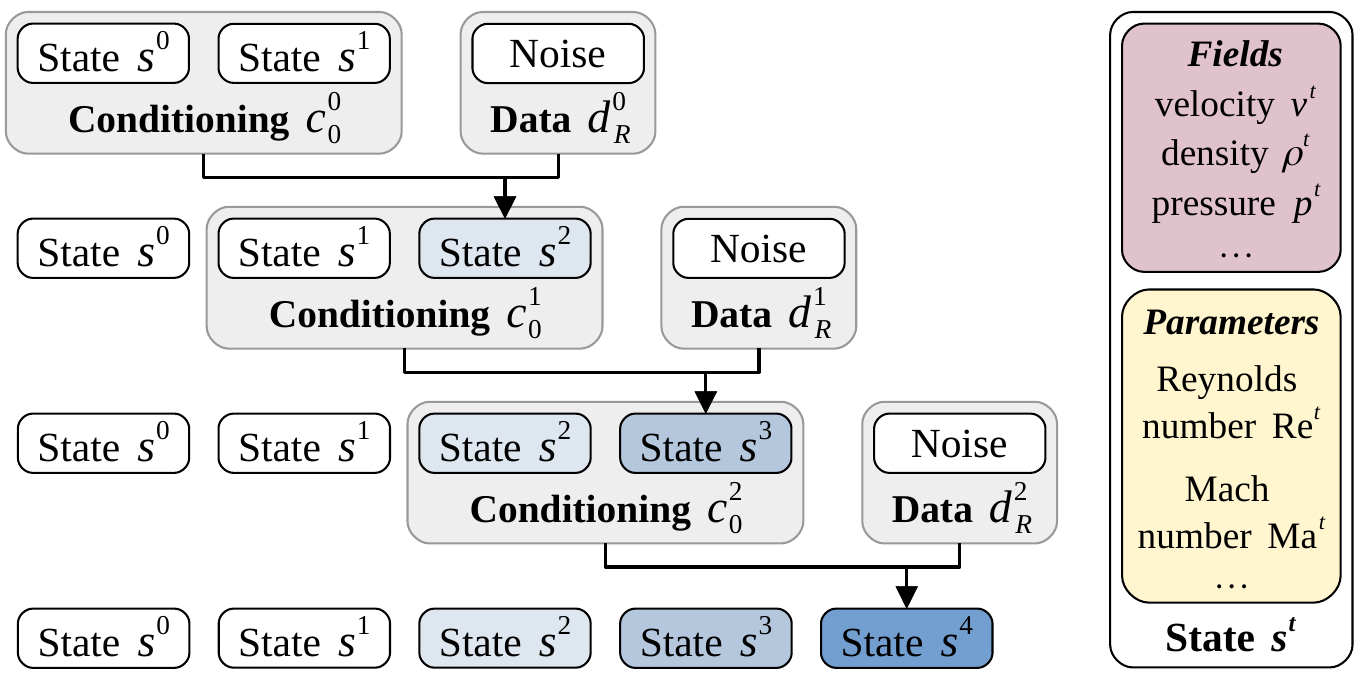}

    \caption{Autoregressive simulation rollout during inference with diffusion models for \( k=2 \) input steps (left), and contents of each simulation state (right).}
    \label{fig: diffusion rollout}
\end{figure}

\paragraph{Inference}
The training objective above allows for producing a single subsequent time step as the final output of the diffusion process \( \vd_0 \) during inference. We can then employ this single-step prediction for sampling simulation rollouts with arbitrary length by autoregressively reusing generated states as conditioning for the next iteration: For each simulation step, \cref{eq: diffusion inference} is unrolled from \( \vx_R^{t} \) to \( \vx_0^{t} \) starting from \( \vd_R^{t} \sim \mathcal{N}(\boldsymbol{0}, \mathbf{I}) \) and \( \vc_0^{t} = (s^{t-{k}}, \ldots, s^{t-1}) \). Then, the predicted next time step is \( s^t = \vd_0^t \). This process is visualized in \cref{fig: diffusion rollout}, and we denote models trained with this approach as \textit{autoregressive conditional diffusion models} (ACDMs) in the following.

The motivation for this combination of conditioning and simulation rollout is the hypothesis that perturbations to the conditioning can be compensated during the diffusion rollout, leading to improved temporal stability. Especially so, when smaller inference errors inevitably accumulate over the course of long simulation rollouts. Furthermore, this autoregressive rollout approach ensures that the network produces a temporally coherent trajectory for every step along the inferred sequence. This stands in contrast to explicitly conditioning the DDPM on physical time \( t \), i.e. treating it in the same way as the diffusion step \( r \) \parencite[as proposed by][]{yang2023_Denoising}. For a prob\-a\-bilis\-tic model it is especially crucial to have access to previously generated outputs during the prediction, as otherwise temporal coherence can only potentially be achieved via tricks such as fixed temporal and spatial noise patterns.

\paragraph{Implementation}
We employ a widely used U-Net \parencite[based on][]{ronneberger2015_UNet} with various established smaller architecture modernizations \parencite{dhariwal2021_Diffusion,ho2020_Denoising}, to learn the reverse process. A standard, linear DDPM variance schedule is used. We use \( k=2 \) previous steps for the model input, and achieved high-quality samples with \( 20 \)--\( 100 \) diffusion steps \(R\), depending on how strongly each setting is conditioned. This is in line with other recent works that achieve competitive results with as little as \( R \approx 30 \) in the image domain \parencite{chung2022_Comecloserdiffusefaster,karras2022_Elucidating}. Combining \(\vd_0\) and \( \vc_0 \) to form \( \vx_0 \), as well as aggregating multiple states for \(\vc_0\) is achieved via concatenation along the channel dimension. Parameters from the simulation are expanded to match the spatial dimensions of the fields and concatenated along the channel dimension as well. For \mACDM{} and the other architectures presented in the following, parameter predictions are overwritten with the known values from the simulation during the autoregressive inference rollout to prevent drift.

\subsection{U-Net Variants}
A crucial question is how much difference the diffusion training itself makes in comparison to more traditional training approaches of the same backbone architecture. Hence, we investigate a range of model variants with identical network architecture to the \mACDM{} model, that are optimized with different supervised setups.

\paragraph{Classic Next-step Predictor}
As a first, simple baseline, the backbone architecture is trained with an MSE loss to directly predict one future simulation state with a single model pass. It is denoted by \mUnet{} in the following.

\paragraph{Unrolled Training}
Several works have reported improvements from unrolling predictions during training instead of a single-step prediction \parencite{geneva2020_Modeling,lusch2018_Deep,um2020_Solverintheloop}. As such, we also analyse U-Net architectures with such an unrolled training (\mUnet{ut}) over \( m \) steps, where the gradient is fully backpropagated through the entire training trajectory. This additional complexity at training time results in substantial stability improvements, due to reducing the data shift from training to inference. For this approach, we found \( m=8 \) to be ideal across experiments.

\paragraph{Training Noise}
The usage of training noise was proposed to reduce problems from error accumulation during inference by explicitly learning to compensate errors in the training data \parencite{sanchez-gonzalez2020_Learning}. We investigate adding normally distributed noise to the U-Net input with varying standard deviation \( n \), and denote this model with training noise by \mUnet{tn}. We found values around \(n=10^{-2}\) to work well.

\paragraph{PDE-Refiner}
PDE-Refiner is a multi-step refinement process to improve the stability of learned PDE predictions \parencite{lippe2023_PDERefiner}. It relies on starting from the predictions of a trained one-step model, and iteratively refining them by adding noise of decreasing variance and denoising the result with the same model. The resulting model is then autoregressively unrolled to form a prediction trajectory. We re-implement this method, closely following the authors' pseudocode, only changing the backbone network to our \mUnet{} implementation for a fair comparison against the remaining architectures. PDE-Refiner, denoted by \mRefiner{} below, relies on two key hyperparameters, the number of refinement steps \(R\), and the minimum noise variance \(\sigma\). We found this approach to be highly sensitive to both parameters as detailed in \cref{app: ablation refiner}. Here, we report representative results using \(R=4\) and \(\sigma = 10^{-6}\) (\(\sigma = 10^{-5}\) for \dIso{}), in line with the authors' recommendations \parencite{lippe2023_PDERefiner}.

\subsection{ResNets and Fourier Neural Operators}
As additional popular approaches from the class of direct next-step predictor models, we investigate dilated ResNets \parencite[based on][]{stachenfeld2022_Learned} and Fourier Neural Operators (FNOs) \parencite{li2021_Fourier}. For the former, the proposed dilated ResNet (\mResNet{dil.}) as well as the same architecture without dilations (\mResNet{}) are included. For the latter, we investigate models using \((16,8)\) Fourier modes in x- and y-direction (\mFNO{16}), as well as \((32,16)\) modes (\mFNO{32}).

\subsection{Latent-space Transformers}
The success of transformer architectures \parencite{vaswani2017_Attention} and their recent application to physics predictions \parencite{geneva2022_Transformers,han2021_Predicting} raises the question how the other approaches fare in comparison to state-of-the-art transformer architectures. Being tailored to sequential processing with a long-term observation horizon, these models operate on a latent space with a reduced size. In contrast, the other investigated architectures by construction operate on the full spatial resolution. Specifically, we test the encoder-processor-decoder architecture from \textcite{han2021_Predicting} adopted to regular grids via a CNN-encoding, denoted by \mTf{MGN} below. Furthermore, we provide an improved variant (\mTf{Enc}) that allows to simulate flows with varying parameters over the simulation rollout, and varies key transformer parameters. Compared to \mTf{MGN} it relies on transformer encoder layers and uses full latent predictions instead of residual predictions. Lastly, we test the transformer-based prediction in conjunction with a probabilistic VAE, denoted by \mTf{VAE}. All transformer architectures have access to a large number of previous steps and use a rollout schedule in line with \parencite{han2021_Predicting} during training, however teacher forcing is removed. By default we use a \(30\) step input window and a training rollout length of \(60\).

\section{Experiments} \label{sec: experiments}
\begin{figure*}[ht]
    \centering
    \includegraphics[width=0.329\textwidth]{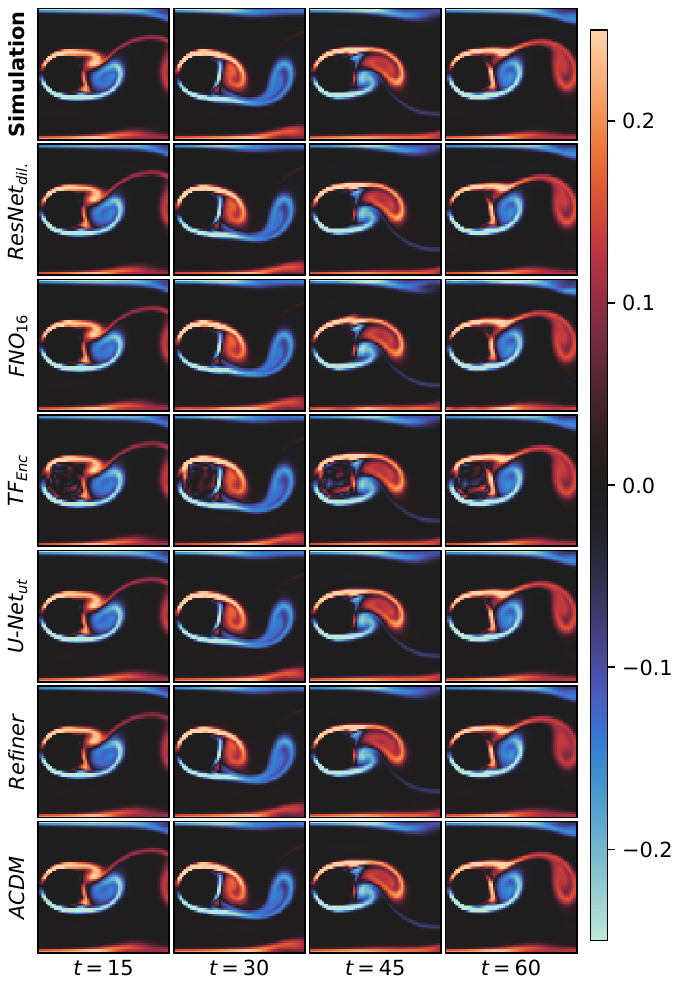}
    \hfill
    \includegraphics[width=0.329\textwidth]{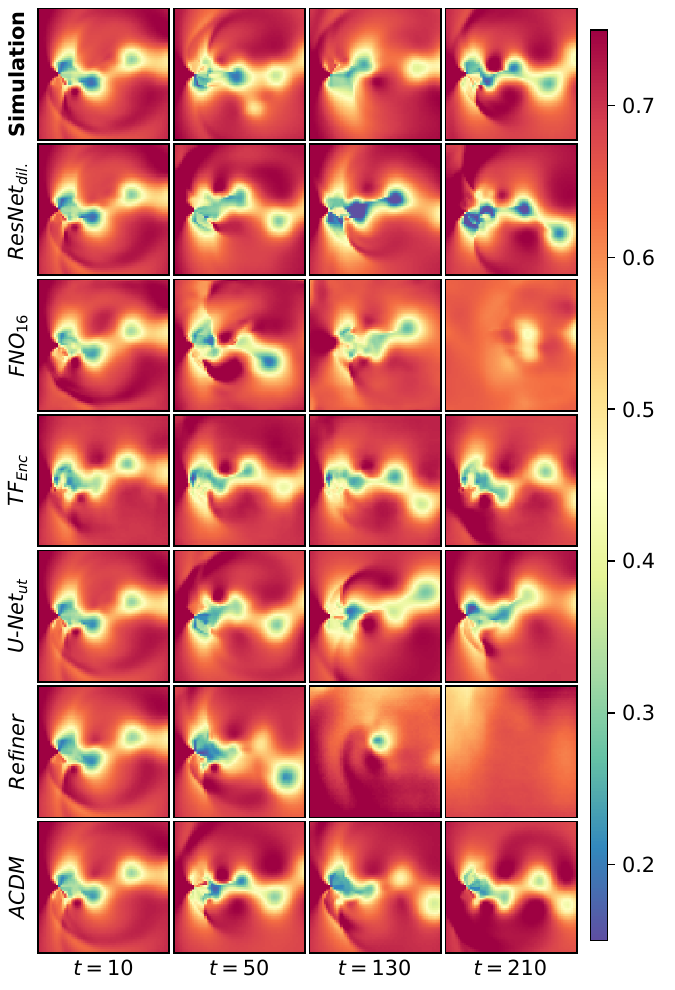}
    \hfill
    \includegraphics[width=0.329\textwidth]{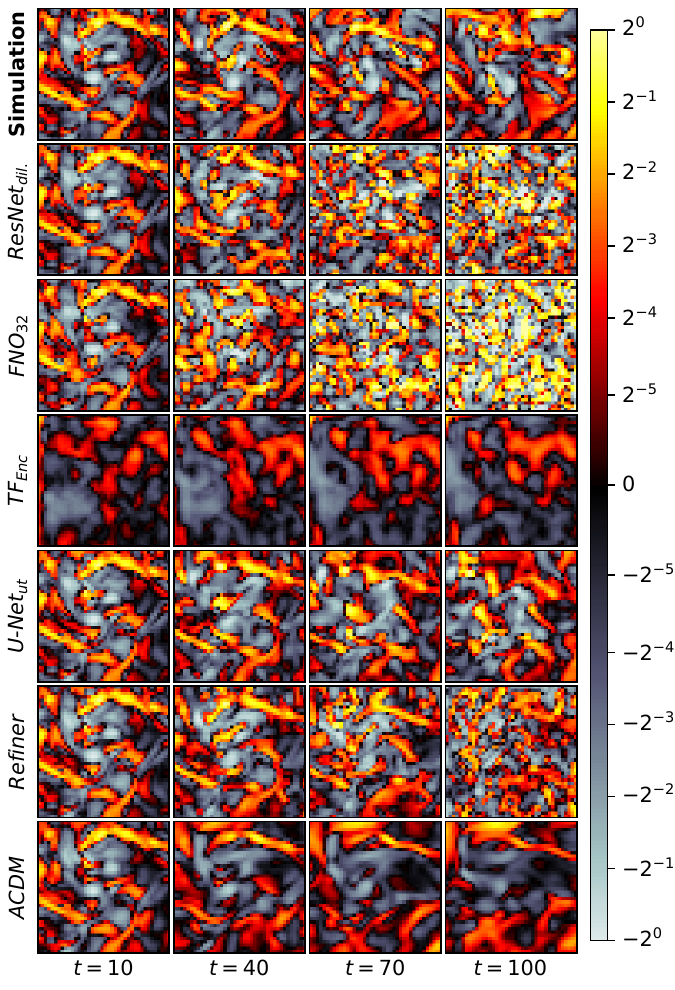}

    \caption{Zoomed example trajectories from \dIncHigh{} with \( \mathit{Re} = 1000 \) (left, vorticity), \dTraLong{} with \( \mathit{Ma} = 0.64 \) (middle, pressure), and \dIso{} with \( z = 280 \) (right, vorticity). Shown are trajectories from the numerical solver, and predictions by key architectures from each model class (also see \href{https://ge.in.tum.de/publications/2023-acdm-kohl/}{accompanying videos}).}
    \label{fig: prediction examples}
\end{figure*}

We quantitatively and qualitatively benchmark the investigated architectures on three flow prediction scenarios with increasing difficulty:
\begin{enumerate*}[label=(\textbf{\textit{\roman*}})]
\item an incompressible wake flow,
\item a transonic cylinder flow with shock waves, and
\item an isotropic turbulence flow.
\end{enumerate*}
Test cases for each scenario contain out-of-distribution data via simulation parameters outside of the training data range. Further experimental details are provided in \cref{app: data}. Examples of the solver trajectory as well as predictions from each model class are shown in \cref{fig: prediction examples}.

\subsection{Incompressible Wake Flow}
Our first case targets incompressible wake flows. These flows already encompass some complexity of the incompressible Navier-Stokes equations with boundary interactions, but represent the simplest of our three scenarios due to their direct unsteady periodic nature and laminar flow behavior. We simulate a fully developed incompressible Karman vortex street behind a cylindrical obstacle with PhiFlow \parencite{holl2020_Learning} for a varying Rey\-nolds number \( \mathit{Re} \leq 1000 \). The corresponding flows capture the transition from laminar to the onset of turbulence. Models are trained on data from simulation sequences with \( \mathit{Re} \in [200, 900] \). We evaluate generalization on the extrapolation test sets \dIncLow{} with \( \mathit{Re} \in [100, 180] \) for \( T=60 \), and \dIncHigh{} with \( \mathit{Re} \in [920, 1000] \) for \( T=60 \). While all training is done with constant \(\mathit{Re}\), we add a case with varying \(\mathit{Re}\) as a particularly challenging test set: \dIncVar{} features a sequence of \( T=250 \) steps with a smoothly varying \( \mathit{Re} \) from \( 200 \) to \( 900 \) over the course of the simulation time.

\subsection{Transonic Cylinder Flow}
As a significantly more complex scenario we target transonic flows. These flows require the simulation of a varying density, and exhibit the formation of shock waves that interact with the flow, especially at higher Mach numbers \( \mathit{Ma} \). These properties make the problem highly chaotic and longer prediction rollouts especially challenging. We simulate a fully developed compressible Karman vortex street using SU2 \parencite{economon2015_SU2} with \( \mathit{Re} = 10000 \), while varying \( \mathit{Ma} \) in a transonic regime where shock waves start to occur. Models are trained on sequences with \( \mathit{Ma} \in [0.53, 0.63] \cup [0.69, 0.90]\). We evaluate extrapolation on \dTraExt{} with \( \mathit{Ma} \in [0.50, 0.52] \) for \( T=60 \), interpolation via \dTraInt{} with \( \mathit{Ma} \in [0.66, 0.68] \) for \( T=60 \), and longer rollouts of about \(8\) vortex shedding periods using \dTraLong{} with \( \mathit{Ma} \in [0.64, 0.65] \) for \( T=240 \).

\subsection{Isotropic Turbulence}
As a third scenario we evaluate the inference of planes from 3D isotropic turbulence. This case is inherently difficult, due to its severely un\-der\-de\-ter\-mined nature, as the information provided in a 2D plane allows for a large space of possible solutions, depending on the 3D motion outside of the plane. Thus, it is expected that deviations from the reference trajectories occur across methods, and we use \( R=100 \) diffusion steps in \mACDM{} as a consequence. For this setup, we observed a tradeoff between accuracy and sampling speed, where additional diffusion steps continued to improve prediction quality. As training data, we utilize 2D z-slices of 3D data from the Johns Hopkins Turbulence Database \parencite{perlman2007_Data}. Models are trained on sequences with \( z \in [1, 199] \cup [351, 1000]\), and we test on \dIso{} with sequences from \( z \in [200, 350] \) for \( T=100 \).

\begin{table*}[th]
    \caption{Quantitative comparison for different network architectures (\best{best} and \bestSec{second best} results are highlighted for each test set).}
    \label{tab: accuracy}
    \centering
    \footnotesize

    \begin{tabular}{l c c c c c c c c c c}
        \toprule
        & \multicolumn{2}{c}{\dIncLow{}} & \multicolumn{2}{c}{\dIncHigh{}} & \multicolumn{2}{c}{\dTraExt{}{}} & \multicolumn{2}{c}{\dTraInt{}} & \multicolumn{2}{c}{\dIso{}}\\
        \cmidrule(lr){2-5} \cmidrule(lr){6-9} \cmidrule(lr){10-11}
        & MSE & LSiM & MSE & LSiM & MSE & LSiM & MSE & LSiM & MSE & LSiM\\

        \textbf{Method}
        & (\(10^{-4}\)) & (\(10^{-2}\)) & (\(10^{-5}\)) & (\(10^{-2}\))
        & (\(10^{-3}\)) & (\(10^{-1}\)) & (\(10^{-3}\)) & (\(10^{-1}\))
        & (\(10^{-2}\)) & (\(10^{-1}\))\\
        \midrule

        \mResNet{}
        & $10\pm9.1$ & $17\pm7.8$  &  $16\pm3.0$ & $5.9\pm1.6$
        & $2.3\pm0.9$ & $1.4\pm0.2$  &  \bestSec{$1.8\pm1.0$} & \best{$1.0\pm0.3$}
        & $6.7\pm2.4$ & $9.1\pm2.2$\\

        \mResNet{dil.}
        & $1.6\pm1.8$ & $7.7\pm5.5$  &  $1.5\pm0.8$ & $2.6\pm0.7$
        & $1.7\pm1.0$ & $1.2\pm0.3$  &  $1.7\pm1.4$ & $1.0\pm0.5$
        & $5.7\pm2.1$ & $8.2\pm2.0$\\
        \midrule

        \mFNO{16}
        & $2.8\pm3.1$ & $8.8\pm7.1$  &  $8.9\pm3.8$ & $2.5\pm1.2$
        & $4.8\pm1.2$ & $3.4\pm1.1$  &  $5.5\pm2.6$ & $2.6\pm1.1$
        & \diverge{$2m\pm6m$} & \diverge{$15\pm1.5$}\\

        \mFNO{32}
        & $160\pm50$ & $80\pm5.4$  &  \diverge{$1k\pm140$} & \diverge{$57\pm4.9$}
        & $4.9\pm1.9$ & $3.6\pm0.9$  &  $6.8\pm3.4$ & $3.1\pm1.1$
        & $14\pm5.3$ & $8.9\pm1.2$\\
        \midrule

        \mTf{MGN}
        & $5.7\pm4.3$ & $13\pm6.4$  &  $10\pm2.9$ & $3.5\pm0.4$
        & $3.9\pm1.0$ & $1.8\pm0.3$  &  $6.3\pm4.4$ & $2.2\pm0.7$
        & $8.7\pm3.8$ & $7.0\pm2.2$\\

        \mTf{Enc}
        & $1.5\pm1.7$ & $6.3\pm4.2$  &  \bestSec{$0.6\pm0.3$} & \bestSec{$1.0\pm0.3$}
        & $3.3\pm1.2$ & $1.8\pm0.3$  &  $6.2\pm4.2$ & $2.2\pm0.7$
        & $11\pm5.2$  & $7.2\pm2.1$\\

        \mTf{VAE}
        & $5.4\pm5.5$ & $13\pm7.2$   &  $14\pm19$ & $4.1\pm1.4$
        & $4.1\pm0.9$ & $2.4\pm0.2$  &  $7.2\pm3.0$ & $2.7\pm0.6$
        & $11\pm5.1$  & $7.5\pm2.1$\\
        \midrule

        \mUnet{}
        & $1.0\pm1.1$ & $5.8\pm3.2$   &  $2.7\pm0.6$ & $2.6\pm0.6$
        & $3.1\pm2.1$  & $3.9\pm2.8$   &  $2.3\pm2.0$ & $3.3\pm2.8$
        & $26\pm35$   & $11\pm3.9$\\

        \mUnet{ut}
        & \best{$0.8\pm1.1$} & \best{$4.5\pm4.0$}  &  \best{$0.2\pm0.1$} & \best{$0.5\pm0.2$}
        & \bestSec{$1.6\pm0.7$} & \best{$1.1\pm0.2$}  &  \best{$1.5\pm1.5$} & $1.0\pm0.5$
        & $4.5\pm2.8$ & \best{$2.4\pm0.5$}\\

        \mUnet{tn}
        & $1.0\pm1.0$ & \bestSec{$5.6\pm3.1$}  &  $0.9\pm0.6$ & $1.5\pm0.6$
        & \best{$1.4\pm0.8$} & \bestSec{$1.1\pm0.3$}  &  $1.8\pm1.1$ & \bestSec{$1.0\pm0.4$}
        & \best{$3.1\pm0.9$} & $4.5\pm2.5$\\
        \midrule

        \mRefiner{}
        & $1.3\pm1.4$ & $7.1\pm4.2$  &  $3.5\pm2.2$ & $2.5\pm1.0$
        & $5.4\pm2.1$ & $2.3\pm0.5$  &  $7.1\pm2.1$ & $3.0\pm1.7$ 
        & $121\pm200$ & $10.2\pm3.5$ \\ 
        \midrule

        \mACDM{ncn}
        & \bestSec{$0.9\pm0.8$} & $6.6\pm2.7$  &  $5.6\pm2.6$ & $3.6\pm1.2$
        & $4.1\pm1.9$ & $1.9\pm0.6$  &  $2.8\pm1.3$ & $1.7\pm0.4$
        & $18.3\pm2.5$ & $8.9\pm1.5$\\
        
        \mACDM{}
        & $1.7\pm2.2$ & $6.9\pm5.7$  &  $0.8\pm0.5$ & \bestSec{$1.0\pm0.3$}
        & $2.3\pm1.4$ & $1.3\pm0.3$  &  $2.7\pm2.1$ & $1.3\pm0.6$
        & \bestSec{$3.7\pm0.8$} & \bestSec{$3.3\pm0.7$}\\

        \bottomrule
    \end{tabular}

\end{table*}

\begin{figure*}[th]
    \centering
    
    \includegraphics[width=0.99\textwidth]{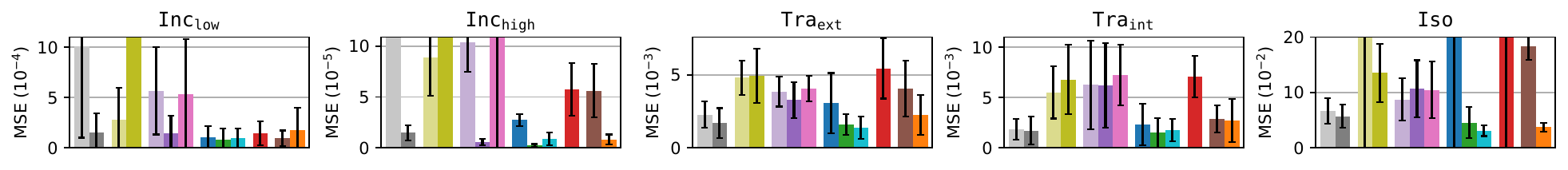}\\
    \vspace{-0.1cm}
    \includegraphics[width=0.99\textwidth]{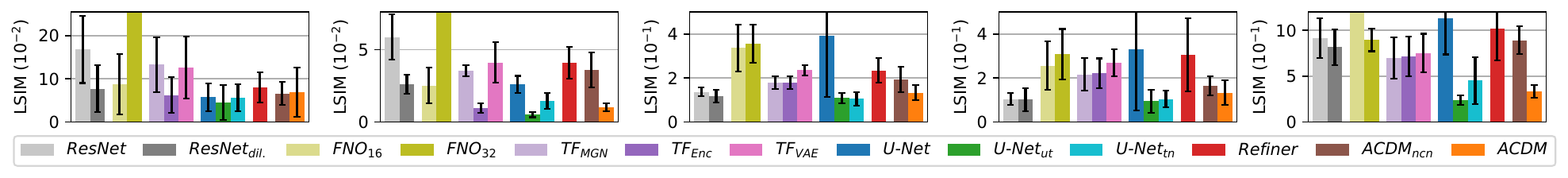}

    \caption{Accuracy visualization for the architectures from \cref{tab: accuracy}. Shown are MSE and LSiM errors with corresponding standard deviation.}
    \label{fig: accuracy}
\end{figure*}

\section{Results} \label{sec: results}
We benchmark the investigated methods in terms of accuracy, posterior sampling, statistical match to the underlying physics, and temporal stability using a range of different metrics. \Cref{app: coherence,app: temporal stability,app: ablations} contain additional results for various aspects discussed in the following. Unless denoted otherwise, mean and standard deviation over all sequences from each data set, multiple training runs, and multiple random model evaluations are reported. We evaluate two training runs with different random seeds for \dIso{}, and three for \dInc{} and \dTra{}. For the probabilistic methods \mTf{VAE}, \mRefiner{}, and \mACDM{}, five random model evaluations are taken into account per trained model.

\subsection{Accuracy}
To assess the quality of flow predictions, we first measure direct errors towards the ground truth sequence. We use a mean-squared-error (MSE) and LSiM, a similarity metric for numerical simulations \parencite[also see \cref{app: lsim}]{kohl2020_Learning}. For both metrics, lower values indicate better reconstruction accuracy. Reported errors are rollout errors, i.e., computed per time step and field, and averaged over the temporal rollout. The full accuracy results are reported in \cref{tab: accuracy}, where mean and standard deviation over all sequences from each data set, multiple training runs, and multiple random model evaluations are shown. Errors of models that diverge during inference, are displayed with factors of \(10^3\) (\(k\)) or \(10^6\) (\(m\)) in addition to the error scaling indicated in the second table row. \Cref{fig: accuracy} contains corresponding bar charts with error visualizations.

For the easiest test case \dInc{}, all model classes can do well, as shown by the overall low errors. The performance of \mResNet{dil.}, \mTf{Enc}, \mUnet{ut}, and \mACDM{} is quite similar. \mFNO{16} and \mRefiner{} also work well on \dIncLow{}, but are slightly less accurate on \dIncHigh{}. On the more complex \dTra{} case, all transformer-based and \mFNO{} architectures, as well as \mRefiner{} are already noticeably less accurate, compared to the remaining methods. \mACDM{} is on par with the established \mResNet{} variants, as well as \mUnet{} models with unrolling or training noise. However, for the longer rollouts in \dTraLong{} the behavior of the architectures is different as temporal stability issues can occur, as discussed below.

On \dIso{}, all models are struggling due to the highly underdetermined nature of this experiment, indicated by the higher errors overall. The transformer-based methods lack accuracy, as the compressed latent representations are unable to capture the high frequency details of this data set. The other architectures with lower relative accuracy accumulate errors over the rollout. This is especially noticeable for \mUnet{}, \mRefiner{}, and the \mResNet{} and \mFNO{} variants. The architectures that remain most stable and accurate are \mUnet{ut} and \mUnet{tn}, which are equipped with state of the art stabilization techniques like unrolled training or training noise. Surprisingly, \mACDM{} remains fully stable on \dIso{} without modification, and achieves comparable accuracy to \mUnet{ut} and \mUnet{tn}. Below, we will mostly focus on the more successful architecture variant in each model class if the behavior is relatively similar, i.e. \mResNet{dil.} as the superior ResNet model, \mTf{Enc} for the latent-space transformer architectures,  and \mFNO{16} (\mFNO{32} for \dIso{}).

The improved performance of \mResNet{dil.} compared to \mResNet{}, and the generally weak results of \mFNO{} on our more complex tasks confirm the findings from previous work \parencite{stachenfeld2022_Learned}. The benefits of unrolling \mUnet{} or adding training noise, also become evident on our more complex cases \dTra{} and \dIso{}: while both require additional parameters and training effort, substantial gains in accuracy can be achieved. The regular \mUnet{}, despite its network structure being identical to \mACDM{}, frequently performs worse than \mACDM{} on \dTra{} and \dIso{}. Thus, we include an ablation on \mACDM{}, that behaves similarly to \mUnet{} in terms of error propagation: For the \mACDM{ncn} model no conditioning noise is applied, i.e., \( \vc_{0} \) is used over the entire diffusion process. This variant performs substantially worse than \mACDM{} across cases, as it does not prevent the buildup of errors similar to the \mUnet{} or \mResNet{} models, due to the tight coupling between conditioning and prediction. This highlights the benefits of creating the next step prediction from scratch, leading to less error propagation and increased temporal stability for \mACDM{}.

\begin{figure}[th]
    \centering
    \includegraphics[width=0.483\textwidth]{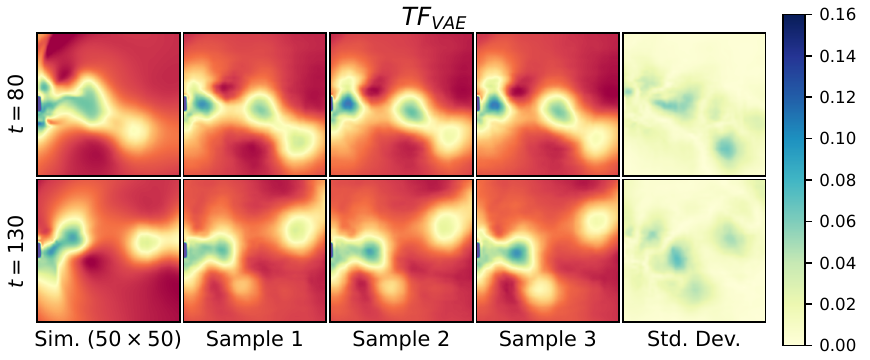}\\
    \vspace{0.15cm}
    \includegraphics[width=0.483\textwidth]{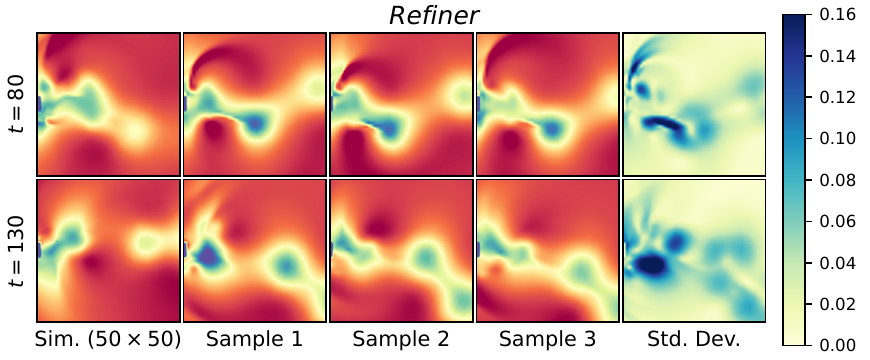}\\
    \vspace{0.15cm}
    \includegraphics[width=0.483\textwidth]{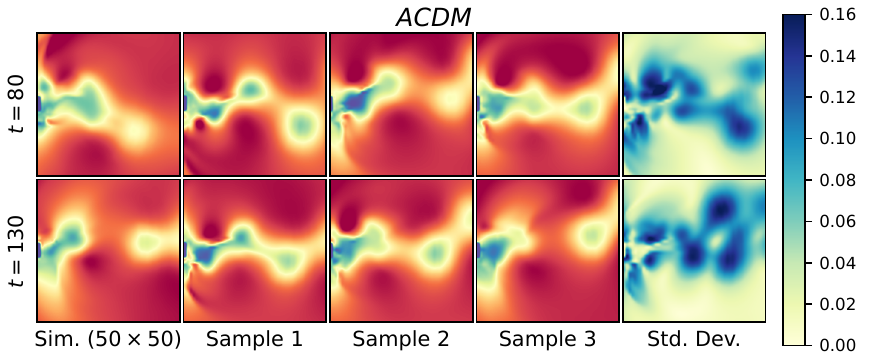}

    \caption{Large-scale posterior sample comparison with corresponding standard deviation. Shown are the vortices downstream of the cylinder on a trajectory from \dTraLong{} with \( \mathit{Ma} = 0.64 \) (pressure) at different time steps \(t\) (also see \href{https://ge.in.tum.de/publications/2023-acdm-kohl/}{accompanying posterior sampling videos}).}
    \label{fig: posterior}
\end{figure}

\begin{figure}[th]
    \centering
    \includegraphics[width=0.483\textwidth]{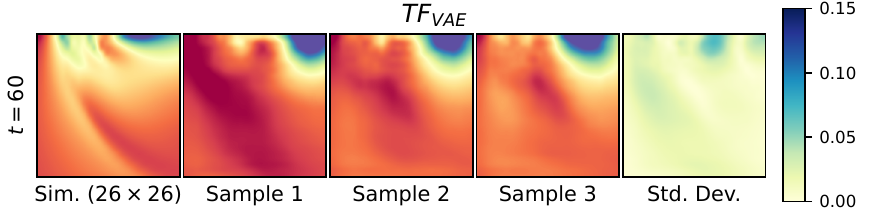}\\
    \vspace{0.15cm}
    \includegraphics[width=0.483\textwidth]{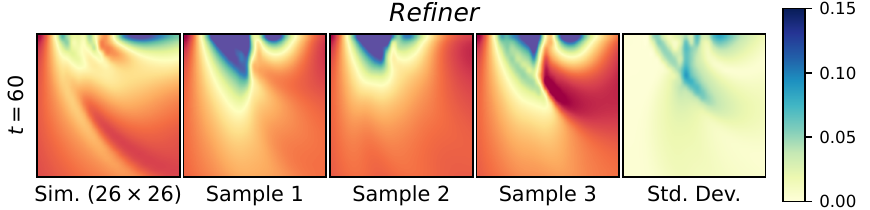}\\
    \vspace{0.15cm}
    \includegraphics[width=0.483\textwidth]{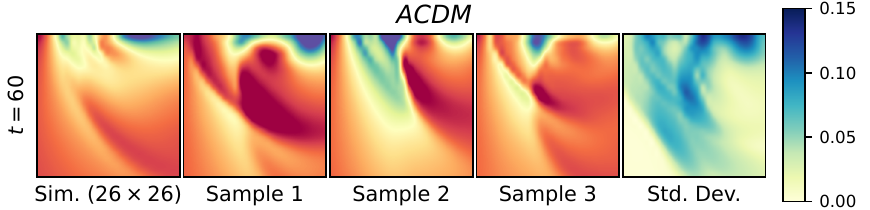}

    \caption{Small-scale posterior sample comparison with according standard deviation. Shown are pressure waves below the cylinder on a sequence from \dTraLong{} with \( \mathit{Ma} = 0.64 \) (pressure) at different time steps \(t\).}
    \label{fig: posterior 2}
\end{figure}

\subsection{Posterior Sampling}
One of the most attractive aspects of a DDPM-based simulator is posterior sampling, i.e., the ability to create different samples from the solution manifold for one initial condition. Below, we qualitatively and quantitatively evaluate the posterior samples of the investigated probabilistic methods \mTf{VAE}, \mRefiner{}, and \mACDM{}. For this purpose, we focus on the transonic flow experiment as a representative case with medium difficulty. For \dInc{}, it is visually difficult to discern predictions due to the simpler learning task, while it can be difficult to judge the larger discrepancies from the simulation reference that can occur for the underdetermined \dIso{} experiment.

\paragraph{Qualitative Analysis}
\Cref{fig: posterior,fig: posterior 2} visualize random, exemplary posterior samples from the different methods at different spatial zoom levels, alongside the snapshot of the corresponding \dTraLong{} reference simulation. Furthermore, the spatially varying standard deviations across the five computed samples from each method are included. First, it becomes apparent that \mTf{VAE} achieves barely any visual or physical differences across samples, as the general vortex structure downstream of the cylinder remain highly similar, even far into the trajectory. \mRefiner{} fares better and creates differences in the predictions, especially for large \(t\). However, as clearly visible in \cref{fig: posterior 2}, both approaches struggle to create physically important, small-scale details with high frequencies, such as the strongly varying formation of shock waves near the immersed cylinder. The predictions from \mTf{VAE} lack these features entirely, while \mRefiner{} is sometimes able to create them, however in a quite similar and sometimes unphysical manner. The diffusion approach produces the most realistic, and diverse features, while even being able to recreate physically plausible shock wave configurations. As expected, the spatial standard deviation increases over time due to the chaotic nature of this test case. The locations of high variance for \mACDM{} match areas that are more difficult to predict, such as vortices and shock wave regions.

\begin{figure*}[th]
    \centering

    \includegraphics[width=0.329\textwidth]{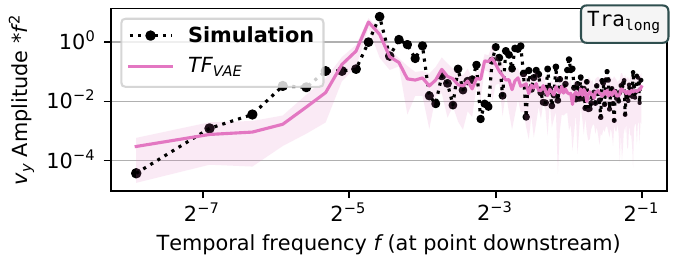}
    \hfill
    \includegraphics[width=0.329\textwidth]{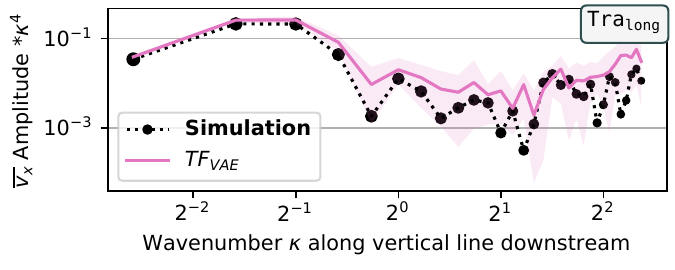}
    \hfill
    \includegraphics[width=0.329\textwidth]{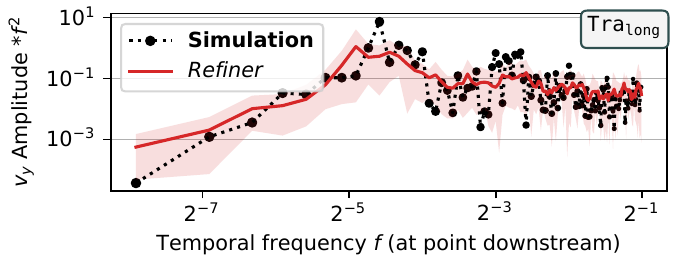}\\

    \includegraphics[width=0.329\textwidth]{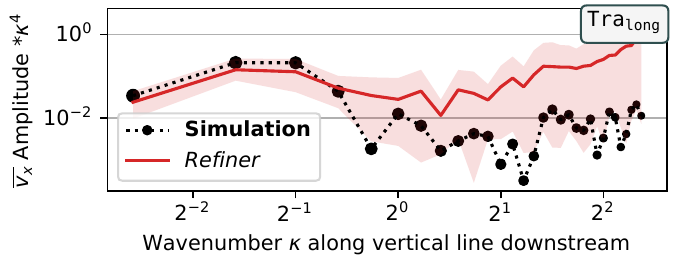}
    \hfill
    \includegraphics[width=0.329\textwidth]{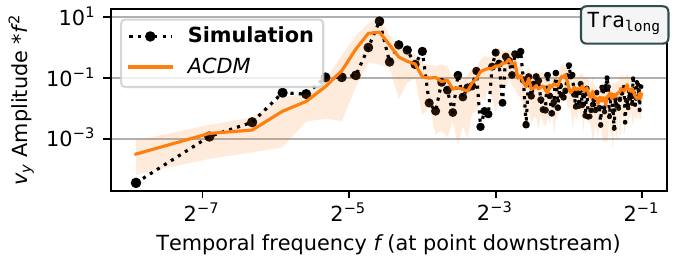}
    \hfill
    \includegraphics[width=0.329\textwidth]{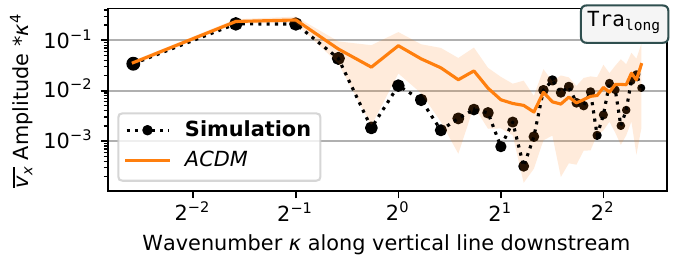}

    \caption{Temporal (top) and spatial (bottom) frequency analysis across posterior samples for a full sequence from \dTraLong{} with \( \mathit{Ma} = 0.64 \). The shaded area shows the \nth{5} to \nth{95} percentile across all trained models and posterior samples.}
    \label{fig: frequency tra}
\end{figure*}

\paragraph{Quantitative Analysis}
To analyze the quality of a distribution of predicted simulation trajectories from a probabilistic algorithm, it is naturally not sufficient to directly compare to a single target sequence, as even highly accurate numerical simulations would eventually decorrelate from a target simulation over time \parencite{hu2020_Risks}. Instead, our experimental setup allows for using temporal and spatial evaluations to measure whether different samples \textit{statistically} match the reference simulation, as established by turbulence research \parencite{dryden1943_Review}. Two metrics for a sequence from \dTraLong{} are analyzed here: We evaluate the wavenumber of the horizontal motion across a vertical line in the flow (averaged over time), and the temporal frequency of the vertical motion at a point probe. Both measurements are taken one cylinder diameter downstream from the back end of each immersed cylinder. The mean and variance across trained models and samples of the resulting spectra are shown in \cref{fig: frequency tra}. While similar at first glance, some key differences between the spectra of the different architectures can be observed. For the temporal analysis, both \mTf{VAE} and \mRefiner{} fail to accurately reproduce the main vortex shedding frequency indicated by the peak around a frequency of \(2^{-5}\), while \mACDM{} comes very close to the reference. The high frequency content on the right side of the spectrum is on average also best reproduced by \mACDM{}, as for example visible by the minor peak around a frequency of \(2^{-3}\). In terms of the spatial spectra at the bottom of \cref{fig: frequency tra}, \mRefiner{} fails to capture the behavior of the reference simulation. The most important low spatial frequencies have high variance, and the amplitudes for medium and high frequencies do not match the reference entirely. \mTf{VAE} and \mACDM{} result in much better spatial spectra, and both accurately capture the major low frequencies. \mACDM{} overshoots in the medium frequency regime while \mTf{VAE} has minor discrepancies for high frequencies. Overall, the samples from \mACDM{} statically most accurately reflect the physical behavior of the reference simulation.

\subsection{Comparing Spectral Statistics}
\begin{figure}[ht]
    \centering
    \includegraphics[width=0.48\textwidth]{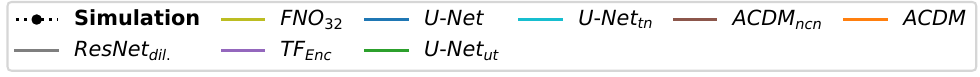}\\
    \includegraphics[width=0.48\textwidth]{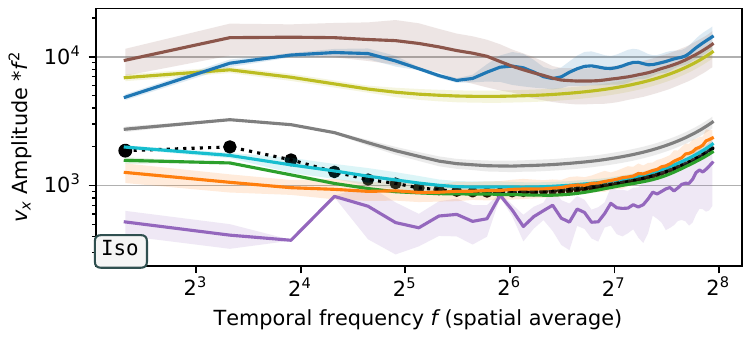}\\
    \includegraphics[width=0.48\textwidth]{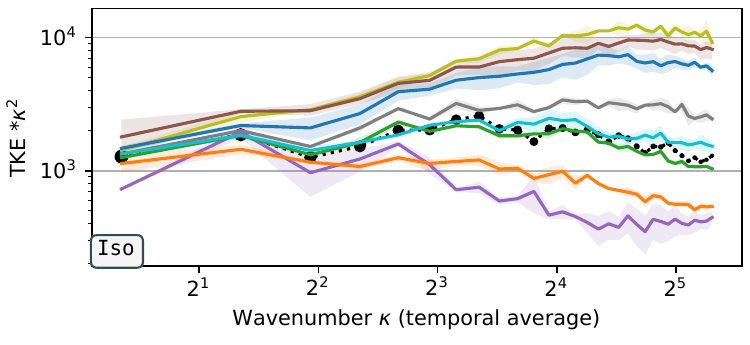}

    \caption{Temporal (top) and spatial (bottom) frequency analysis on a sequence from \dIso{} with \( z = 300 \). The shaded area shows the \nth{5} to \nth{95} percentile across all trained models and posterior samples. PDE-Refiner is omitted here, as some of its models and samples are unstable, leading to substantially worse results compared to all other methods on both evaluations.}
    \label{fig: frequency iso}
\end{figure}

Spectral statistics also highlight the differences between the other model architectures under consideration. On \dIso{}, the temporal frequency of the x-velocity is evaluated and shown at the top of \cref{fig: frequency iso}. As this case is isotropic, we average the evaluation across every spatial point for a more stable analysis. Furthermore, we also analyze the spatial behavior on \dIso{} via an energy spectrum at the bottom of \cref{fig: frequency iso}. In that case, the turbulent kinetic energy (TKE) is evaluated and aggregated in x- and y-direction, and averaged across all time steps of the simulation. \mTf{Enc} is lacking across the spatial and temporal frequency band, and \mTf{MGN} and \mTf{VAE} which are not shown behave similarly. This behavior is caused by the compression to the latent space, where spatial details are lost, and meaningful temporal evolution is challenging due to the overall complexity of the \dIso{} experiment.

Architectures which operate as direct next-step predictors such as \mResNet{dil.}, \mFNO{16}, \mUnet{}, and \mACDM{ncn} clearly overshoot, due to the direct error propagation that causes instabilities. Introducing explicit stabilization techniques via unrolling or training noise to \mUnet{} substantially improves spatial and temporal spectral behavior, making them the most accurate architectures in this evaluation. However, early signs of instabilities can be observed for \mUnet{tn}, as additional energy is introduced in high spatial frequencies compared to the reference which will eventually result in instabilities. \mACDM{} follows close behind the stabilized \mUnet{} variants: High temporal and low spatial frequencies are modeled well, but it deviates in terms of lower temporal and higher spatial frequencies. This is most likely caused by the strongly under-determined nature of \dIso{}. It causes \mACDM{} to unnecessarily dissipate spatial high-frequency motions, which in turn impacts low temporal frequencies over longer rollouts. \mRefiner{} is omitted in both visualizations, as several posterior samples across training runs were unstable, causing substantially worse result compared to all other methods.

\subsection{Temporal Stability}
\begin{figure}[ht]
    \centering
    \includegraphics[width=0.48\textwidth]{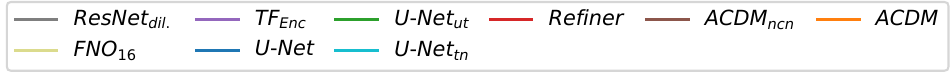}\\
    \includegraphics[width=0.48\textwidth]{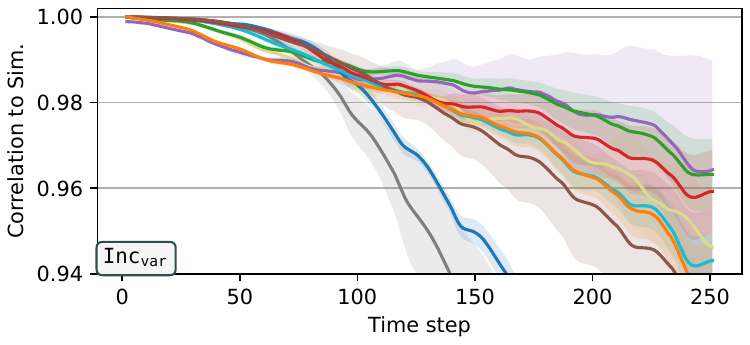}

    \caption{Correlation of predictions from different methods with the reference simulation over the rollout on \dIncVar{}. The shaded area shows the standard deviation across all trained models and posterior samples.}
    \label{fig: correlation inc}
\end{figure}

A central motivation for analysing diffusion models in the context of transient simulations is the hypothesis that the stochastic training procedure leads to a more robust temporal behavior at inference time. This is especially crucial for practical applications of fluid simulations, where rollouts with thousands of steps are common. We now evaluate this aspect in more detail, first, by measuring the Pearson correlation coefficient \parencite{pearson1920_Notes} between prediction and reference over time. We evaluate this for the \dIncVar{} test, which contains \( T = 250 \) steps with a previously unseen change of the Reynolds number over the rollout, displayed in \cref{fig: correlation inc}.

For this relatively simple task, all architectures perform well and manage to achieve a high correlation over the entire rollout. While \mUnet{} and \mResNet{dil.} are accurate initially, both exhibit faster decorrelation rates over time compared to the other investigated architectures. This already indicates a lack of tolerance to rollout errors observed on our more complex cases. \mFNO{16} performs quite well, as \dIncVar{} mainly contains low frequencies. However, the \mFNO{} variants have difficulties with high-frequency information, as shown below and reported in related work \parencite{stachenfeld2022_Learned}. Similarly, \mRefiner{} substantially outperforms \mUnet{} on this simple case confirming the authors results \parencite{lippe2023_PDERefiner}, however PDE-Refiner does not consistently work on more complex tests in our experiments as reported below. \mACDM{ncn} without noise on the conditioning is on par with \mACDM{}, indicating that tolerances to rollout errors are less crucial for this easier case compared to the remaining experiments. \mTf{Enc} and \mUnet{ut} keep the highest correlation levels on \dIncVar{}, showing the performance benefits of learning with unrolling, as both methods learn from longer, coherent trajectories during training. \mUnet{tn} does not have this advantage and performs similar to \mACDM{}.

To assess the dynamics of more complex cases, we measure the magnitude of the rate of change of \( s \), computed as \( \lVert \nicefrac{(s^{t}-s^{t-1})}{\Delta t} \rVert_1 \) for every normalized time step. Compared to the correlation, this metric stays meaningful even for long rollout times on complex data, where sequence can diverge from the specific solution trajectory while still remaining physical. It indicates whether a simulator preserves the expected evolution of states as given by the reference simulation: If predictions explode, the rate of change substantially grows beyond the reference, and if they collapse into an incorrect steady state, the change approaches zero.

\begin{figure}[ht]
    \centering
    \includegraphics[width=0.48\textwidth]{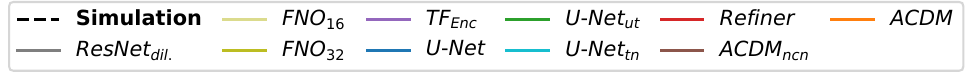}\\
    \includegraphics[width=0.48\textwidth]{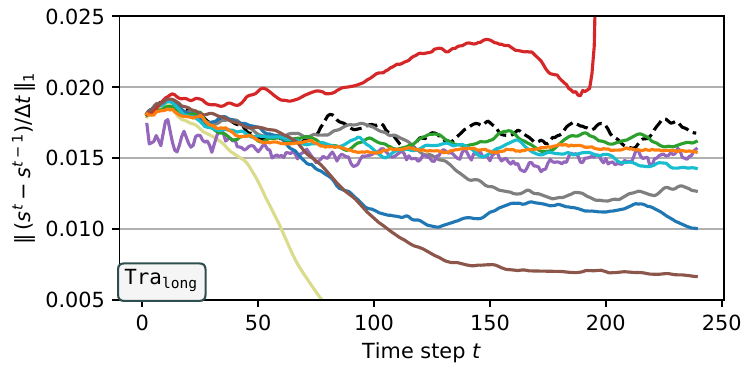}\\
    \includegraphics[width=0.48\textwidth]{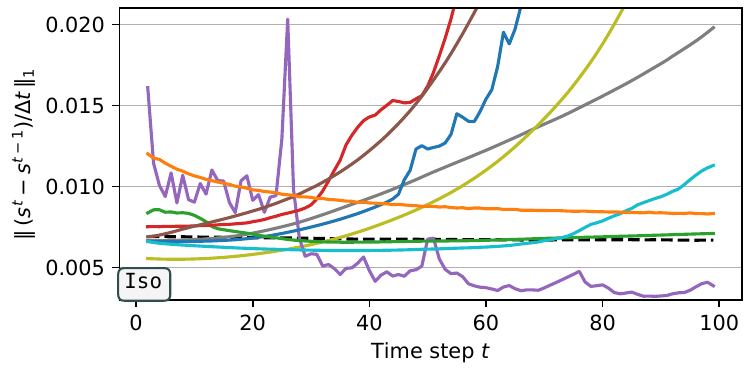}

    \caption{Stability analysis via error to previous time step on \dTraLong{} (top) and \dIso{} (bottom). Standard deviations are omitted for visual clarity.}
    \label{fig: error to previous}
\end{figure}

\begin{table*}[ht]
    \caption{Overview with our assessment of advantages for different flow prediction approaches at similar training memory (\goodIII{}: excellent, \goodII{}: good, \goodI{}: acceptable, \badI{}: suboptimal, \badII{}: bad, \badIII{}: not available, ---: not investigated in detail).}
    \label{tab: overview}
    \centering
    \scriptsize

    \begin{tabular}{l | c c c c c c c c}
        \toprule

        \multirow{2}*{\textbf{Aspect}} & Latent-space & \multirow{2}*{FNO} & Dilated &\multirow{2}*{U-Net} & U-Net & U-Net & PDE & Autoreg.\\
         & Transformer & & ResNet &  & (train. noise) & (unrolled) & Refiner & Diffusion\\
        \midrule
        
        Training Speed (per Epoch) & \goodII{} & \goodIII{} & \badII{} & \goodI{} & \goodI{} & \badII{} & \goodI{} & \goodI{}\\
        Inference Speed & \goodIII{} & \goodII{} & \goodII{} & \goodI{} & \goodI{} & \goodI{} & \badI{} & \badII{}\\
        \midrule

        Accuracy (\dInc{}) & \goodIII{} & \goodI{} & \goodIII{} & \goodII{} & \goodIII{} & \goodIII{} & \goodII{} & \goodIII{}\\
        Accuracy (\dTra{}) & \goodI{} & \goodI{} & \goodIII{} & \goodI{} & \goodIII{} & \goodIII{} & \goodI{} & \goodIII{}\\
        Accuracy (\dIso{}) & \badI{} & \badII{} & \goodI{} & \badII{} & \goodIII{} & \goodIII{} & \badII{} & \goodIII{}\\
        \midrule

        Posterior Sample Diversity & \badI{} (VAE) & \badIII{} & \badIII{} & \badIII{} & \badIII{} & \badIII{} & \goodI{} & \goodIII{}\\
        \midrule

        Temporal Stability (\dInc{}) & \goodIII{} & \goodIII{} & \goodII{} & \goodII{} & \goodIII{} & \goodIII{} & \goodIII{} & \goodIII{}\\
        Temporal Stability (\dTra{}) & \goodII{} & \badI{} & \goodI{} & \badI{} & \goodII{} & \goodIII{} & \goodI{} & \goodIII{}\\
        Temporal Stability (\dIso{}) & \badII{} & \badI{} & \badI{} & \badII{} & \goodII{} & \goodIII{} & \badI{} & \goodII{}\\
        \midrule

        \multirow{2}*{Key Hyperparameters} & Latent & Fourier & Dilation & \# Up/Down  & Noise & Rollout & Diffusion Steps, & Diffusion\\
        & Size & Modes & Rates & Blocks & Variance & Length & Noise Variance & Steps\\
        Key Hyperparam. Stability & \miss{} & \badI{} & \miss{} & \miss{} & \badI{} & \goodI{} & \badII{} & \goodII{}\\

        \bottomrule
    \end{tabular}
\end{table*}

\Cref{fig: error to previous} shows this evaluation for \dTraLong{} and \dIso{}. Model variants from each class that are omitted perform similar to the included variant in both cases. For \dTraLong{}, the reference simulation features steady oscillations as given by the main vortex shedding frequency. Note that these vortex shedding oscillations are averaged out over posterior samples and training runs in this evaluation, once architecture diverge from the exact reference trajectory. \mRefiner{} is highly unstable in this evaluations across its hyperparameters. Even though some trained models and samples are fully stable, the temporal stability on average is worst across the investigated architectures. Next-step predictor architectures like \mResNet{dil.}, \mFNO{16}, \mUnet{} diverge at different points during the \dTraLong{} rollout and mostly settle into a stable but wrong state of a mean flow prediction without vortices. \mACDM{ncn} which does not explicitly target error mitigation performs similar as well. \mTf{Enc} generally remains stable due to the long training rollout, indicated by a mostly constant rate of change. However, it exhibits minor temporal inconsistencies and slightly undershoots compared to the reference, most likely due to temporal updates being performed suboptimally in the latent space. \mUnet{tn} is stable, however early deterioration signs are visible towards the end of the trajectory. \mACDM{} and \mUnet{ut} remain fully stable, even for many additional rollouts steps.

In \cref{fig: error to previous} at the bottom, this evaluation is repeated for the \dIso{} experiment. Its isotropic nature in combination with forcing leads to an almost constant rate of change over time for the reference simulation. All methods struggle to replicate this accurately due the highly underdetermined learning task. In addition to issues with the reconstruction quality, \mTf{Enc} exhibits undesirable spikes corresponding to their temporal prediction window of \(k=25\) previous steps. It also undershoots after one rollout window, making it highly undesirable for this task. Similar as observed on \dTraLong{}, next-step predictor architectures like \mResNet{dil.}, \mFNO{32}, and \mUnet{} initially predict a quite accurate rate of change, but diverge at different points over the rollout. \mRefiner{} is also highly unstable across trained models and samples. Here, the common failure mode across models is an incorrect addition of energy to the system that causes a quick and significant divergence from the reference.

\mUnet{ut} and \mUnet{tn} show the best temporal stability in this experiment, however both exhibit minor issues at the start or end of the simulation rollout. For \mUnet{ut}, this can be further mitigated via a learning schedule to incrementally increase the rollout during training, or by pre-training with few rollout steps. Despite the initially slightly larger rate of change, and the decay corresponding to an overly dissipative prediction, \mACDM{} fares well and remains stable over the simulation rollout. For this case, we did observe mild temporal coherence issues in the vorticity computed from the \mACDM{} velocity predictions, which are analyzed in more detail in \cref{app: coherence}. Nevertheless, these experiments clearly show the increased error tolerance of \mACDM{} compared to \mACDM{ncn}, as the latter performs very similar to \mUnet{} for both evaluated experiments in \cref{fig: error to previous}, while \mACDM{} remains fully stable.

In an additional stability evaluation (see \cref{app: temporal stability}), we unrolled the three most stable architectures \mUnet{ut}, \mUnet{tn}, and \mACDM{} for \( T = 200\,000 \) steps to investigate extremely long prediction horizons which are highly desirable for e.g.~PDE, video, or climate prediction models \parencite{cachay2023_DYffusion,harvey2022_Flexible,watt-meyer2023_ACE}. All architectures remained fully stable and statistically accurate on extended sequences from \dIncHigh{}. In addition, we considered more challenging extended sequences from \dTraExt{}: while one training run from \mUnet{tn} did diverge, all runs from \mACDM{} and \mUnet{ut} were fully stable across the entire prediction horizon, indicating their potential for future applications.

\subsection{Discussion and Overview}
We summarize the findings of the performed experiments in \cref{tab: overview}, and include additional performance results from \cref{app: performance}. In general, latent-space transformers and \mFNO{} variants are fast to train and evaluate. As they use an identical backbone model, \mUnet{}, \mUnet{ut}, and \mUnet{tn} have the same inference speed. Note however, that \mUnet{ut} requires substantially more time and memory during training due to the additional rollout steps. Diffusion-based models like PDE-Refiner and \mACDM{} have an inference slow-down factor roughly proportional to the number of diffusion steps \(R\), i.e.,\( 2 \)--\( 8\) for PDE-Refiner and \( 20 \)--\( 100 \) for \mACDM{}. Furthermore, the stability with respect to key hyperparameters for some architectures is reported, experiments for which can be found in \cref{app: ablations}. To summarize, we draw the following main conclusions:
\begin{enumerate}[label=(\textbf{\textit{\roman*}})]
\item Latent-space transformers are highly accurate and stable, as long as the input space can be compressed easily.
\item Next-step predictors like \mResNet{dil.}, \mFNO{}, and \mUnet{} work very well for simple tasks, but require explicit stabilization techniques for more complex cases.
\item If training \emph{and} inference performance are of concern, using training noise provides substantial stabilization benefits.
\item Unrolling is resource-intensive during training and requires tuning of the rollout length, but pays off in terms of improved accuracy and stability.
\item Autoregressive diffusion models like \mACDM{} are as accurate and stable as unrolled training at lower training cost, but are expensive during inference.
\item PDE-Refiner can improve the temporal stability compared to \mUnet{}, however for complex cases it is very sensitive to repeated sampling or training, and its hyperparameters.
\item If accurate and diverse posterior samples are required, \mACDM{} models are the ideal option.
\end{enumerate}

\section{Conclusion and Future Work}
We investigated the attractiveness of autoregressive conditional diffusion models for the simulation of complex flow phenomena. Our results show that using even a simple diffusion-based approach is on par with established stabilization approaches, while at the same time enabling probabilistic inference. Furthermore, there are several interesting directions for future work. We believe that recent advances in sampling procedures, such as distillation \parencite{salimans2022_Progressive}, are a promising avenue for improving the inference performance of autoregressive diffusion models. However, the advantages in terms of stability of diffusion-based approaches most likely stem from its iterative nature over the diffusion rollout, and hence we anticipate that a constant factor over deterministic, single-pass inference will remain. Naturally, considering other PDEs, or larger, three-dimensional flows is likewise a highly interesting direction. For the latter, single-step diffusion approaches are particularly attractive, as they avoid the substantial costs of temporal training rollouts \parencite{sirignano2020_DPM} to achieve stability.

\section*{CRediT Authorship Contribution Statement}
\textbf{Georg Kohl:} Data curation, Formal analysis, Methodology, Software, Validation, Visualisation, Writing -- original draft, Writing -- review and editing. \textbf{Li-Wei Chen:} Data curation, Supervision, Writing -- review and editing. \textbf{Nils Thuerey:} Conceptualization, Supervision, Writing -- review and editing.

\section*{Declaration of Competing Interests}
The authors declare that they have no known competing financial interests or personal relationships that could have appeared to influence the work reported in this paper.

\section*{Acknowledgments}
This work was supported by the ERC Consolidator Grant \textit{SpaTe} (CoG-2019-863850). The authors would like to thank Bj\"orn List and Benjamin Holzschuh for helpful discussions and comments during the creation of this work.

\section*{Data Availability}
The source code, data sets, and trained models that support the findings in this work are available at \url{https://github.com/tum-pbs/autoreg-pde-diffusion}. Supplementary videos for this article can be found at \url{https://ge.in.tum.de/publications/2023-acdm-kohl}.

{
\emergencystretch=1em
\hbadness=99999 

\newrefcontext[sorting=nyt] 
\printbibliography
}


\cleardoublepage
\onecolumn
\appendix

\gdef\thesection{\Alph{section}}
\makeatletter
\renewcommand\@seccntformat[1]{\appendixname\ \csname the#1\endcsname.\hspace{0.5em}}
\makeatother

\section{Data Details} \label{app: data}
In the following, we provide details for each simulation setup: the incompressible wake flow \dInc{} in \cref{app: incompressible flow}, the transonic cylinder flow \dTra{} in \cref{app: transonic flow}, and the isotropic turbulence \dIso{} in \cref{app: isotropic turbulence}. Further details can be found in our source code  at \url{https://github.com/tum-pbs/autoreg-pde-diffusion}.

\subsection{Incompressible Flow Simulation} \label{app: incompressible flow}
To create the incompressible cylinder flow we employ the fluid solver PhiFlow\footnote{\url{https://github.com/tum-pbs/PhiFlow}} \parencite{holl2020_Learning}. Velocity data is stored on a staggered grid, we employ an advection scheme based on the MacCormack method, and use the adaptive conjugate gradient method as a pressure solver. We enforce a given Reynolds number in \( [100, 1000] \) via an explicit diffusion step.

\begin{figure}[ht]
    \centering
    \includegraphics[width=0.46\textwidth]{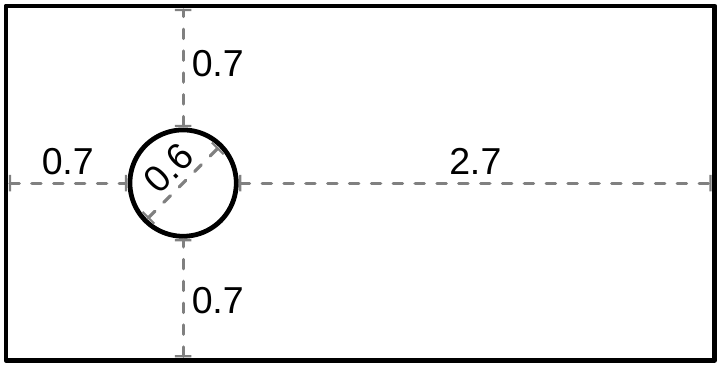}

    \caption{Simulation domain for incompressible flow simulation.}
    \label{fig-app: setup phiflow}
\end{figure}

Our domain setup is illustrated in \cref{fig-app: setup phiflow}. We use Neumann boundary conditions in vertical x-direction of the domain and around the cylinder, and a Dirichlet boundary condition for the outflow on the right of the domain. For the inflow on the left of the domain we prescribe a fixed freestream velocity of \( \binom{0}{0.5} \) during the simulation. To get oscillations started, the y-component of this velocity is replaced with \( 0.5 \cdot ( \cos(\pi \cdot x) + 1 ) \), where \( x \) denotes normalized vertical domain coordinates in \( [0,1] \), during a warmup of \( 20 \) time steps. We run and export the simulation for \( 1300 \) iterations at time step \( 0.05 \), using data after a suitable warmup period \( t > 300 \). The spatial domain discretization is \( 256 \times 128 \), but we train and evaluate models on a reduced resolution via downsampling the data to \( 128 \times 64 \). Velocities are resampled to a regular grid before exporting, and pressure values are exported directly. In addition, we normalize all fields and scalar components to a standard normal distribution. The velocity is normalized in terms of magnitude. During inference we do not evaluate the cylinder area; i.e., all values inside the cylinder are set to zero via a multiplicative binary mask before every evaluation or loss computation.

We generated a data set of \( 91 \) sequences with Reynolds number \( \mathit{Re} \in \{100, 110, \ldots, 990, 1000\} \). Running and exporting the simulations on a machine with an NVIDIA GeForce GTX 1080 Ti GPU and an Intel Core i7-6850k CPU with 6 cores at 3.6 GHz took about 5 days. Models are trained using the data of \( 81 \) sequences with \( \mathit{Re} \in \{200, 210, \ldots, 890, 900\}\) for \( t \in [800,1300] \). Training and test sequences employ a temporal stride of \( 2 \). As test sets we use:
\begin{itemize}
    \item \dIncLow{}: five sequences with \( t \in [1000,1120) \) and thus \( T=60 \), for \( \mathit{Re} \in \{100, 120, 140, 160, 180\} \).
    \item \dIncHigh{}: five sequences  with \( t \in [1000,1120) \) and thus \( T=60 \), for \( \mathit{Re} \in \{920, 940, 960, 980, 1000\} \).
    \item \dIncVar{}: one sequence for \( t \in [300,800) \) with \( T=250 \), and a smoothly varying \( \mathit{Re} \) from \( 200 \) to \( 900 \) during the simulation. This is achieved via linearly interpolating the diffusivity to the corresponding value at each time step.
\end{itemize}

For the \dIncVar{} test set, we replace the model predictions of \( \mathit{Re} \) that are learned to be constant for \mACDM{}, \mUnet{}, \mResNet{}, and \mFNO{} with the linearly varying Reynolds numbers over the simulation rollout during inference. The transformer-based methods \mTf{Enc} and \mTf{VAE} receive all scalar simulation parameters as an additional input to the latent space for each iteration of the latent processor. Note that the architectural design of \mTf{MGN} does not allow for varying simulation parameters over the rollout, as only one fixed parameter embedding is provided as a first input step for the latent processor, i.e. the model is expected to diverge quickly due to the data shift in such cases.

\subsection{Transonic Flow Simulation} \label{app: transonic flow}
To create the transonic cylinder flow we use the simulation framework SU2\footnote{\url{https://su2code.github.io}} \parencite{economon2015_SU2}. We employ the delayed detached eddy simulation model (SA-DDES) for turbulence closure, which is derived from the one-equation Spalart-Allmaras model \parencite{spalart2006_New}. By modifying the length scale, the model behaves like RANS for the attached flow in the near wall region and resolves the detached flows in the other regions. No-slip and adiabatic conditions are applied on the cylinder surface. The farfield boundary conditions are treated by local, one-dimensional Riemann-invariants. The governing equations are numerically solved by the finite-volume method. Spatial gradients are computed with weighted least squares, and the biconjugate gradient stabilized method (BiCGSTAB) is used as the implicit linear solver. For the freestream velocity we enforce a given Mach number in \( [0.5, 0.9] \) while keeping the Reynolds number at a constant value of \( 10^4 \).

To prevent issues with shockwaves from the initial flow phase, we first compute a steady RANS solution for each case for \( 1000 \) solver iterations and use that as the initialization for the unsteady simulation. We run the unsteady simulation for \( 150\,000 \) iterations overall, and use every \nth{50} step once the vortex street is fully developed after the first \( 100\,000 \) iterations. This leads to \( T = 1000 \) exported steps with velocity, density, and pressure fields. The non-dimensional time step for each simulation is $0.002*\Tilde{D}/\Tilde{U}_{\infty}$, where $\Tilde{D}$ is the dimensional cylinder diameter, and $\Tilde{U}_{\infty}$ the free-stream velocity magnitude.

\begin{figure}[ht]
    \centering
    \begin{adjustbox}{valign=c}
    \includegraphics[width=0.48\textwidth]{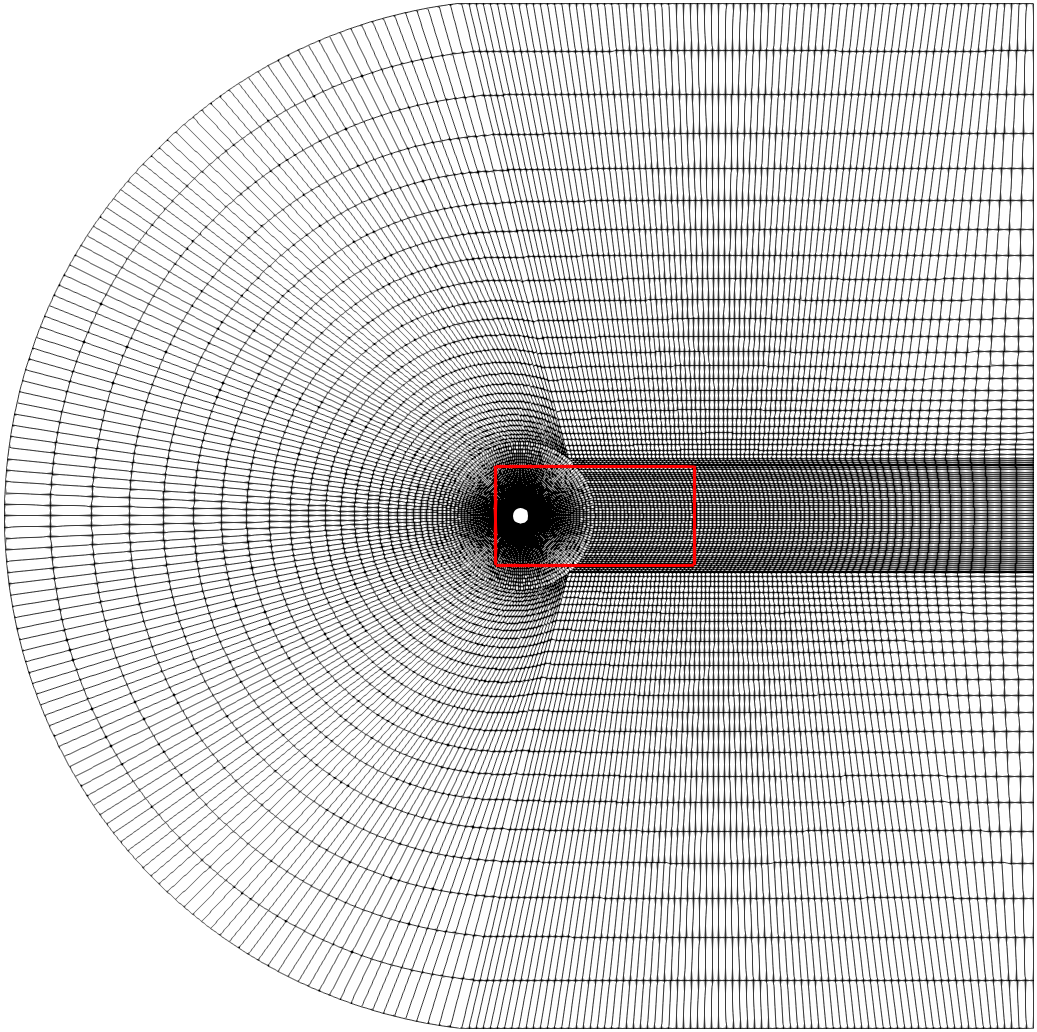}
    \end{adjustbox}
    \hfill
    \begin{adjustbox}{valign=c}
    \includegraphics[width=0.48\textwidth]{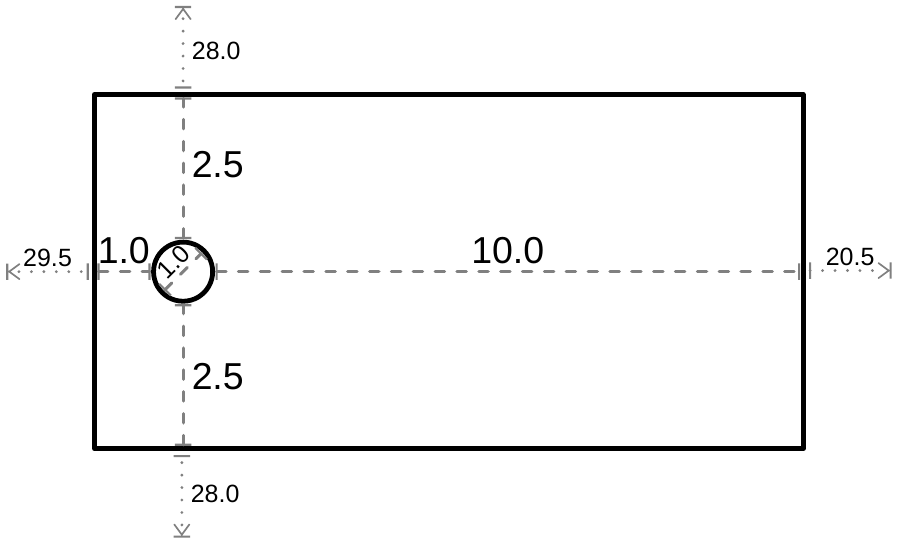}
    \end{adjustbox}

    \caption{Full simulation mesh with highlighted resampling area (left) and resampling domain setup (right) for the transonic flow simulation.}
    \label{fig-app: setup su2}
\end{figure}

The computational mesh is illustrated on the top in \cref{fig-app: setup su2}. Inference is focused on the near field region around the obstacle (marked in red on the top, and shown in detail on the bottom). To interpolate from the original mesh to the resampled training and testing domain, which is a regular, Cartesian grid with resolution \( 128 \times 64 \), we use an interpolation based on radial basis functions. It employs a linear basis function across the \( 5 \) nearest data points of the original mesh. In terms of field normalization and masking the cylinder area during inference, we treat this case in the same way as described in \cref{app: incompressible flow}. 

We created a data set of \( 41 \) sequences with Mach number \( \mathit{Ma} \in \{0.5, 0.51, \ldots, 0.89, 0.90\} \) at Reynolds number \( 10^4 \) with the \( T=1000 \) exported steps each. We sequentially ran the simulations on one CPU cluster node that contains 28 Intel Xeon E5-2690 v3 CPU cores at 2.6 GHz in about 5 days. Each simulation was computed in parallel with 56 threads, and one separate thread simultaneously resampled and processed the simulation outputs online during the simulation. All models are trained on the data of \( 33 \) sequences with \( \mathit{Ma} \in \{0.53, 0.54, \ldots, 0.62, 0.63\} \cup \{0.69, 0.70, \ldots, 0.89, 0.90\}\). Training and test sequences use a temporal stride of \( 2 \). The used test cases for this compressible, transonic flow setup are:
\begin{itemize}
    \item \dTraExt{}: six sequences from \( \mathit{Ma} \in \{0.50, 0.51, 0.52\} \), for \( t \in [500,620) \) and for \( t \in [620,740) \) with \( T=60 \).
    \item \dTraInt{}: six sequences from \( \mathit{Ma} \in \{0.66, 0.67, 0.68\} \), for \( t \in [500,620) \) and for \( t \in [620,740) \) with \( T=60 \).
    \item \dTraLong{}: four sequences from \( \mathit{Ma} \in \{0.64, 0.65\} \), for \( t \in [0,480) \) and for \( t \in [480,960) \) with \( T=240 \).
\end{itemize}

\subsection{Isotropic Turbulence} \label{app: isotropic turbulence}
For the isotropic turbulence experiment, we make use of the 3D \textit{isotropic1024coarse} simulation from the Johns Hopkins Turbulence Database\footnote{\url{https://turbulence.pha.jhu.edu/}} \parencite{perlman2007_Data}. It contains simulations of forced turbulence with a direct numerical simulation (DNS) using a pseudo-spectral method on \( 1024^3 \) nodes for \( 5028 \) time steps. The database allows for direct download queries of parameterized simulation cutouts; filtering and interpolation are already provided. We utilize sequences of individual 2D slices with a spatio-temporal starting point of \( (s_x,s_y,s_z,s_t) = (1,1,z,1) \) and end point of \( (e_x,e_y,e_z,e_t) = (256,128,z+1,1000) \) for different values of \( z \). A spatial striding of \( 2 \) leads to the training and evaluation resolution of \( 128 \times 64 \). We use the pressure, as well as the velocity field including the velocity z-component. We normalize all fields to a standard normal distribution before training and inference. In this case, the velocity components are normalized individually, which is statistically comparable to a normalization in terms of magnitude for isotropic turbulence.

We utilize \( 1000 \) sequences with \( z \in \{1, 2, \ldots, 999, 1000\} \) and \( T=1000 \). Models are trained on \( 849 \) sequences with \( z \in \{1, 2, \ldots, 198, 199\} \cup \{351, 352, \ldots, 999, 1000\}\). The test set in this case is \dIso{} using 16 sequences from \( z \in \{200, 210, \ldots, 340, 350\} \) for \( t \in [500,600)\), meaning \( T=100 \).

\section{Implementation and Model Details} \label{app: implementation}
Using the data generated with the techniques described above, the deep learning aspects of this work are implemented in PyTorch \parencite{paszke2019_PyTorch}. For every model we optimize network weights using the Adam optimizer \parencite{kingma2015_Adam} with a learning rate of \( 10^{-4} \) (using \(\beta_1=0.9\) and \(\beta_2=0.999\)), where the batch size is chosen as \(64\) by default. If models would exceed the available GPU memory, the batch size is reduced accordingly. For each epoch, the long training sequences are split into shorter parts according to the required training sequence length for each model and the temporal strides described in \cref{app: data}.
To prevent issues with a bias towards certain initial states, the start (and corresponding end) of each training sequence is randomly shifted forwards or backwards in time by half the sequence length every time the sequence is loaded. This is especially crucial for the oscillating cylinder flows when training models with longer rollouts. For instance, training a model with a training rollout length of \(60\) steps on a data set that contains vortex shedding oscillations with a period of \(30\) steps would lead to a correlation between certain vortex arrangements and the temporal position in the rollout during training (and inference). This could potentially lead to generalization problems when the model is confronted with a different vortex arrangement than expected at a certain time point in the rollout. The sequences for each test set are used directly without further modifications. In the following, we provide architectural and training details for the different model architectures discussed above.

\subsection{ACDM Implementation} \label{app: acdm implementation}
For the \mACDM{} models, we employ a ``modern'' U-Net architecture commonly used for diffusion models: The setup at its core follows the traditional U-Net architecture \parencite{ronneberger2015_UNet} with an initial convolution layer, several downsampling blocks, one bottleneck block, and several upsampling blocks followed by a final convolution layer. The downsampling and upsampling block at one resolution are connected via skip connections in addition to the connections through lower layers. The modernizations mainly affect the number and composition of the blocks: We use three feature map resolutions (\(128 \times 64\), \(64 \times 32\), and \(32 \times 16\)), i.e. three down- and three upsamling blocks, with a constant number of channels of \(128\) at each resolution level. The down- and upsampling block at each level consists of two ConvNeXt blocks \parencite{liu2022_ConvNet} and a linear attention layer \parencite{shen2021_Efficient}. The bottleneck block uses a regular multi-head self-attention layer \parencite{vaswani2017_Attention} instead. As proposed by \textcite{ho2020_Denoising}, we:
\begin{itemize}
    \item use group normalization \parencite{wu2018_Group} throughout the blocks,
    \item use a diffusion time embedding for the diffusion step \(r\) via a Transformer sinusoidal position embedding layer \parencite{vaswani2017_Attention} combined with an MLP consisting of two fully connected layers, that is added to the input of every ConvNeXt block,
    \item train the model via reparameterization,
    \item and employ a linear variance schedule.
\end{itemize}

Since the variance hyperparameters provided by \textcite{ho2020_Denoising} only work for a large number of diffusion steps \(R\), we adjust them accordingly to fewer diffusion steps: \( \beta_0 = 10^{-4} * (\nicefrac{500}{R}) \) and \( \beta_R = 0.02 * (\nicefrac{500}{R}) \). We generally found \( R=20 \) to be sufficient on the strongly conditioned data set \dInc{} and \dTra{}, but on the highly complex \dIso{} data, \mACDM{} showed improvements up to about \( R=100 \). The same value of \(R\) is used during training and inference. In early exploration runs, we found \( k=2 \) input steps to show slightly better performance compared to \( k=1 \) used by \mUnet{} below, and kept this choice for consistency across diffusion evaluations. However, the differences for changing the number of input steps from \( k \in \{1,2,3,4\} \) are minor compared to the performance difference between architectures. The resulting models are trained for \( 3100 \) epochs on \dInc{} and \dTra{}, and \( 100 \) epochs on \dIso{}. All setups use a batch size of \(64\) during training, and employ a Huber loss, which worked better than an MSE loss. However, the performance difference between the losses are marginal, compared to the difference between architectures.

For the \mACDM{ncn} variants, we leave all these architecture and training parameters untouched, and only change the conditioning integration: Instead of adding noise to \( \vc_0 \) in the forward and reverse diffusion process at training and inference time, \( \vc_0 \) is used without alterations over the entire diffusion rollout.

\subsection{Implementation of U-Net and Variants with Stabilization} \label{app: unet implementation}
For the implementation of \mUnet{} we use an identical U-Net architecture as described above in \cref{app: acdm implementation}. The only difference being that the diffusion time embeddings are not necessary. The resulting model is trained with an MSE loss on the subsequent time step. In early exploration runs, we found \( k=1 \) input steps to perform best for this direct next-step prediction setup with \mUnet{} (and similarly for \mResNet{} and \mFNO{} below), when investigating \( k \in \{1,2,3,4\} \). However, compared to the difference between architectures, these changes are minor.

The additional \mUnet{} variants with time unrolling during training share the same architecture. They are likewise trained with an MSE loss applied equally to every step of the predicted rollout with length \(m\) against the ground truth. A \mUnet{} trained with, e.g., \(m=8\) is denoted by \mUnet{m8} below. To keep a consistent memory level during training, the batch size is reduced correspondingly when \( m \) is increased. Thus, the training time of \mUnet{} significantly depends on \( m \). While \( m=2 \) allows for a batch size of \(64\), \( m=4 \) reduces that to \(32\), \( m=4 \) leads to \(16\), and finally, for \( m=16 \) the batch size is only \(8\).

We also analyze \mUnet{} variants with training noise to stabilize predictions \parencite{sanchez-gonzalez2020_Learning}. Normally distributed noise with standard deviation \(n\) is added to every model input during training, while leaving the prediction target untouched. At inference time, the models operate identically to their counterparts without training noise. In the following, \mUnet{} models trained with training noise of e.g., \(n=10^{-1}\), are denoted by \mUnet{n1e-1}. All \mUnet{} variants were trained for \( 1000 \) epochs on \dInc{} and \dTra{}, and \( 100 \) epochs on \dIso{}.

\subsection{PDE-Refiner Implementation} \label{app: refiner implementation}
PDE-Refiner is a recently proposed multi-step refinement process to improve the stability of learned PDE predictions \parencite{lippe2023_PDERefiner}. This approach relies on starting from the predictions of a trained one-step model, and iteratively refining them by adding noise of decreasing variances and denoising the result. The resulting model is then autoregressively unrolled to form a prediction trajectory. This method implies, that only probabilistic refinements are applied to a deterministic initial prediction. To train a model that can predict and refine at the same time, a random step \(r \in [0,R]\) in the refinement process is sampled, and the model is trained with a next-step MSE objective if \(r=R\) and with a standard denoising objective otherwise\footnote{Compared to \textcite{lippe2023_PDERefiner}, we switch the notation to \(R\) being the first step in the reverse process here, in line with our notation above, which also matches the notation in the original DDPM \parencite{ho2020_Denoising}.}. We re-implement this method, closely following the provided pseudocode in their paper, only changing the backbone network to our \mUnet{} implementation (see \cref{app: unet implementation}) for a fair comparison against our architectures. The resulting models are trained for \( 3000 \) epochs on \dInc{} and \dTra{}, and \( 100 \) epochs on \dIso{}, with a batch size of \(64\) and \( k=1 \) input steps.

The authors report that \mRefiner{} models with around \(R=4\) refinement steps perform best, when paired with a custom, exponential noise schedule, parameterized with a minimum noise variance\footnote{For brevity, we use \(\sigma\) for the minimum noise variance here, \textcite{lippe2023_PDERefiner} refer to it as \(\sigma^2_{min}\).} around \(\sigma=10^{-6}\). As such, we use these values for our main results above (only changing \(\sigma = 10^{-5}\) for \dIso{}). Below, we additionally sweep over combinations of \(R \in \{2,4,8\}\) and \(\sigma \in \{10^{-7}, 10^{-6}, 10^{-5}, 10^{-4}, 10^{-3}\}\), to investigate the stability with respect to these hyperparameters in our setting. We denote models trained with e.g., \(R=2\) and \(\sigma = 10^{-3}\) by \mRefiner{R2,$\sigma$1e-3} in the following.

\subsection{Implementation of dilated ResNets} \label{app: resnet implementation}
For the implementation of \mResNet{dil.} and \mResNet{}, we follow the setup proposed by \textcite{stachenfeld2022_Learned} that relies on a relatively simple architecture: both models consist of \(4\) blocks connected with skip connections as originally proposed by \cite{he2016_Deep}. Furthermore, one encoder layer before and one decoder convolution layer after the blocks are used to achieve the desired number of input and output channels. Each block contains \(7\) convolution layers with kernel size \(3\), stride \(1\), and \(144\) feature channels, followed by ReLU activations. For the \mResNet{dil.} model, the convolution layers in each block employ the following dilation and padding values: \((1,2,4,8,4,2,1)\). For \mResNet{}, all dilation and padding values are set to \(1\). Both models use a batch size of \(64\), receive \(k=1\) input steps, predict a single next step, and are trained via an mean-squared-error (MSE) on the prediction against the simulation trajectory as described in \cref{app: unet implementation}.

\subsection{Implementation of FNOs} \label{app: fno implementation}
For the implementation of the \mFNO{} variants, we follow the official PyTorch FNO implementation.\footnote{\url{https://github.com/NeuralOperator/neuraloperator}} The lifting and projection block setups are directly replicated from \textcite{li2021_Fourier}, and all models use \(4\) FNO layers. We vary the number of modes that are kept in in x- and y-direction in each layer as follows: \mFNO{16} uses \((16,8)\) modes and \mFNO{32} uses \((32,16)\) modes. To ensure a fair comparison, the hidden size of all models are parameterized to reach a number of trainable parameters similar to \mACDM{}, i.e. \(112\) for \mFNO{16} and \(56\) for \mFNO{32}. Both models use a batch size of \(64\), receive \(k=1\) input steps, predict a single next step, and are trained via an mean-squared-error (MSE) on the prediction against the simulation trajectory as described in \cref{app: unet implementation}.

\subsection{Latent Transformer Implementation} \label{app: transformer implementation}
To adapt the approach from \textcite{han2021_Predicting} to regular grids instead of graphs, we rely on CNN-based networks to replace their Graph Mesh Reducer (GMR) network for encoding and their Graph Mesh Up-Sampling (GMUS) network for decoding. Our encoder model consists of convolution+ReLU blocks with MaxPools and skip connections. In the following, convolution parameters are given as \textit{"input channels \(\rightarrow\) output channels, kernel size, stride, and padding"}. Pooling parameters are given as \textit{"kernel size, stride"}, and Upsampling parameters are give as \textit{"scale factor in x, scale factor in y, interpolation mode"}. The number of channels of the original flow state are denoted by \(\mathit{in}\), the encoder width is \(w_e\), the decoder width is \(w_d\), and \(L\) is the size of the latent space. The encoder layers are:
\begin{enumerate}
    \item Conv(\( \mathit{in} \rightarrow w_e, 11, 4, 5 \)) + ReLU + MaxPool(\(2, 2\))
    \item Conv(\( w_e + \mathit{in}_1 \rightarrow 3 * w_e, 5, 1, 2 \)) + ReLU + MaxPool(\(2, 2\))
    \item Conv(\( 3*w_e + \mathit{in}_2 \rightarrow 6 * w_e, 3, 1, 1 \)) + ReLU
    \item Conv(\( 6*w_e + \mathit{in}_2 \rightarrow 4 * w_e, 3, 1, 1 \)) + ReLU
    \item Conv(\( 4*w_e + \mathit{in}_2 \rightarrow w_e, 3, 1, 1 \)) + ReLU
    \item Conv(\( w_e + \mathit{in}_2 \rightarrow L, 1, 1, 0 \)) + ReLU + MaxPool(\(2, 2\))
\end{enumerate}
Here, \(\mathit{in}_1\) and \(\mathit{in}_2\) are skip connections to spatially reduced inputs that are computed directly on the original encoder input with an AvgPool(\(8, 8\)) and AvgPool(\(16, 16\)) layer, respectively. Finally, the output from the last convolution layer is spatially reduced to a size of \(1\) via an adaptive average pooling operation. This results in a latent space with \(L\) elements. This latent space is then decoded with the following decoder model based on convolution+ReLU blocks with Upsampling layers:
\begin{enumerate}
    \item Conv(\( L \rightarrow w_d, 1, 1, 0 \)) + ReLU + Up(\(4, 2, \mathit{nearest}\))
    \item Conv(\( w_d + L \rightarrow w_d, 3, 1, 1 \)) + ReLU + Up(\(2, 2, \mathit{nearest}\))
    \item Conv(\( w_d + L \rightarrow w_d, 3, 1, 1 \)) + ReLU + Up(\(2, 2, \mathit{nearest}\))
    \item Conv(\( w_d + L \rightarrow w_d, 3, 1, 1 \)) + ReLU + Up(\(2, 2, \mathit{nearest}\))
    \item Conv(\( w_d + L \rightarrow w_d, 3, 1, 1 \)) + ReLU + Up(\(2, 2, \mathit{nearest}\))
    \item Conv(\( w_d + L \rightarrow w_d, 3, 1, 1 \)) + ReLU + Up(\(2, 2, \mathit{bilinear}\))
    \item Conv(\( w_d + L \rightarrow w_d, 5, 1, 2 \)) + ReLU
    \item Conv(\( w_d + L \rightarrow w_d, 3, 1, 1 \)) + ReLU
    \item Conv(\( w_d \rightarrow \mathit{in}, 3, 1, 1 \))
\end{enumerate}
Here, the latent space is concatenated along the channel dimension and spatially expanded to match the corresponding spatial input size of each layer for the skip connections. In our implementation, an encoder width of \(w_e=32\), a decoder width of \(w_d=96\) with a latent space dimensionality of \(L=32\) worked best across experiments. For the model \mTf{Enc} on the experiments \dInc{} and \dTra{}, we employ \(L=31\) and concatenate the scalar simulation parameter that is used for conditioning, i.e., Reynolds number for \dInc{} and Mach number for \dTra{}, to every instance of the latent space. For \mTf{VAE} we proceed identically, but here every latent space element consists of two network weights for mean and variance via reparameterization as detailed by \textcite{kingma2014_Autoencoding}. For \mTf{MGN}, we use an additional first latent space of size \(L\) that contains a simulation parameter encoding via an MLP as proposed by \textcite{han2021_Predicting}. Compared to our improved approach, this means \mTf{MGN} is not capable to change this quantity over the course of the simulation.

For the latent processor in \mTf{MGN} we directly follow the original transformer specifications by \textcite{han2021_Predicting} via a single transformer decoder layer with four attention heads and a layer width of \(1024\). Latent predictions are learned as a residual from the previous step. For our adaptations \mTf{Enc} and \mTf{VAE}, we instead use a single transformer encoder layer and learn a full new latent state instead of a residual prediction.

To train the different transformer variants end-to-end, we always use a batch size of \(8\). We train each model with a training rollout of \(m=60\) steps (\(m=50\) for \dIso{}) using a transformer input window of \(k=30\) steps (\(k=25\) for \dIso{}). We first only optimize the encoder and decoder to obtain a reasonably stable latent space, and then the training rollout is linearly increased step by step as proposed by \textcite{han2021_Predicting}. We start increasing the rollout at epoch \(300\) (\(40\) for \dIso{}) until the full sequence length is reached at epoch \(1200\) (\(160\) for \dIso{}). Each model is trained with an MSE loss over the full sequence (adjusted to the current rollout length). On \dInc{} and \dTra{} these transformer-based models were trained for \( 5000 \) epochs, and on \dIso{} for \( 200 \) epochs.

We do not train the decoder to recover values inside the cylinder area for \dInc{} and \dTra{}, by applying a binary masking (also see \cref{app: incompressible flow} for details) before the training loss computation. Note that this masking is not suitable for autoregressive approaches in the input space, as the masking can cause a distribution shift via unexpected values in the masked area during inference, leading to instabilities. The pure reconstruction, i.e. the first step of the sequence that is not processed by the latent processor, receives a relative weight of \(1.0\), and all steps of the rollout \textit{jointly} receive a weight of \(1.0\) as well, to ensure that the model balances reconstruction and prediction quality. For \mTf{VAE}, an additional regularization via a Kullback–Leibler divergence on the latent space with a relative weight of \(0.1\) is used. As detailed by \textcite{kingma2014_Autoencoding}, for a given mean \(l_m^i\) and log variance \(l_v^i\) of each latent variable \(l^i\) with \(i \in 0, 1, \ldots, L \), the regularization \(\mathcal{L}_\mathit{KL}\) is computed as
\[ \mathcal{L}_\mathit{KL} = -0.5 * \frac{ 1 }{ L } * \sum_{i=0}^{L} 1 + l_v^i - {l_m^i}^2 - e^{l_v^i}. \]

\begin{table}[th]
    \caption{Overview of training and inference performance for different model architectures.}
    \label{tab-app: performance}
    \centering
    \footnotesize

    \begin{tabular}{l c c c c c c}
        \toprule

        Architecture & Training & Batch & Training & Training Speed & Inference & Inference \\
        & Epochs & Size & Time [h] & per Epoch [min] & Speed [s] & Speed [s] \\
        & \dInc{} / \dTra{} / \dIso{} &  & \dInc{} / \dTra{} / \dIso{} & \dInc{} / \dTra{} / \dIso{} & without I/O & with I/O\\
        \midrule
        \mACDM{R20} & \multirow{2}*{3100 / 3100 / 100} & \multirow{2}*{64} & \multirow{2}*{66 / 42 / 62} & \multirow{2}*{0.80 / 0.74 / 37.4} & 193.9 & 195.7\\
        \mACDM{R100} & & & & & 973.2 & 975.0 \\
        \midrule
        \mUnet{} & \multirow{4}*{1000 / 1000 / 100} & 64 &  26 / 19 / 90  & 1.16 / 1.10 / 55.0  & \multirow{4}*{9.4} & \multirow{4}*{11.1}\\
        \mUnet{m4} & & 32 & 34 / 31 / 156 & 1.97 / 1.82 / 95.3 & \\
        \mUnet{m8} & & 16 & 45 / 43 / 217 & 2.67 / 2.49 / 130.8 & \\
        \mUnet{m16} & & 8 & 54 / 51 / 262 & 3.20 / 3.03 / 161.1 & \\
        \midrule
        \mTf{MGN} & \multirow{3}*{5000 / 5000 / 200} & \multirow{3}*{8} & 43 / 42 / 69 & 0.51 / 0.48 / 20.8 & 0.8 & 2.8\\
        \mTf{Enc} & & & 38 / 38 / 67 & 0.42 / 0.43 / 19.9 & 0.6 & 2.8 \\
        \mTf{VAE} & & & 38 / 38 / 68 & 0.42 / 0.42 / 20.0 & 0.7 & 2.7 \\
        \midrule
        \mResNet{dil.} & \multirow{2}*{1000 / 1000 / 100} & \multirow{2}*{64} & \multirow{2}*{52 / 49 / 263} & \multirow{2}*{3.15 / 2.95 / 152.5} & \multirow{2}*{4.2} & \multirow{2}*{6.0} \\
        \mResNet{} & & & & & & \\
        \midrule
        \mFNO{16} & \multirow{2}*{2000 / 2000 / 200} & \multirow{2}*{64} & 13 / 12 / 55 & 0.36 / 0.34 / 16.5 & 2.3 & 4.1 \\
        \mFNO{32} & & & 8 / 8 / 33 & 0.22 / 0.21 / 9.9 & 2.5 & 4.2\\
        \midrule
        \mRefiner{R2} & \multirow{3}*{3000 / 3000 / 100} & \multirow{3}*{64} & \multirow{3}*{59 / 55 / 61} & \multirow{3}*{1.18 / 1.10 / 37.2} & 30.5 & 32.8\\
        \mRefiner{R4} & & & & & 60.7 & 63.2 \\
        \mRefiner{R8} & & & & & 92.7 & 94.4 \\

        \bottomrule
    \end{tabular}
\end{table}

\section{Training and Inference Performance} \label{app: performance}
All model architectures were trained, evaluated, and benchmarked on a server with an NVIDIA RTX A5000 GPU with 24GB of video memory and an Intel Xeon Gold 6242R CPU with 20 cores at 3.1 GHz. A performance overview across models can be found in \cref{tab-app: performance}. The training time column indicates how many hours are approximately required to fully train a single model according to the epochs and batch size given further left. For each architecture, we train 3 models (2 for \dIso{}) based on randomly seeded runs for the evaluations in the paper, and report the maximum training time. The training speed per epoch in minutes in the next column is averaged over a set of training epochs across all trained models.

All model architectures were trained on each data set until their training loss curves were visually fully converged. This means, architectures with more complex learning objectives require more epochs compared to simpler methods. As such, the transformer variants are highly demanding, as they first need to learn a good latent embedding via the encoder and decoder, and afterwards need to learn the transformer unrolling schedule that is faded in during training time. \mACDM{} which needs to learn a full denoising schedule via random sampling can also require more training iterations compared to direct next-step predictors such as \mUnet{}, \mResNet{}, or \mFNO{}. Furthermore, we found the performance of next-step predictors to degrade when trained substantially past the point of visual convergence in early exploration runs. As mentioned above, the default training batch size of \( 64 \) is reduced for architectures that exceed available GPU memory, so the training time comparison is performed at roughly equal memory. Thus, training unrolled \mUnet{} models is highly expensive, especially on \dIso{}, both via higher memory requirements that result in a lower batch size, but also in the number of computations required for the training rollout.

The right side of \cref{tab-app: performance} features the inference speed of each method. It is measured on a single example sequence consisting of \( T=1000 \) time steps. We report the overall time in seconds that each architecture required during inference for this sequence. Shown in the table is the pure model inference time, as well as the performance including I/O operations and data transfers from CPU to GPU. Note that compared to the performance of \mUnet{}, the inference speed slowdown factor of the diffusion-based architecture \mRefiner{} and \mACDM{} is closely related to the number of refinement or diffusion steps \( R \). This directly corresponds to the number of backbone model evaluations.

\section{Accuracy Evaluation with LSiM} \label{app: lsim}
The LSiM metric \parencite{kohl2020_Learning} is a deep learning-based similarity measure for data from numerical simulations. It is designed to more accurately capture the similarity behavior of larger patterns or connected structures that are neglected by the element-wise nature of point-based metrics like MSE. As a simple example, consider a vortex inside a fluid flow that is structurally correctly predicted, but spatially misplaced compared to a reference simulation. While MSE would result in a large distance value, LSiM results in a relatively low distance, especially compared to another vortex that is spatially correctly positioned, but structurally different. LSiM works by embedding both inputs that should be compared in a latent space of a feature extractor network, computing an element-wise difference, and aggregating this difference to a scalar distance value via different operations. The metric is trained on a range of data sets consisting of different transport-based PDE simulations like advection-diffusion equations, Burgers' equation, or the full Navier-Stokes equations. It has been shown to generalize well to flow simulation data outside its training domain like isotropic turbulence.

\section{Evaluating Temporal Coherence} \label{app: coherence}

\begin{figure}[th]
    \centering
    \includegraphics[height=12.3cm]{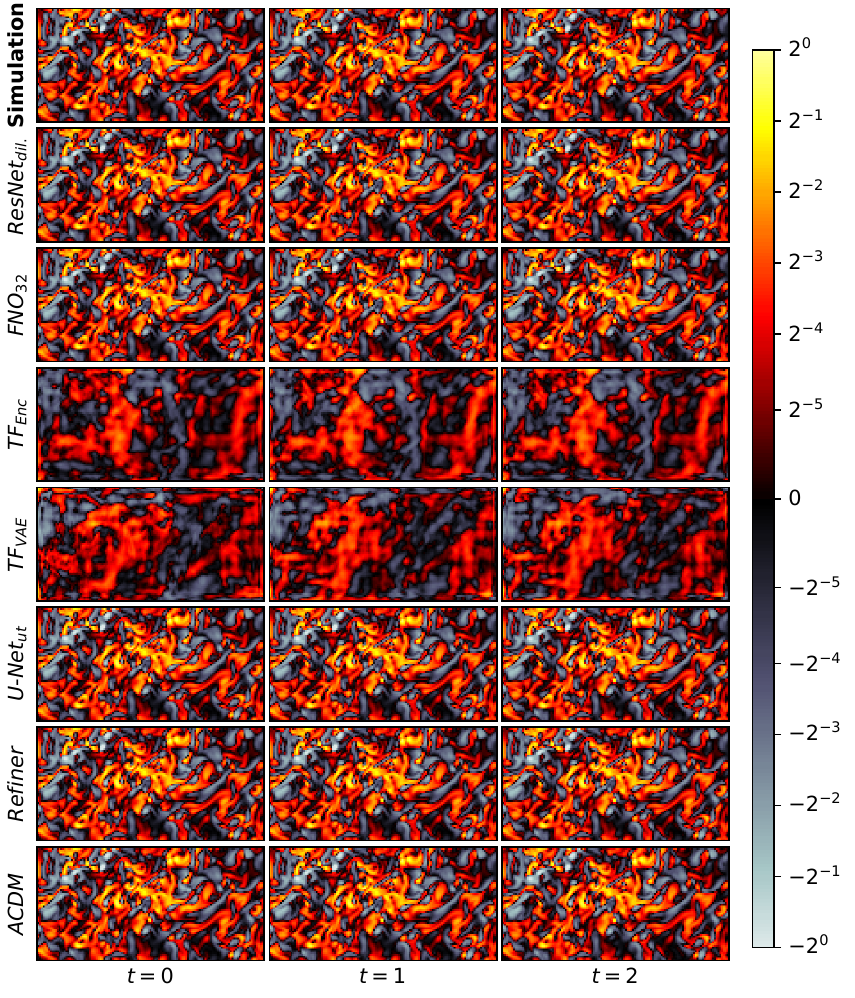}
    \hfill
    \includegraphics[height=12.3cm]{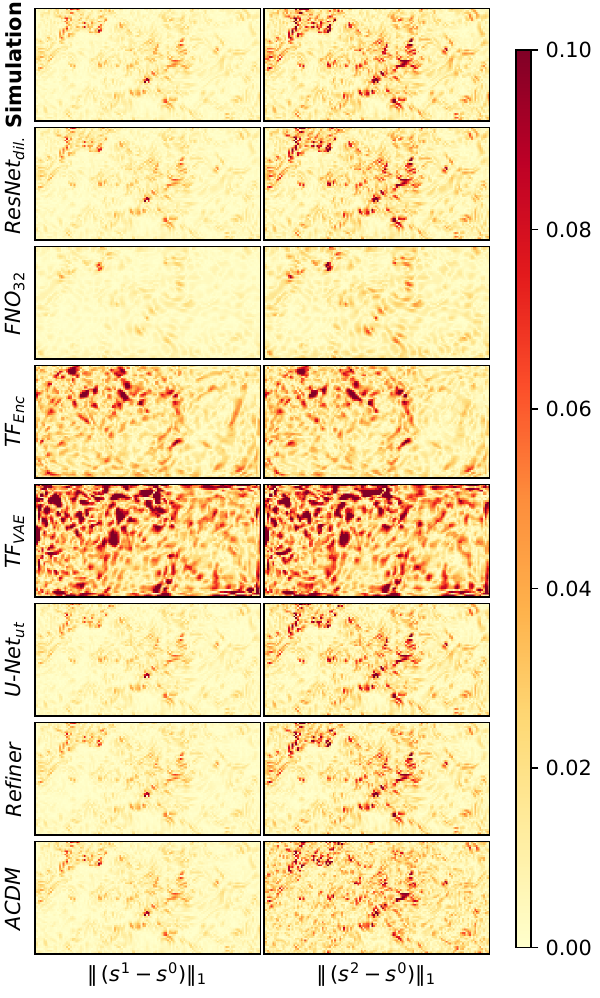}

    \caption{Temporal coherence analysis of different model architectures. Model predictions (left) and the differences between the first and following two prediction steps (right) are shown an example from \dIso{} with \( z=300 \).}
    \label{fig-app: coherence}
\end{figure}

Here, we analyze the temporal coherence between individual time steps of the architectures under consideration. In \cref{fig-app: coherence} on the left, we display the first three simulation steps of a sequence from \dIso{}, along with the corresponding predictions of models from each architecture class. In addition, the spatial changes between the first two predicted steps and \(s^0\) are visualized on the right. Most architectures do not exhibit issues with temporal coherence and closely follow to difference pattern produced by the simulation, except from the transformer variants: both, \mTf{Enc} or \mTf{VAE}, struggle to reproduce the original vorticity field at \(t=0\), and furthermore show large difference between prediction steps. The key difference for these architectures is, that the decoders of \mTf{Enc} or \mTf{VAE} do not have access to previously generated time steps, as their input is only a sample from the latent space at every step. This leads to temporal artifacts where large differences between consecutive time steps can occur. Note that this problem is in general worse for \mTf{VAE} indicated by the larger overall difference magnitude, as the probabilistic nature of the model makes temporal coherence even more challenging. \mACDM{} with \(R=100\) does show some coherence issues in the vorticity, indicated by some very small, slightly darker areas in the difference field, especially for \(s^2\) (see \href{https://ge.in.tum.de/publications/2023-acdm-kohl/}{accompanying videos}). However, they are quite minor, only visible in the vorticity and not in the raw predicted velocities, and can be further mitigated with additional diffusion steps.

\section{Temporal Stability on Extremely Long Rollouts} \label{app: temporal stability}

\begin{figure}[ht]
    \centering
    \includegraphics[width=0.99\textwidth]{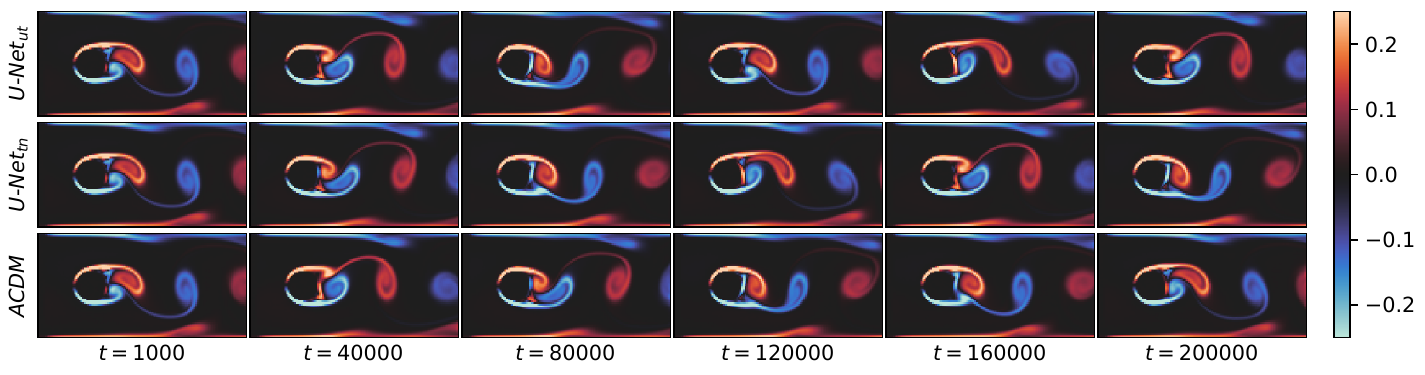}\\
    \vspace{0.1cm}
    \includegraphics[width=0.99\textwidth]{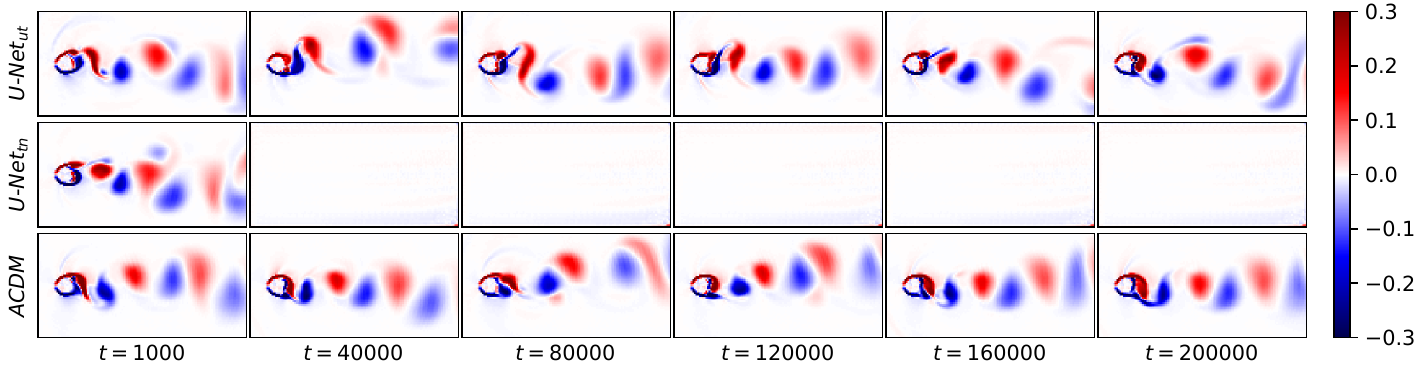}

    \caption{Predictions (vorticity) from the most temporally stable architectures, when unrolled for \( T = 200\,000 \) steps on a sequence from \dIncHigh{} with \( \mathit{Re} = 1000 \) (top), and on a sequence from \dTraExt{} with \( \mathit{Ma} = 0.52 \) (bottom).}
    \label{fig-app: prediction long}
\end{figure}

\begin{figure}[hb]
    \centering
    \includegraphics[width=0.5\textwidth]{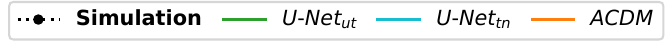}\\
    \includegraphics[width=0.49\textwidth]{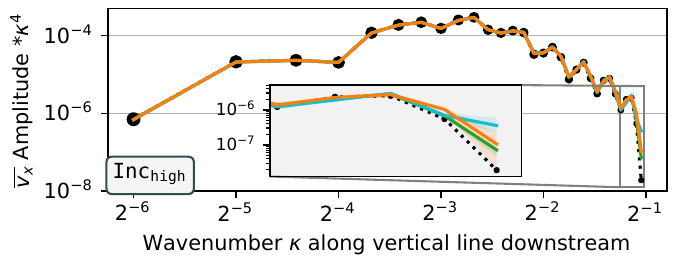}
    \hfill
    \includegraphics[width=0.49\textwidth]{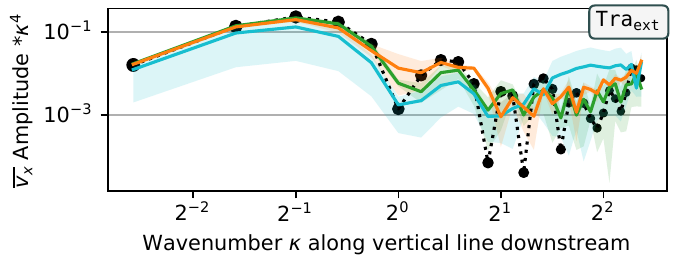}

    \caption{Spatial frequency along a vertical line one cylinder diameter downstream for a sequence from \dIncHigh{} with \( \mathit{Re} = 1000 \) (left), and on a sequence from \dTraExt{} with \( \mathit{Ma} = 0.50 \) (right). The prediction spectra are computed on the mean flow achieved from averaging every \nth{100} step from a prediction with a horizon of \( T = 200\,000 \) steps.}
    \label{fig-app: frequency long}
\end{figure}

To further investigate the temporal stability of the most stable methods \mUnet{ut}, \mUnet{tn}, and \mACDM{} in our comparison, we provide results on extremely long inference rollouts. We choose sequences from the extrapolation areas of our data sets for this purpose: three sequences from \dIncHigh{} with \( \mathit{Re} \in \{960, 980, 1000\} \) and three from \dTraExt{} with \( \mathit{Ma} \in \{0.50, 0.51, 0.52\} \). Every architecture is unrolled over \( T = 200\,000 \) steps, for the three training runs each. \Cref{fig-app: prediction long} visualizes the resulting predictions. All methods and training runs remain fully stable over the entire rollout on the sequence from \dIncHigh{}. Since this case is unsteady but fully periodic, the results of all models is a simple, periodic trajectory that prevents error accumulation. For the sequences from \dTraExt{}, one from the three trained \mUnet{tn} models has stability issues within the first few thousand steps and deteriorates to a simple, mean flow prediction without vortices. The remaining training runs do however remain stable. \mUnet{ut} and \mACDM{} on the other hand are fully stable across training runs for this case, indicating a fundamentally higher resistance to rollout errors which eventually could cause instabilities. Due to its highly chaotic nature, relying on simple, periodic predictions is not sufficient for this case: As displayed in \cref{fig-app: prediction long}, the predictions of \mUnet{ut}, but especially \mACDM{}, exhibit differences at the same stage of the vortex shedding period. This is most clearly visible when comparing the predictions for \( t = 120\,000 \), \( t = 160\,000 \), and \( t = 200\,000 \), where a blue vortex facing downwards is about to detach behind the cylinder.

In \cref{fig-app: frequency long}, the statistical match of these long-term predictions to the physical behavior of the simulation trajectories is analyzed. We evaluate the spatial frequency spectrum of the mean horizontal flow velocity along a vertical line one cylinder diameter downstream of the cylinder. For the predictions, every \nth{100} step of the predicted rollout from every training run is used to compute the mean flow. Note that the corresponding simulation spectrum is only computed over the simulated time range of \( t \in [300,1300) \) for \dIncHigh{}, and \( t \in [0,1000) \) for \dTraExt{} with temporal strides of \( 2 \) leading to \( T=500 \) time steps for this evaluation. For the rather simple sequence from \dIncHigh{}, all methods perfectly match the simulation spectrum. For the sequence from \dTraExt{}, the spectrum from \mUnet{tn} has a high standard deviation across the frequency band due the diverging training run. \mUnet{ut} and \mACDM{} statistically match the low simulation frequencies here very well, and only exhibit minor deviations for the higher frequencies, indicating that the predictions do not drift substantially over extremely long rollout horizons.

\subsection{Stability Criteria for Unrolled Training}
\begin{figure}[hp]
    \centering
    \includegraphics[width=0.99\textwidth]{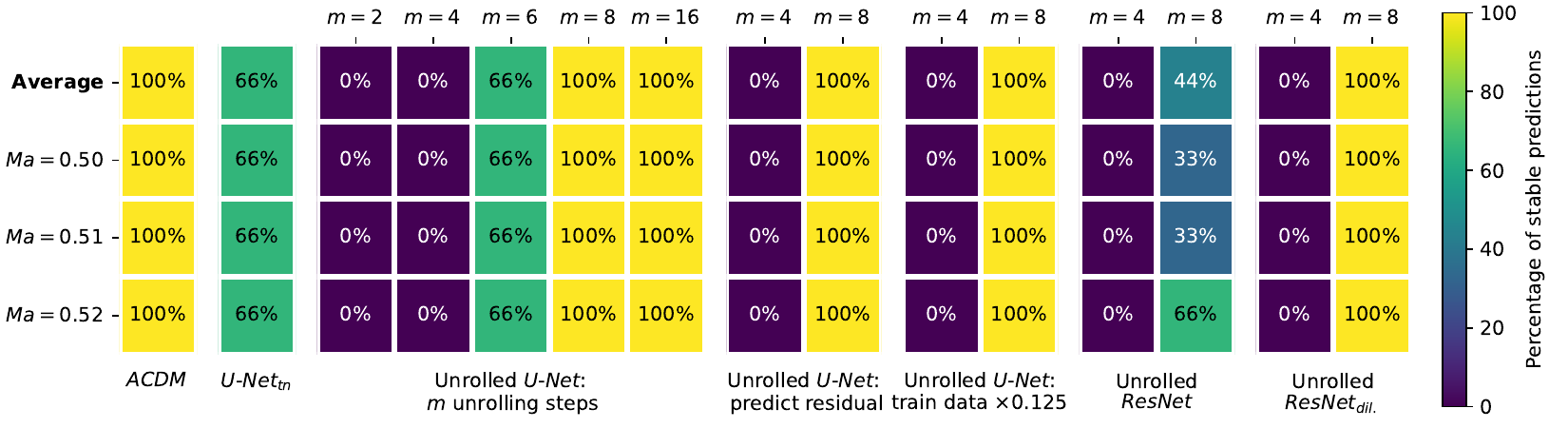}

    \caption{Important aspects to achieve stable models for extremely long rollouts on \dTraExt{}. Shown is the percentage of stable runs for three model seeds across sequences with \( \mathit{Ma} \in \{0.50, 0.51, 0.52\} \) and their average. While \mACDM{} is fully stable out-of-the-box, one \mUnet{tn} model diverges across Mach numbers. The most important aspect for fully stable unrolled \mUnet{} models is the rollout length \(m\), as displayed in the third block. However, the training methodology of predicting residuals instead of full next states did not impact the stability. Similarly, reducing the training data by a factor of \(8\) (with an \(8 \times\) longer training) did not alter the results across values of \(m\). Other architectures like \mResNet{} and \mResNet{dil.} are also capable to achieve full stability when trained with unrolling, even though \mResNet{} requires slightly larger \(m\) for consistent stability.}
    \label{fig-app: stability long heatmap}
\end{figure}

For the \mUnet{} models with unrolled training we also investigated key criteria to achieve fully stable rollouts over extremely long horizons. For this purpose, different ablation architectures are evaluated on the long rollout experiments over \( T = 200\,000 \) rollout steps described above. \Cref{fig-app: stability long heatmap} displays the percentage of stable runs across architectures for three trained models for every sequence from \dTraExt{}, as well as an average stability. As shown above \mACDM{} remains fully stable while only two out of three \mUnet{tn} are stable. The most important stability criterion for \mUnet{} trained with unrolling is the number of unrolling steps \(m\): while models with \(m \leq 4\) do not achieve stable rollouts, using \(m\geq8\) is sufficient for stability across different Mach numbers.

Three factors that did not substantially impact rollout stability in our experiments are 
\begin{enumerate*}[label=(\textbf{\textit{\roman*}})]
    \item the prediction strategy,
    \item the amount of training data, and
    \item the backbone architecture.
\end{enumerate*}
First, using residual predictions, i.e., \( s^t = s^{t-1} + f_\theta(s^{t-1})\) instead of \( s^t =  f_\theta(s^{t-1})\) does not impact stability for different values of \(m\) as shown in the middle of \cref{fig-app: stability long heatmap}. Second, the stability is not affected when reducing the amount of available training data by a factor of \(8\) from \(1000\) steps per Mach number to \(125\) steps. Note that these models are also trained with \(8\times\) more epochs to ensure a fair comparison. This training data reduction still retains the full physical behavior, i.e., complete vortex shedding periods. Third, it possible to train other backbone architectures with unrolling to achieve fully stable rollouts, as shown on the right in \cref{fig-app: stability long heatmap}: \mResNet{dil.} models trained with \(m = 8\) are also able to keep predictions stable across the entire prediction horizon. For \mResNet{} only one trained model is stable, most likely due to the reduced receptive field. However, we expect achieving full stability is also possible with longer training rollout horizons.

\begin{figure}[hp]
    \centering
    \includegraphics[width=0.49\textwidth]{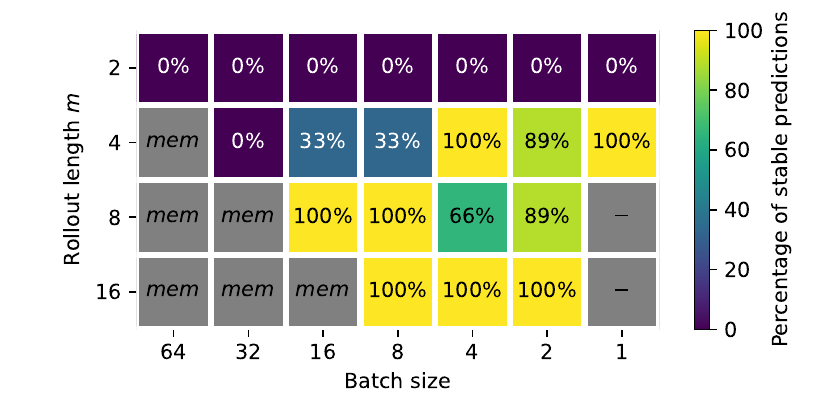}
    \hfill
    \includegraphics[width=0.49\textwidth]{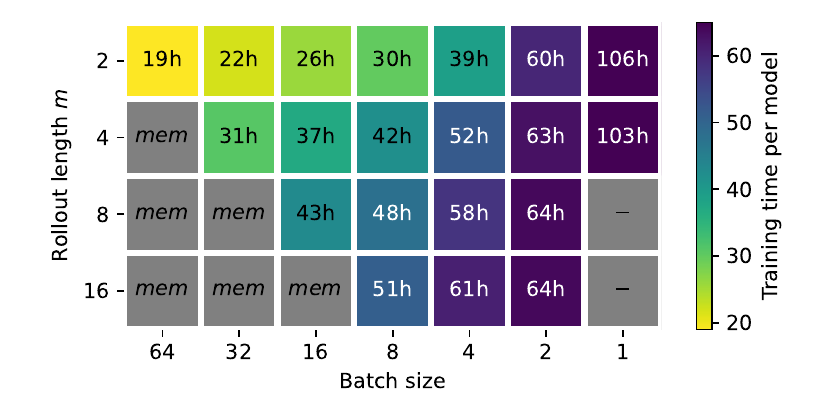}

    \caption{Stability investigation for unrolled \mUnet{} models with different combinations of batch size and rollout length \(m\) on \dTraExt{}. Shown are the percentage of stable predictions across three models and three sequences with \( T = 200\,000 \) steps each (left) and approximate training time for each parameter combination (right). Grey configurations are infeasible due to memory constraints (\textit{mem}) or omitted due to high computational resource demands (--). Notice that decreases in batch size can lead to more stable models for medium \(m\), but incur a higher training cost compared to increasing \(m\).}
    \label{fig-app: stability long heatmap tra}
\end{figure}

\begin{figure}[hp]
    \centering
    \includegraphics[width=0.49\textwidth]{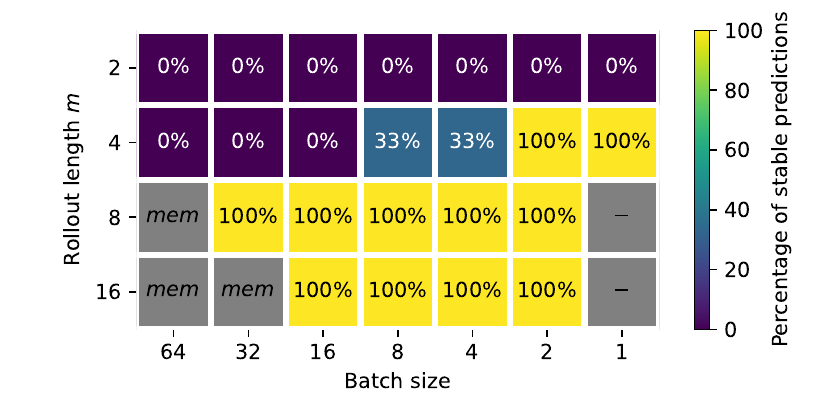}
    \hfill
    \includegraphics[width=0.49\textwidth]{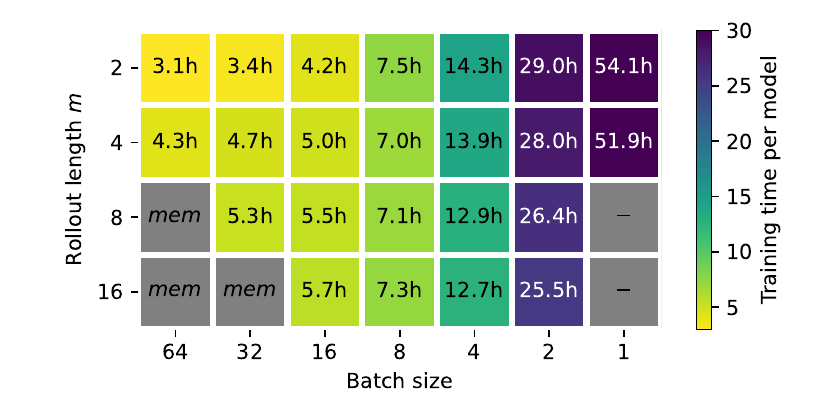}

    \caption{Stability investigation for unrolled \mUnet{} models (with a width reduced by a factor of \(8\)) with different combinations of rollout length \(m\) and batch size on \dIncHigh{}. Shown are the percentage of stable predictions across three sequences with \( T = 200\,000 \) steps and three models each (left) and approximate training time for each configurations (right). Grey parameter combinations are infeasible due to memory constraints (\textit{mem}) or omitted due to high computational resource demands (--). Similar to \cref{fig-app: stability long heatmap tra} decreasing the batch size can lead to higher stability models for medium \(m\), but this affects training cost. Note that there is barely any overhead for additional unrolling steps, as the model is small and longer training rollouts are implemented via fewer training sequences.}
    \label{fig-app: stability long heatmap inc}
\end{figure}

Finally, we observed that the batch size can impact the stability of models trained with unrolling. In the image domain, it has been documented that smaller batch sizes exhibit better generalization properties compared to larger mini-batch sizes or even full gradient decent without mini-batches \parencite{goodfellow2016_Deep}. As described in more detail in \cref{app: unet implementation}, we adjust the batch size of \mUnet{} for different values of \(m\), such that each batch contains the same amount of data, e.g., halving the batch size when doubling \(m\). By default, we chose the largest possible batch size that fits in GPU memory in the remainder of this work, to ensure computational efficiency. Here, we investigate different configurations by varying of batch size and rollout length on \dTraExt{} and \dIncHigh{} in \cref{fig-app: stability long heatmap tra,fig-app: stability long heatmap inc}, respectively. Training models with smaller batches for the same amount of network updates did not improve stability, so all networks are trained with the same amount of data, i.e., an equal number of epochs with more network updates for small batch sizes. Note that the width of the \mUnet{} architecture was substantially reduced by a factor of \(8\) across all network layers for \dIncHigh{} in \cref{fig-app: stability long heatmap inc} to artificially increase the difficulty of the learning task, as otherwise every model configuration would be fully stable.

\begin{figure}[th]
    \centering
    \includegraphics[width=0.49\textwidth]{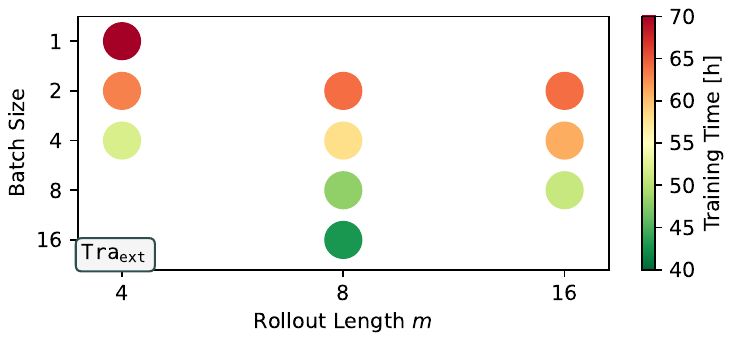}
    \hfill
    \includegraphics[width=0.49\textwidth]{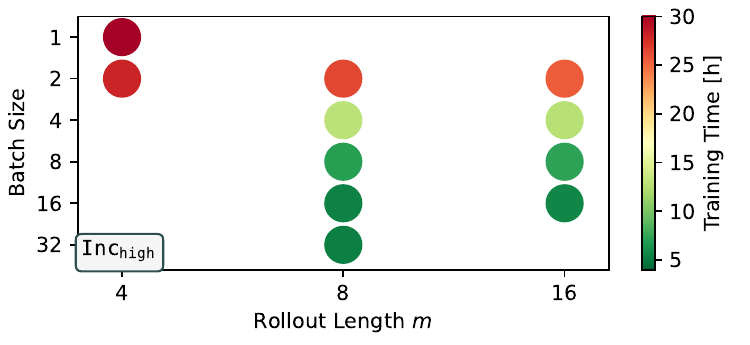}

    \caption{Training time for different combinations of rollout length \(m\) and batch size on \dTraExt{} (left) and \dIncHigh{} (right). Only configurations that result in highly stable rollouts are shown (percentage of stable runs across three trained models and three sequences \(\geq 89\%\)). The number of channels across the \mUnet{} architecture is reduced by a factor of \(8\) for the  \dIncHigh{} case, as all configurations were already fully stable for the full-size \mUnet{}. Note that the training time increases faster for smaller batches compared to longer rollouts.}
    \label{fig-app: stability long scatter}
\end{figure}

For both test sets, \mUnet{m4} with the largest batch size that fits in memory is not stable, while perfect stability can be achieved with lowering the batch size. However, this effect can not fully replace the stabilization from longer training rollouts, as \mUnet{m2} is never stable. Furthermore, smaller batch sizes are less memory efficient meaning model training takes longer. Since we implement longer training rollouts via fewer training sequences, larger values of \(m\) do not necessarily induce higher training times as shown for the small \mUnet{} on \dIncHigh{} in \cref{fig-app: stability long heatmap inc}. Nevertheless, larger training rollouts did lead to longer training times for the full-size \mUnet{} model as shown in \cref{fig-app: stability long heatmap tra} and also listed in \cref{tab-app: performance}. Thus, we investigate the most efficient combination of rollout length and batch size to achieve fully stable rollouts. As shown in \cref{fig-app: stability long scatter}, choosing longer rollouts over smaller batches leads to high stability at lower training cost, for both \dIncHigh{} and \dTraExt{}. However, the difference did change depending on the model size in our experiments. 

To summarize, the most important criterion for full stable models is the training rollout length. While lowering the batch size did have an effect, especially in situations where the rollout is slightly smaller than required, choosing large batch sizes was superior in terms of training time. Model architecture, larger amounts of training data without a substantial change of physical behavior, as well as the alternate strategy of predicting residuals did not have a measurable impact on the model stability in our experiments.

\clearpage

\section{Ablations} \label{app: ablations}
In the following, we provide various ablation studies on the number of diffusion steps in \cref{app: ablation diffusion steps}, as well as ablations on the stabilization techniques of longer training rollouts in \cref{app: ablation training rollout} and training noise in \cref{app: ablation training noise}. We investigate different loss formulations in \cref{app: ablation lsim training}, and analyze the impact of the recently proposed architecture modernizations for U-Nets in \cref{app: ablation unet modernizations}. Finally, we provide ablations on the PDE-Refiner method \parencite{lippe2023_PDERefiner} in \cref{app: ablation refiner}.

\subsection{Ablation on Diffusion Steps} \label{app: ablation diffusion steps}
\begin{figure}[p]
    \centering
    \includegraphics[width=0.99\textwidth]{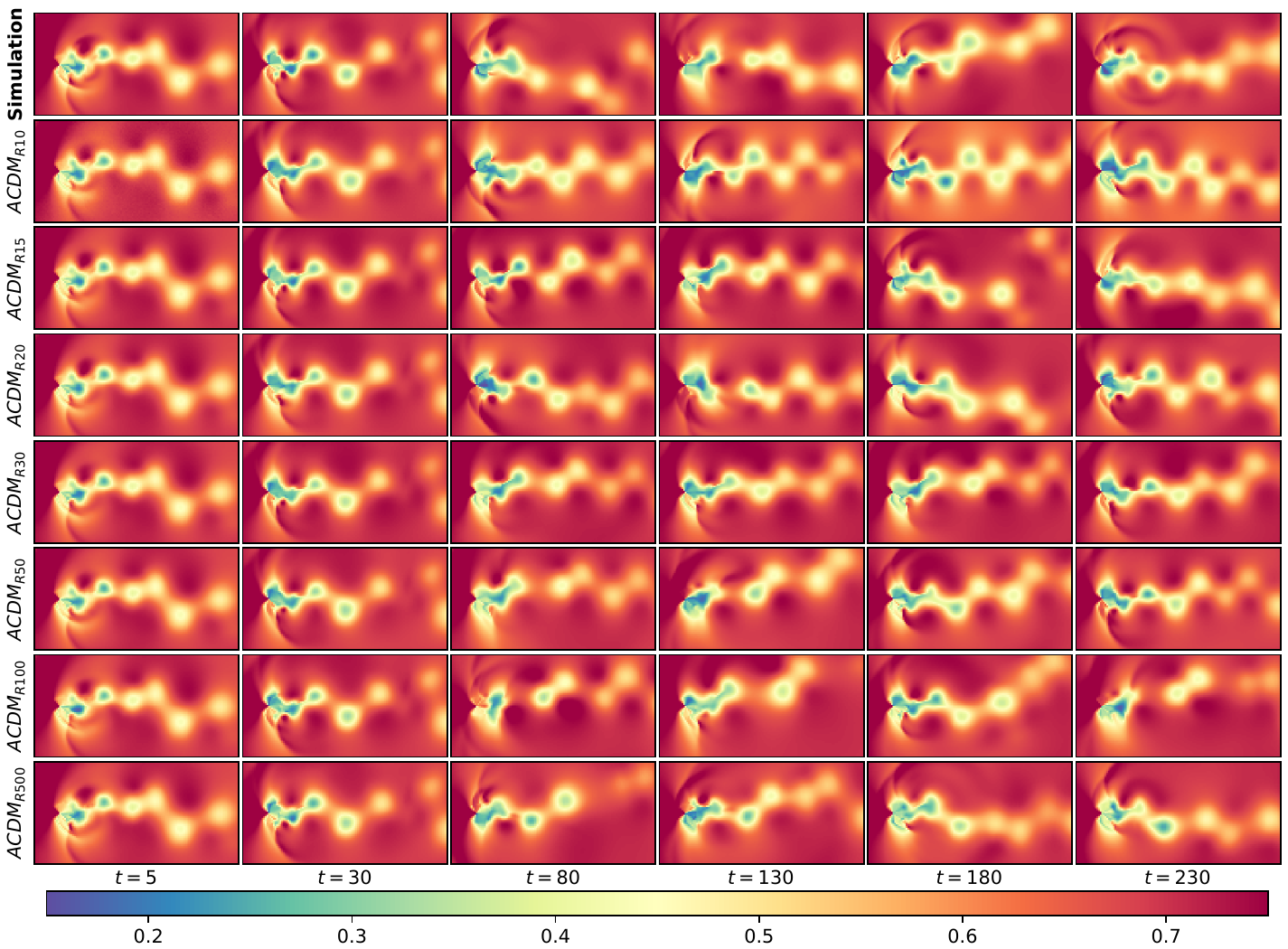}

    \caption{Diffusion Step Ablation (see \cref{app: ablation diffusion steps}): Pressure predictions from \dTraLong{}.}
    \label{fig-app: example tra diffusion steps}
\end{figure}

\begin{figure}[p]
    \centering
    \includegraphics[width=0.99\textwidth]{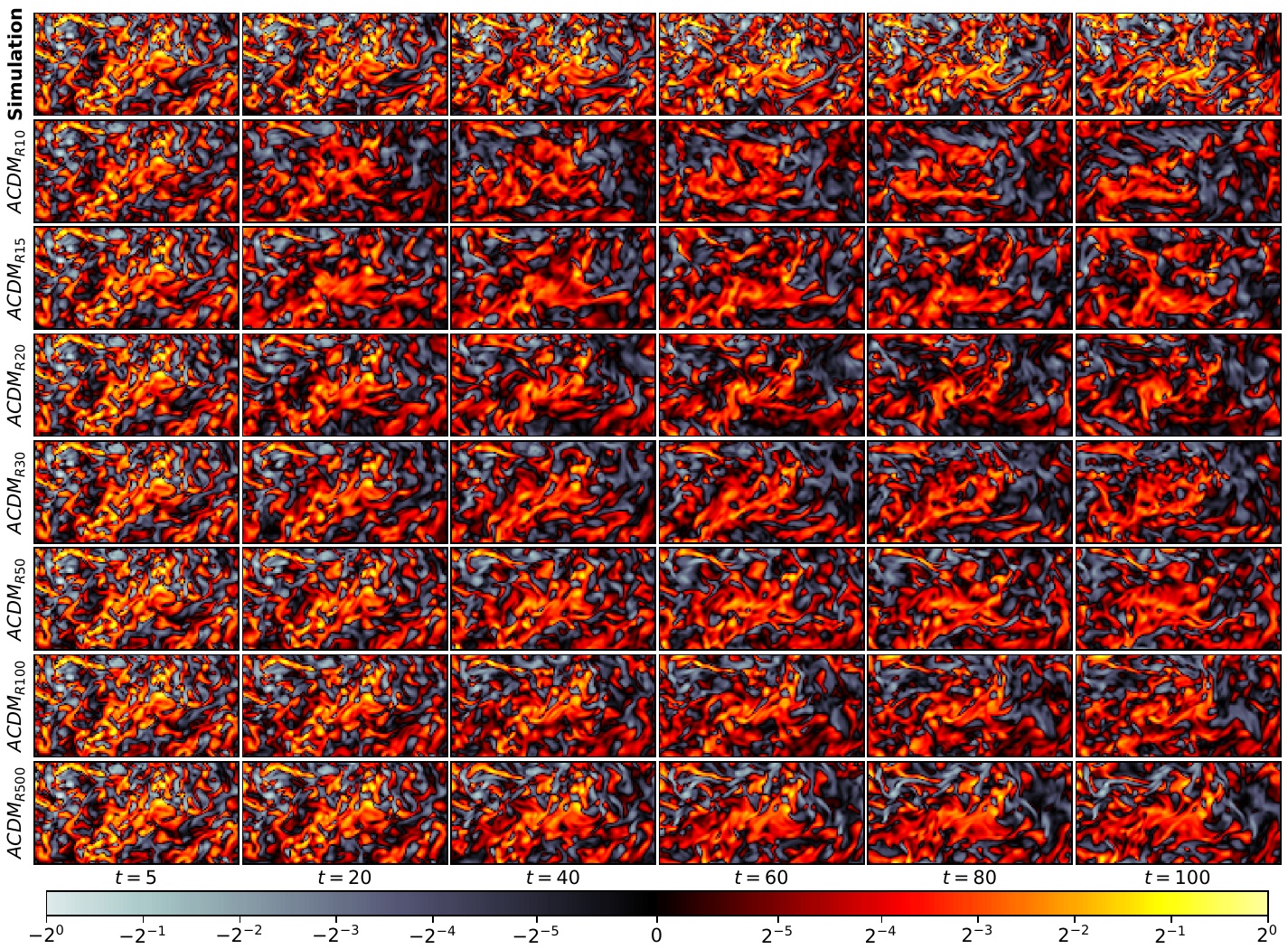}

    \caption{Diffusion Step Ablation (see \cref{app: ablation diffusion steps}): Vorticity predictions from \dIso{}.}
    \label{fig-app: example iso diffusion steps}

\end{figure}

In the following, we will investigate the \mACDM{} approach with respect to the effect of the number of diffusion steps \(R\) in each autoregressive prediction step. We use the adjusted linear variance schedule as discussed in \cref{app: acdm implementation}, according to the investigated diffusion step \(R\). At training and inference time, models always use \(R\) diffusion steps. Prediction examples for this evaluation can be found in \cref{fig-app: example tra diffusion steps,fig-app: example iso diffusion steps} in \cref{app: prediction examples ablation}.

\begin{table}[th]
    \caption{Accuracy ablation for different diffusion steps \(R\).}
    \label{tab-app: accuracy diffusion steps}
    \centering
    \small

    \begin{tabular}{l c c c c c c c}
        \toprule
        & & \multicolumn{2}{c}{\dTraExt{}{}} & \multicolumn{2}{c}{\dTraInt{}} & \multicolumn{2}{c}{\dIso{}}\\
        \cmidrule(lr){3-6} \cmidrule(lr){7-8}
        & & MSE & LSiM & MSE & LSiM & MSE & LSiM\\

        \textbf{Method} & \(R\)
        & (\(10^{-3}\)) & (\(10^{-1}\)) & (\(10^{-3}\)) & (\(10^{-1}\))
        & (\(10^{-2}\)) & (\(10^{-1}\))\\
        \midrule

        \mACDM{} & 10
        & $3.8\pm1.4$ & $1.8\pm0.3$  &  $6.2\pm2.5$ & $2.1\pm0.6$
        & $15.1\pm7.4$ & $6.6\pm1.4$\\
    
        \mACDM{} & 15
        & $2.5\pm1.5$ & $1.4\pm0.3$  &  $2.7\pm2.0$ & $1.4\pm0.5$
        & $4.8\pm1.6$ & $4.3\pm1.0$\\
        
        \mACDM{} & 20
        & $2.3\pm1.4$ & $1.3\pm0.3$  &  $2.7\pm2.1$ & $1.3\pm0.6$
        & $4.5\pm1.3$ & $4.1\pm0.8$\\
        
        \mACDM{} & 30
        & $2.5\pm1.9$ & $1.4\pm0.4$  &  $2.7\pm2.3$ & $1.3\pm0.6$
        & $4.8\pm1.9$ & $4.1\pm0.9$\\

        \mACDM{} & 50
        & $2.3\pm1.4$ & $1.3\pm0.3$  &  $2.4\pm2.1$ & $1.3\pm0.6$
        & $3.4\pm0.9$ & $3.4\pm0.7$\\

        \mACDM{} & 100
        & $2.3\pm1.3$ & $1.3\pm0.3$  &  $3.1\pm2.7$ & $1.4\pm0.6$
        & $3.7\pm0.8$ & $3.3\pm0.7$ \\

        \mACDM{} & 500
        & $2.5\pm1.5$ & $1.4\pm0.4$  &  $3.1\pm2.5$ & $1.4\pm0.6$
        & $3.5\pm0.9$ & $3.2\pm0.7$\\

        \bottomrule
    \end{tabular}
\end{table}

\paragraph{Accuracy}
\cref{tab-app: accuracy diffusion steps} contains the accuracy, of \mACDM{} models with a different number of diffusion steps \(R\). While too few diffusion steps on \dTra{} are detrimental, as visible for \mACDM{R10}, adding more steps after around \(R=20\) does not improve accuracy. However, on \dIso{} the accuracy of \mACDM{} does continue to improve slightly with increased values of \(R\) up to our evaluation limit of \(R=500\). We believe this results from the highly underdetermined setting of the \dIso{} experiment. Note that there is a relatively sharp boundary between too few and a sufficient number of steps; in our experiments \(15-20\) steps on \dTra{} and \(50-100\) steps on \dIso{}.

\begin{figure}[ht]
    \centering
    \includegraphics[width=0.5\textwidth]{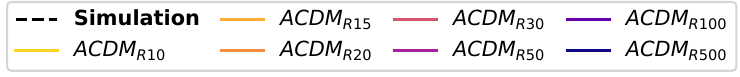}\\
    \includegraphics[width=0.49\textwidth]{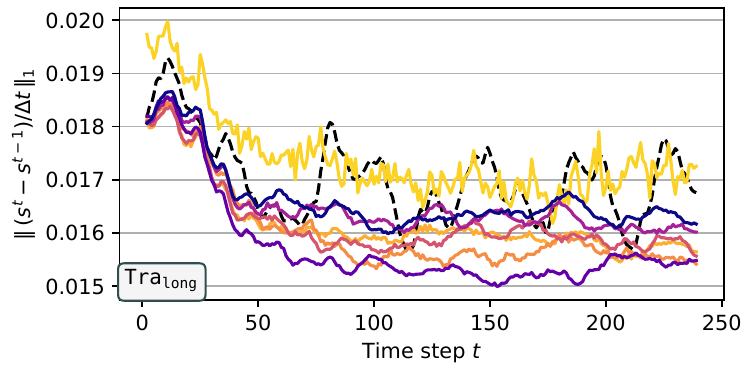}
    \hfill
    \includegraphics[width=0.49\textwidth]{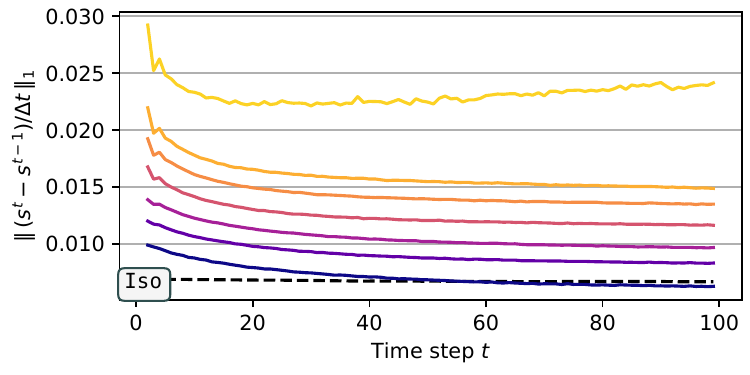}
    
    \caption{Temporal stability evaluation via error to previous time step for different diffusion steps \(R\) on \dTraLong{} (left) and \dIso{} (right). Standard deviations are omitted for visual clarity.}
    \label{fig-app: stability diffusion steps}
\end{figure}

\paragraph{Temporal Stability}
In \cref{fig-app: stability diffusion steps}, we evaluate the temporal stability via the magnitude of the rate of change of \(s\), as detailed above. Here, different behavior for the ablation models with respect to the number of diffusion steps emerges on \dTraLong{} and \dIso{}. For the former, too little steps, i.e., for \mACDM{R10}, result in unwanted, high-frequency temporal spikes that are also visible as slightly noisy predictions. For \(15-20\) diffusion steps, these issues vanished, and adding further iterations does not substantially improve temporal stability. Only a slightly higher rate of change can be observed for \mACDM{R50} and \mACDM{R500}.

\begin{figure}[ht]
    \centering
    \includegraphics[width=0.49\textwidth]{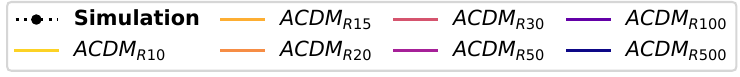}\\
    \includegraphics[width=0.49\textwidth]{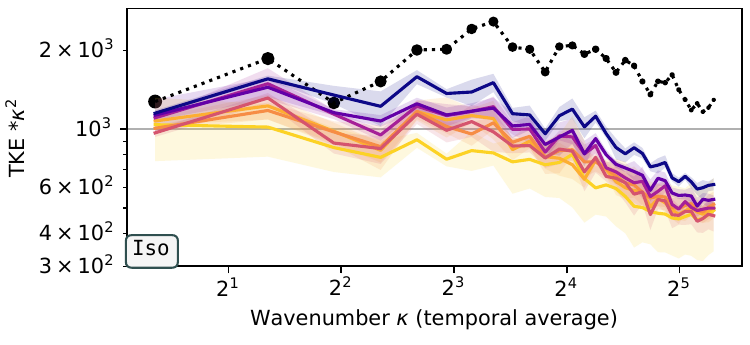}

    \caption{Spatial frequency analysis via the turbulent kinetic energy (TKE) on a sequence from \dIso{} with \( z = 300 \) for the diffusion step ablation.}
    \label{fig-app: frequency diffusion steps}
\end{figure}

On \dIso{}, a tradeoff between prediction accuracy and sampling speed occurs. Even though there are some minor temporal inconsistencies in the first few time steps for very low \(R\), all variants result in a stable prediction. However, the magnitude of the rate of change consistently matches the reference trajectory more closely when increasing \(R\). This also corresponds to a slight reduction in the overly diffusive prediction behavior for large \(R\), both visually and in a spatial spectral analysis via the TKE, as shown in \cref{fig-app: frequency diffusion steps}. We believe this tradeoff is caused by the highly underdetermined nature of the \dIso{} experiment, that leads to a weaker conditioned learning setting, that naturally requires more diffusion steps for high-quality results. Furthermore, the predictions of \mACDM{R100} exhibit minor visually visible temporal coherence issues on \dIso{}, where small-scale details can flicker quickly. This is caused by highly underdetermined nature of \dIso{}, and can be mitigate by more diffusion steps as well, as \mACDM{R500} reduces this behavior.

\paragraph{Summary}
\mACDM{} works well out-of-the-box with a large number of diffusion steps, but \(R\) can be used to balance accuracy and inference performance. Finding the number of diffusion steps for the best tradeoff is dependent on the data set and learning problem formulation. Generally, setups with stronger conditioning work with few diffusion steps, while less restrictive learning problems can benefit from more diffusion samples. In our experiments, the ideal thresholds emerged relatively clearly.

\subsection{Ablation on Training Rollout} \label{app: ablation training rollout}

\begin{figure}[p]
    \centering
    \includegraphics[width=0.99\textwidth]{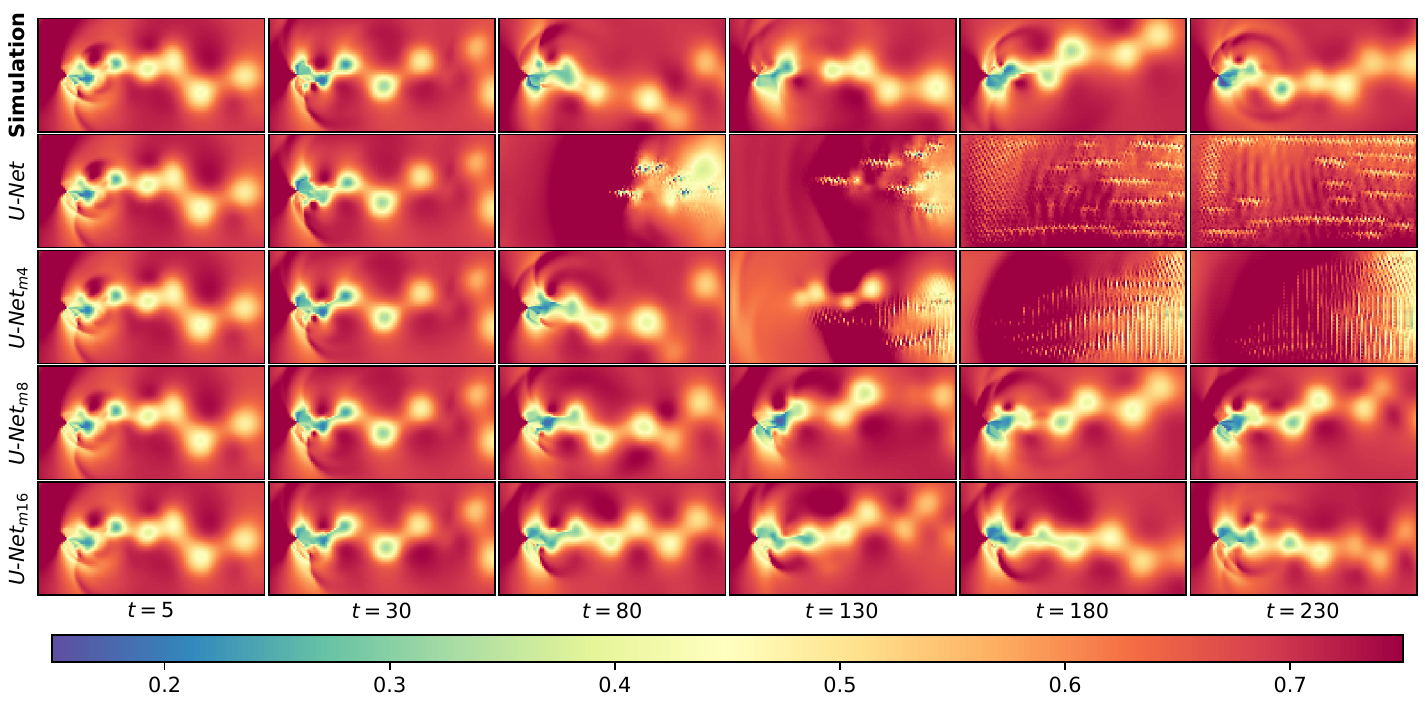}

    \caption{Training Rollout Ablation (see \cref{app: ablation training rollout}): Pressure predictions from \dTraLong{}.}
    \label{fig-app: example tra training rollout}
\end{figure}

\begin{figure}[p]
    \centering
    \includegraphics[width=0.99\textwidth]{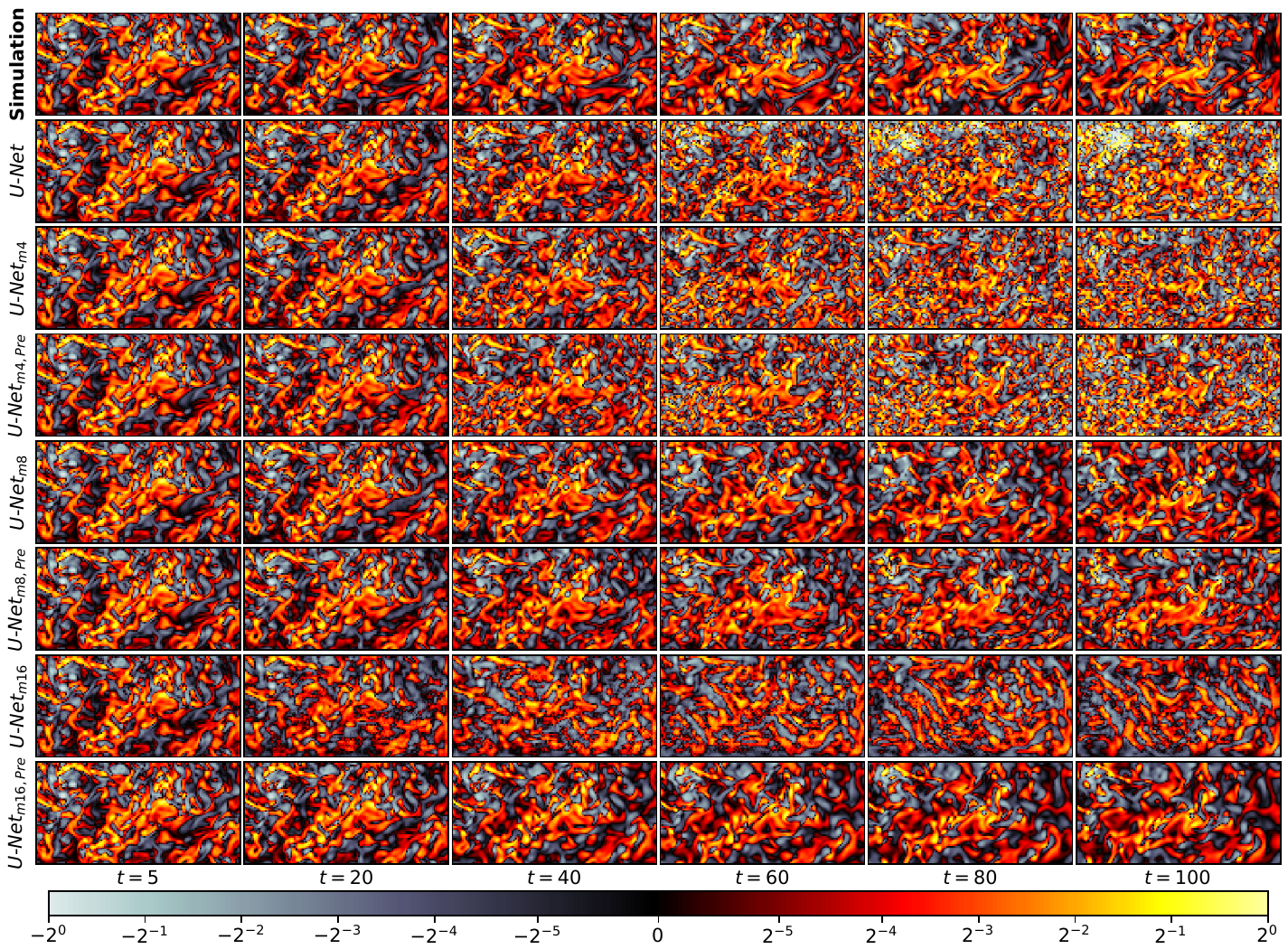}

    \caption{Training Rollout Ablation (see \cref{app: ablation training rollout}): Vorticity predictions from \dIso{}.}
    \label{fig-app: example iso training rollout}
\end{figure}

Here, we investigate the impact of unrolling the \mUnet{} model at training time, via varying the training rollout length \(m\). For these models, we use the U-Net architecture as described in \cref{app: unet implementation} with \( k=1 \) input steps. However, gradients are propagated through multiple state predictions during training, and corresponding MSE loss over all predicted steps is applied. Prediction examples can be found in \cref{fig-app: example tra training rollout,fig-app: example iso training rollout} in \cref{app: prediction examples ablation}.

\begin{table}[th]
    \caption{Accuracy ablation for different training rollout lengths \(m\) and pre-training (Pre.).}
    \label{tab-app: accuracy training rollout}
    \centering
    \small

    \begin{tabular}{l c c c c c c c c}
        \toprule
        & & & \multicolumn{2}{c}{\dTraExt{}{}} & \multicolumn{2}{c}{\dTraInt{}} & \multicolumn{2}{c}{\dIso{}}\\
        \cmidrule(lr){4-7} \cmidrule(lr){8-9}
        & & & MSE & LSiM & MSE & LSiM & MSE & LSiM\\

        \textbf{Method} & \(m\) & Pre.
        & (\(10^{-3}\)) & (\(10^{-1}\)) & (\(10^{-3}\)) & (\(10^{-1}\))
        & (\(10^{-2}\)) & (\(10^{-1}\))\\
        \midrule

        \mUnet{} & \(2\) & no
        & $3.1\pm2.1$  & $3.9\pm2.8$   &  $2.3\pm2.0$ & $3.3\pm2.8$
        & $25.8\pm35$ & $11.3\pm3.9$\\

        \mUnet{} & \(4\) & no
        & $1.6\pm1.0$ & $1.4\pm0.8$  &  $1.1\pm1.0$ & $0.9\pm0.4$
        & $3.7\pm0.8$ & $2.8\pm0.5$\\

        \mUnet{} & \(8\) & no
        & $1.6\pm0.7$ & $1.1\pm0.2$  &  $1.5\pm1.5$ & $1.0\pm0.5$
        & $4.5\pm2.8$ & $2.4\pm0.5$\\

        \mUnet{} & \(16\) & no
        & $2.2\pm1.1$ & $1.3\pm0.3$  &  $2.4\pm1.3$ & $1.3\pm0.5$
        & $13.0\pm11$ & $3.8\pm1.5$\\

        \midrule

        \mUnet{} & \(4\) & yes
        & \miss{} & \miss{}  &  \miss{} & \miss{}
        & $5.7\pm2.6$ & $3.6\pm0.8$\\

        \mUnet{} & \(8\) & yes
        & \miss{} & \miss{}  &  \miss{} & \miss{}
        & $2.6\pm0.6$ & $2.3\pm0.5$\\

        \mUnet{} & \(16\) & yes
        & \miss{} & \miss{}  &  \miss{} & \miss{}
        & $2.9\pm1.4$ & $2.3\pm0.5$\\

        \bottomrule
    \end{tabular}
\end{table}

\paragraph{Accuracy}
\Cref{tab-app: accuracy training rollout} shows models trained with different rollout lengths, and also includes the performance of \mUnet{} with \(m=2\) for reference. For the transonic flow, \(m=4\) is already sufficient to substantially improve the accuracy compared to \mUnet{} for the relatively short rollout of \(T=60\) steps during inference for \dTraExt{} and \dTraInt{}. Increasing the training rollout further does not lead to additional improvements and only slightly changes the accuracy. However, note that there is still a substantial difference between the temporal stability of \mUnet{m4} compared to \mUnet{m8} or \mUnet{m16} for cases with a longer inference rollout as analyzed below.

On \dIso{}, the behavior of \mUnet{} models with longer training rollout is clearly different as models with \( m > 4 \) substantially degrade compared to \(m=4\). The main reason for this behavior is that gradients from longer rollouts can be less useful for complex data when predictions strongly diverge from the ground truth in early training stages. Thus, we also considered variants, with \( m>2 \) that are finetuned from an initialization of a trained basic \mUnet{}, denoted by e.g., \mUnet{m4,Pre}. With this pre-training the previous behavior emerges, and \mUnet{m8,Pre} even clearly improves upon \mUnet{m4}.

\begin{figure}[ht]
    \centering
    \includegraphics[width=0.5\textwidth]{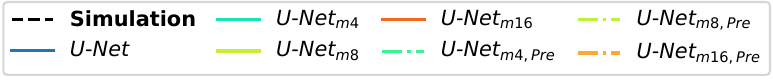}\\
    \includegraphics[width=0.49\textwidth]{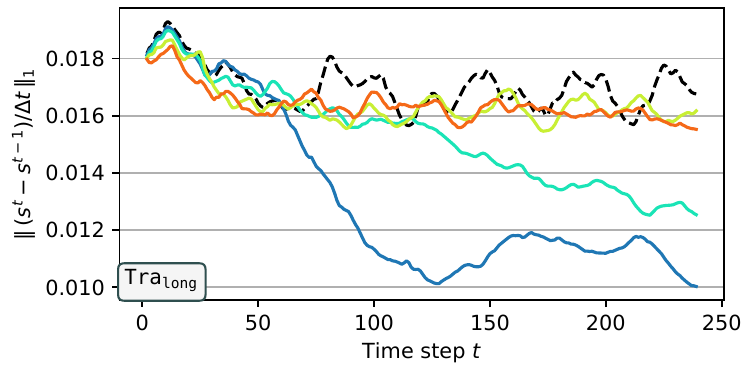}
    \hfill
    \includegraphics[width=0.49\textwidth]{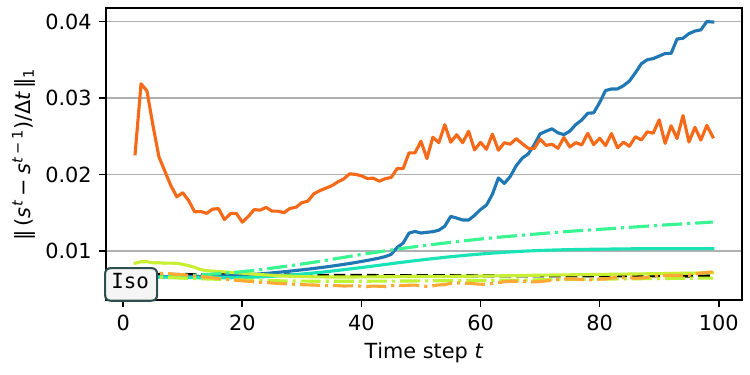}

    \caption{Temporal stability evaluation via error to previous time step for different training rollout lengths \(m\) on \dTraLong{} (left) and \dIso{} (right). Standard deviations are omitted for visual clarity.}
    \label{fig-app: stability training rollout}
\end{figure}

\paragraph{Temporal Stability}
In \cref{fig-app: stability training rollout}, we evaluate the temporal stability via the magnitude of the rate of change of \(s\), as detailed above. On \dTraLong{} all models perform similar until about \(t=50\) where \mUnet{} deteriorates. \mUnet{m4} also exhibits similar signs of deterioration around \(t=130\) during the rollout. Only \mUnet{m8} and \mUnet{m16} are fully stable across the entire rollout of \(T=240\) steps. On \dIso{}, \mUnet{m8} achieves comparable stability to \mACDM{}, with an almost constant rate of change for the entire rollout. Models with shorter rollouts, i.e., \mUnet{} and \mUnet{m4} deteriorate after an initial phase, and longer rollouts prevent effective training for \mUnet{m16} as explained above. The variants with additional pre-training are also included: \mUnet{m4,Pre} does not substantially improve upon \mUnet{m4}, and \mUnet{m8,Pre} performs very well, similar to \mUnet{m8}. Only for the longer rollouts in the \mUnet{m16,Pre} model pre-training clearly helps, as \mUnet{m16,Pre} is also fully stable.

\begin{table}[th]
    \caption{Accuracy ablation for different training noise standard deviations \(n\).}
    \label{tab-app: accuracy training noise}
    \centering
    \small

    \begin{tabular}{l c c c c c c c}
        \toprule
        & & \multicolumn{2}{c}{\dTraExt{}{}} & \multicolumn{2}{c}{\dTraInt{}} & \multicolumn{2}{c}{\dIso{}}\\
        \cmidrule(lr){3-6} \cmidrule(lr){7-8}
        & & MSE & LSiM & MSE & LSiM & MSE & LSiM\\

        \textbf{Method} & \(n\)
        & (\(10^{-3}\)) & (\(10^{-1}\)) & (\(10^{-3}\)) & (\(10^{-1}\))
        & (\(10^{-2}\)) & (\(10^{-1}\))\\
        \midrule

        \mUnet{} & \miss{}
        & $3.1\pm2.1$  & $3.9\pm2.8$   &  $2.3\pm2.0$ & $3.3\pm2.8$
        & $25.8\pm35$ & $11.3\pm3.9$\\

        \mUnet{} & \(1e\)--\(4\)
        & $2.7\pm1.8$ & $3.9\pm2.1$  &  $1.9\pm0.8$ & $2.4\pm2.1$
        & $16.0\pm22$ & $9.6\pm3.0$\\

        \mUnet{} & \(1e\)--\(3\)
        & $5.6\pm2.2$ & $3.3\pm2.5$  &  $3.5\pm1.6$ & $3.0\pm2.2$
        & $36.4\pm39$ & $12.9\pm2.2$\\

        \mUnet{} & \(1e\)--\(2\)
        & $1.4\pm0.8$ & $1.1\pm0.3$  &  $1.8\pm1.1$ & $1.0\pm0.4$
        & $3.1\pm0.9$ & $4.5\pm2.5$\\

        \mUnet{} & \(1e\)--\(1\)
        & $1.8\pm0.8$ & $1.2\pm0.2$  &  $2.2\pm2.0$ & $1.2\pm0.6$
        & $3.2\pm0.5$ & $2.9\pm0.6$\\

        \mUnet{} & \(1e0\)
        & $4.0\pm1.5$ & $1.8\pm0.3$  &  $11.4\pm6.3$ & $2.9\pm1.3$
        & $16.2\pm7.8$ & $7.5\pm2.7$\\

        \midrule
        
        \mACDM{ncn} & \miss{}
        & $4.1\pm1.9$ & $1.9\pm0.6$  &  $2.8\pm1.3$ & $1.7\pm0.4$
        & $18.3\pm2.5$ & $8.9\pm1.5$\\
        
        \mACDM{ncn} & \(1e\)--\(4\)
        & $3.8\pm1.5$ & $2.0\pm0.3$  &  $4.3\pm2.3$ & $1.7\pm0.4$
        & $14.2\pm1.7$ & $8.1\pm1.2$\\

        \mACDM{ncn} & \(1e\)--\(3\)
        & $3.6\pm1.4$ & $2.2\pm0.3$  &  $3.9\pm2.3$ & $1.8\pm0.4$
        & $11.1\pm3.8$ & $8.5\pm1.6$\\

        \mACDM{ncn} & \(1e\)--\(2\)
        & $3.6\pm1.6$ & $1.7\pm0.4$  &  $2.6\pm2.3$ & $1.3\pm0.5$
        & $26.7\pm25$ & $12.2\pm2.8$\\

        \mACDM{ncn} & \(1e\)--\(1\)
        & $3.6\pm1.9$ & $1.5\pm0.4$  &  $2.5\pm2.2$ & $1.2\pm0.6$
        & $2.8\pm0.6$ & $4.0\pm2.2$\\

        \mACDM{ncn} & \(1e0\)
        & $4.2\pm1.7$ & $1.8\pm0.4$  &  $6.2\pm2.8$ & $2.0\pm0.6$
        & $11.1\pm1.4$ & $6.2\pm0.9$\\
        
        \bottomrule
    \end{tabular}
\end{table}

\paragraph{Summary}
Compared to \mACDM{}, the variants of \mUnet{} with longer training rollouts can achieve similar or slightly higher accuracy and a equivalent temporal stability at a faster inference speed. However, this method requires additional computational resources during training, both in terms of memory over the rollout as well as training time. For example, \mUnet{m16,Pre} on \dIso{} increases the required training time (\(90h\) of pre-training + \(260h\) of refinement) by a factor of more than $5.6\times$ at equal epochs and memory compared to \mACDM{R100} as shown in \cref{tab-app: performance}. Naturally, longer training rollouts do not provide \mUnet{} with the ability for posterior sampling.

\subsection{Ablation on Training Noise} \label{app: ablation training noise}

\begin{figure}[p]
    \centering
    \includegraphics[width=0.99\textwidth]{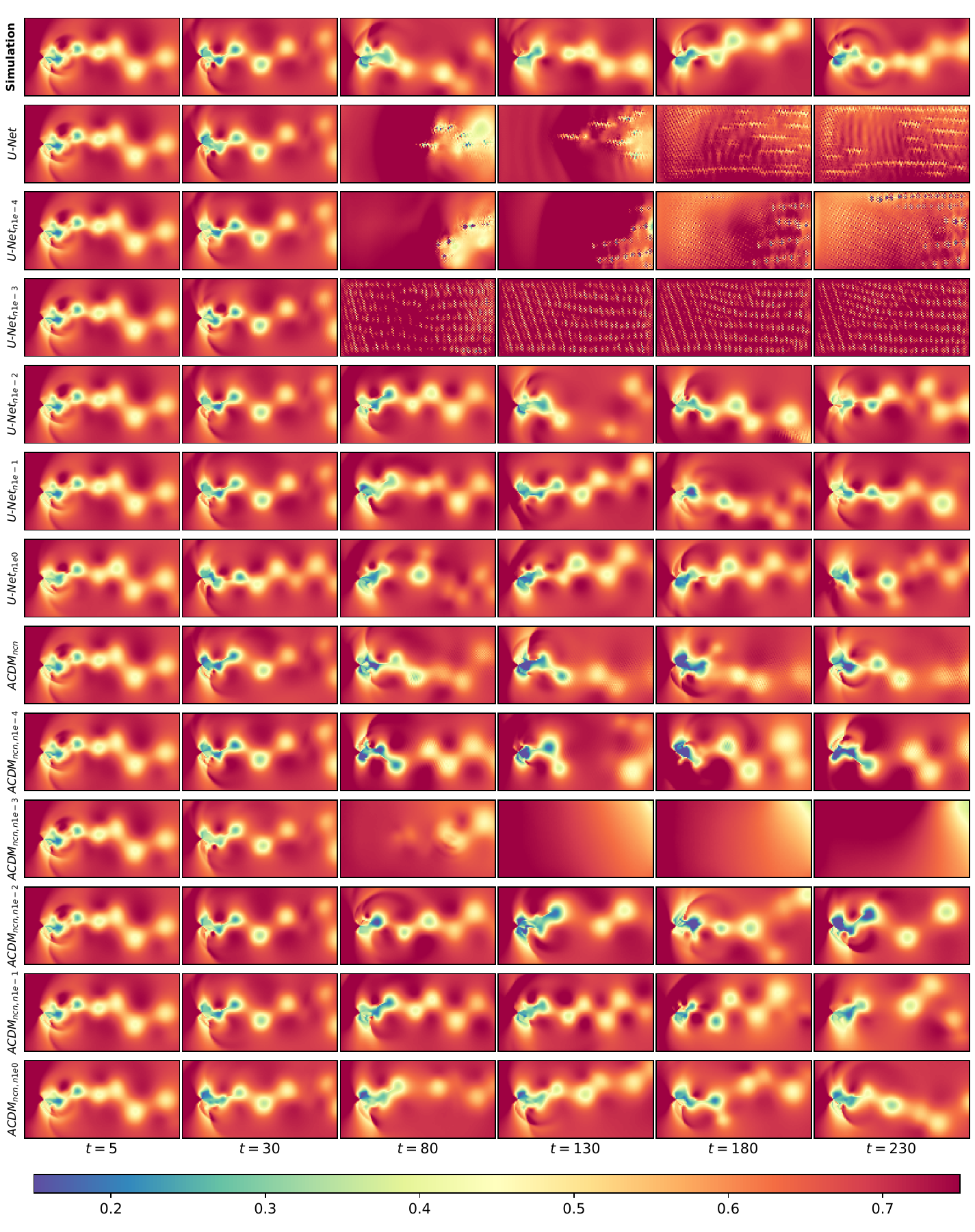}

    \caption{Training Noise Ablation (see \cref{app: ablation training noise}): Pressure predictions from \dTraLong{}.}
    \label{fig-app: example tra training noise}
\end{figure}

\begin{figure}[p]
    \centering
    \includegraphics[width=0.99\textwidth]{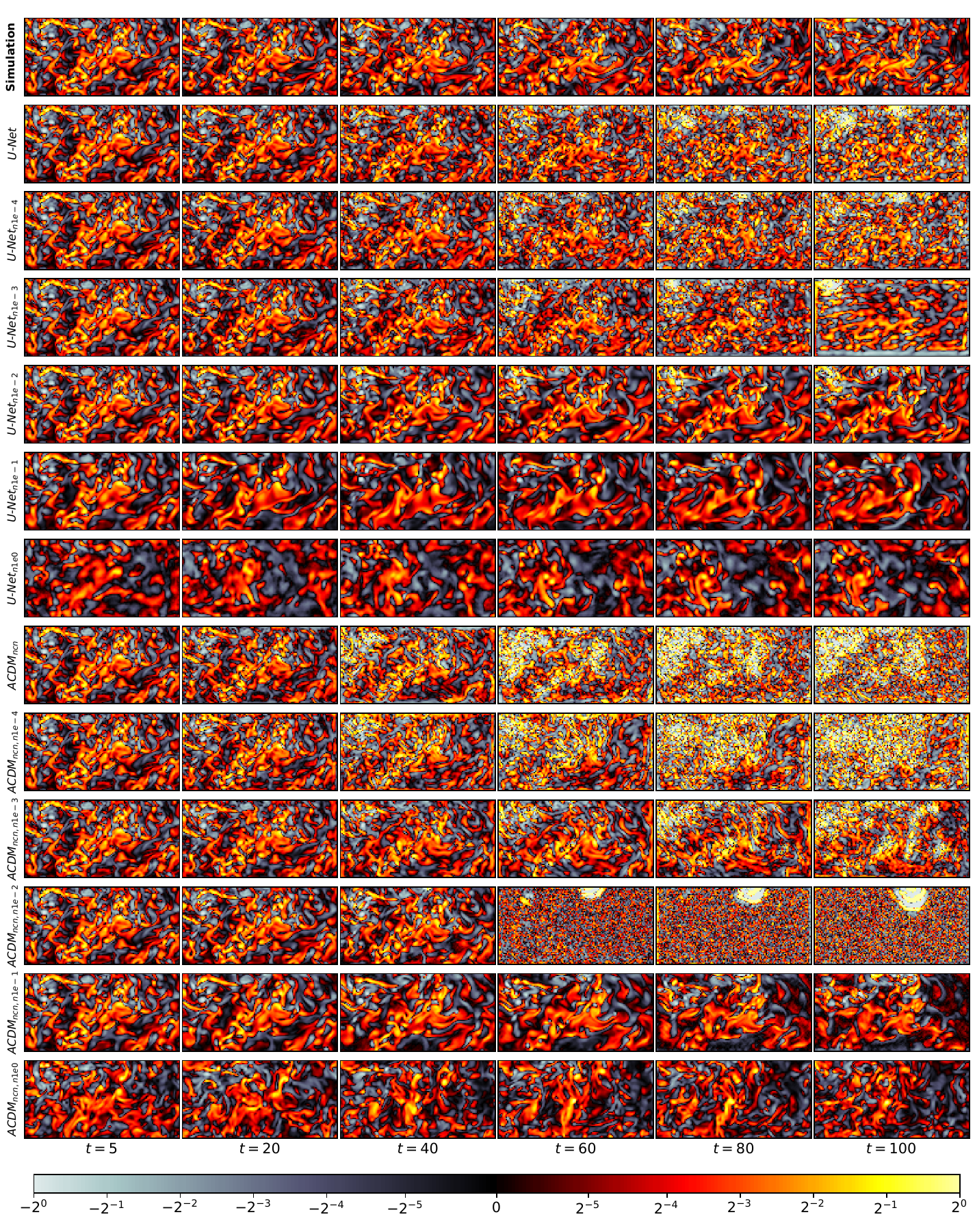}

    \caption{Training Noise Ablation (see \cref{app: ablation training noise}): Vorticity predictions from \dIso{}.}
    \label{fig-app: example iso training noise}
\end{figure}

We investigate the usage of training noise \parencite{sanchez-gonzalez2020_Learning} to stabilize predictions, as it features interesting connections to our method. Instead of generating predictions from noise to achieve temporal stability, this method relies on the addition of noise to the training inputs, to simulate error accumulation during training. In this way, the model adapts to disturbances during training, such that the data shift is reduced once errors inevitably accumulate during the inference rollout, leading to increased temporal stability. We test this approach on \mUnet{} and on \mACDM{ncn}. The latter evaluation serves as an example to understand if the lost tolerance for error accumulation in \mACDM{ncn}, the setup without conditioning noise, can be replaced with training noise. This \mACDM{ncn} version is not intended as a practical architecture as it inherits the drawbacks of both methods, the inference cost from diffusion models, and the overhead and additional hyperparameters from added training noise. For this ablation, we use the \mACDM{ncn} model as described in \cref{app: acdm implementation}, but add training noise to every model input in the same way as for the \mUnet{} with training noise (see \cref{app: unet implementation}). Here, \mUnet{} or \mACDM{ncn} models trained using training noise with a standard deviation of e.g., \(n=10^{-1}\), are denoted by \mUnet{n1e-1} or \mACDM{ncn,n1e-1} respectively. Prediction examples can be found in \cref{fig-app: example tra training noise,fig-app: example iso training noise} in \cref{app: prediction examples ablation}.

\begin{figure}[hbt]
    \centering
    \includegraphics[width=0.49\textwidth]{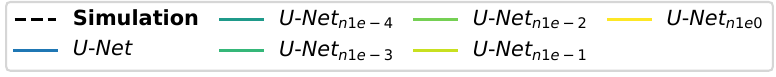}
    \hfill
    \includegraphics[width=0.49\textwidth]{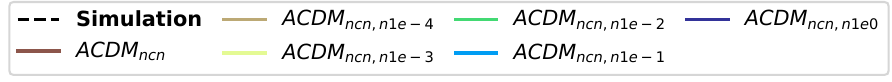}\\
    \includegraphics[width=0.49\textwidth]{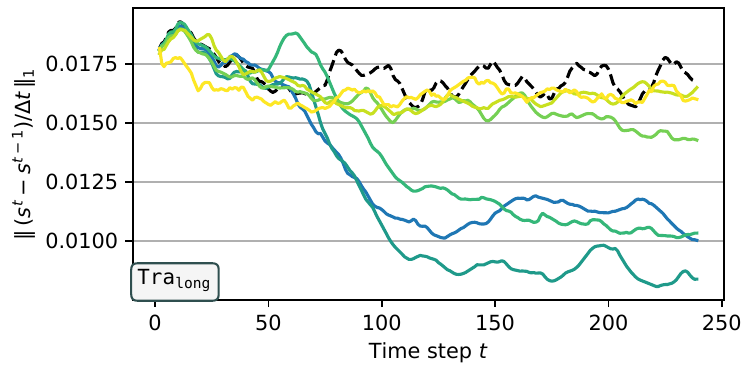}
    \hfill
    \includegraphics[width=0.49\textwidth]{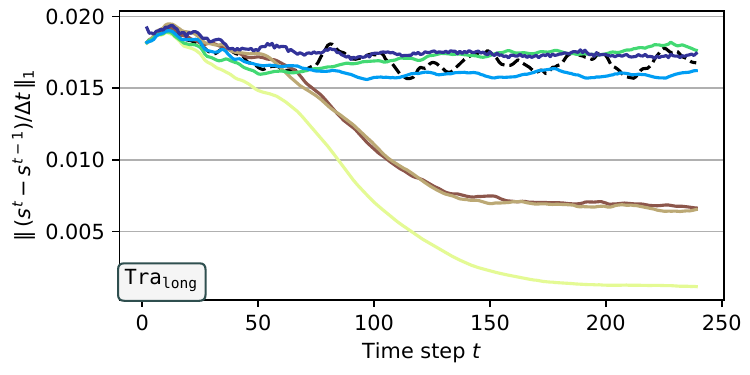}\\
    \includegraphics[width=0.49\textwidth]{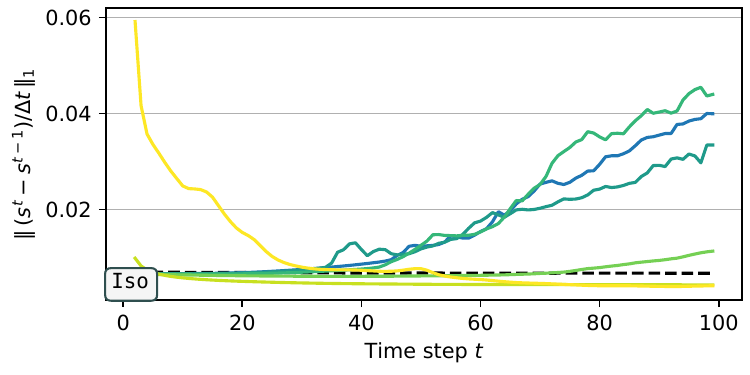}
    \hfill
    \includegraphics[width=0.49\textwidth]{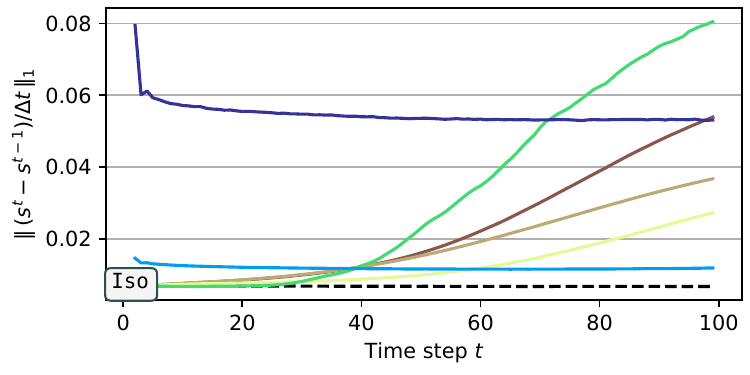}

    \caption{Temporal stability evaluation for different training noise standard deviations \(n\) of \mUnet{} (left) and \mACDM{ncn} (right) on \dTraLong{} (top) and \dIso{} (bottom). Standard deviations are omitted for visual clarity.}
    \label{fig-app: stability training noise}
\end{figure}

\paragraph{Accuracy}
The accuracy of \mUnet{} and \mACDM{ncn} setups with training noise using different standard deviations \(n\) is analyzed in \cref{tab-app: accuracy training noise}. On \dTra{}, the accuracy trend is not fully consistent. Small values of \(n\) such as \(10^{-4}\) and \(10^{-3}\) occasionally even reduce the final performance, but training noise with a well-tuned standard deviation between \(10^{-2}\) and \(10^{-1}\) does increase accuracy. Choosing very large standard deviations corrupts the training data too much, and reduces accuracy again as expected. The results on the isotropic turbulence experiment show a similar behavior for \mUnet{} as well as \mACDM{ncn}.

\paragraph{Temporal Stability}
In \cref{fig-app: stability training noise}, we evaluate the temporal stability of models with training noise via the magnitude of the rate of change of \(s\), as detailed above. On \dTraLong{} both architectures \mUnet{} and \mACDM{ncn} behave similarly: while training noise with a standard deviation \(n\) that is too low does not improve the stability and occasionally even deteriorates it, finding a suitable magnitude is key for stable inference rollouts. In both cases, values of \(n\) between \(10^{-2}\) and \(10^{-1}\) produce the best results. Increasing the noise further has detrimental effects, as for example slight overshooting and high-frequency fluctuations occur for \mACDM{ncn,n1e0} or predictions can diverge early from the simulation for \mUnet{n1e0}. On \dIso{}, a similar stabilizing effect from training noise can be observed, given the noise magnitude is tuned sufficiently: While lower standard deviations barely alter the time point \(t=40\), where predictions diverge from the reference simulation, too much training noise already causes major problems at the very beginning of the prediction.

\begin{figure}[ht]
    \centering
    \includegraphics[width=0.49\textwidth]{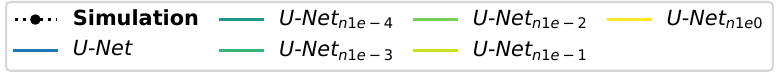}
    \hfill
    \includegraphics[width=0.49\textwidth]{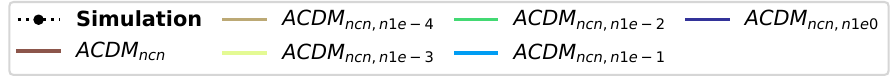}\\
    
    \includegraphics[width=0.49\textwidth]{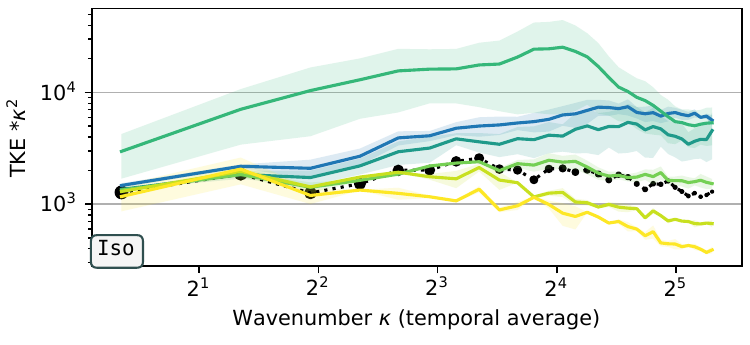}
    \hfill
    \includegraphics[width=0.49\textwidth]{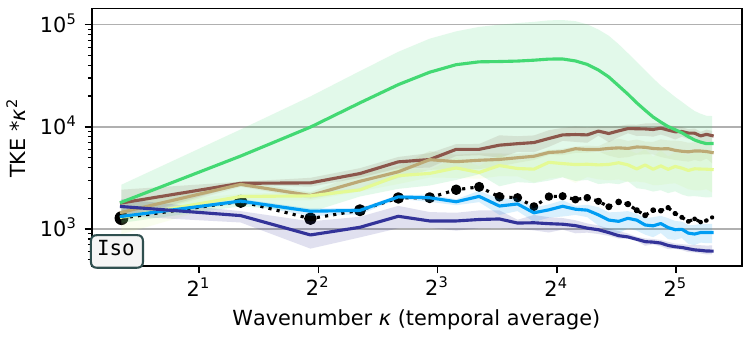}

    \caption{Spatial frequency analysis via the turbulent kinetic energy (TKE) on a sequence from \dIso{} with \( z = 300 \) for the training noise ablations on \mUnet{} (left) and \mACDM{ncn} (right).}
    \label{fig-app: frequency training noise}
\end{figure}

This behavior can also be observed on a spatial spectral analysis via the TKE in \cref{fig-app: frequency training noise}, where the training noise can balance predictions between under- and overshooting. For both \mUnet{} and \mACDM{ncn}, training noise with a suitable magnitude can result in a comparable temporal stability to \mACDM{}, that includes noise on the conditioning.

\paragraph{Summary}
Training \mUnet{} with training noise can achieve similar or slightly higher accuracy and a competitive temporal stability compared to \mACDM{}. While this method exhibits faster inference speeds, it does rely on the additional noise variance hyperparameter, that can even reduce performance if not tuned well. Furthermore, training noise does not provide deterministic models with the ability for posterior sampling. Interestingly, the lost error tolerance of the \mACDM{ncn} architecture without conditioning noise, can be mostly restored with training noise of suitable magnitude.

\subsection{Ablation on Training with an LSiM Loss} \label{app: ablation lsim training}

In this section, we investigate usage of the LSiM metric \parencite{kohl2020_Learning} as an additional loss term, similar to perceptual losses in the computer vision domain \parencite{dosovitskiy2016_Generating, johnson2016_Perceptual}. This means, in addition to training \mUnet{} with an MSE loss as above, the differentiable learned LSiM metric model is also used during back-propagation. Given a predicted state \(s^{t}\) and the corresponding ground truth state \(\hat{s}^{t}\), we evaluate the training loss as
\[ \mathcal{L}_\mathit{MSE+LSiM} = \left( s^{t} - \hat{s}^{t} \right)^2 + \lambda * \operatorname{LSiM}(s^{t}, \hat{s}^{t}) \]
while leaving the inference of the models untouched. To use LSiM, each field from both states is individually normalized to \([0,255]\). The resulting loss values are aggregated with an average operation across fields. Fields containing the scalar simulation parameters are not evaluated with this metric. In the following, the impact of \(\lambda\), the weight that controls the influence of the LSiM loss, is investigated. \mUnet{} models trained with e.g., \(\lambda=10^{-1}\), are denoted by \mUnet{$\lambda$1e-1}.

\begin{table}[th]
    \caption{Accuracy ablation for training with LSiM losses of different strengths \(\lambda\).}
    \label{tab-app: accuracy lsim training}
    \centering
    \small

    \begin{tabular}{l c c c c c c c}
        \toprule
        & & \multicolumn{2}{c}{\dTraExt{}{}} & \multicolumn{2}{c}{\dTraInt{}} & \multicolumn{2}{c}{\dIso{}}\\
        \cmidrule(lr){3-6} \cmidrule(lr){7-8}
        & & MSE & LSiM & MSE & LSiM & MSE & LSiM\\

        \textbf{Method} & \(\lambda\)
        & (\(10^{-3}\)) & (\(10^{-1}\)) & (\(10^{-3}\)) & (\(10^{-1}\))
        & (\(10^{-2}\)) & (\(10^{-1}\))\\
        \midrule

        \mUnet{} & \miss{}
        & $3.1\pm2.1$  & $3.9\pm2.8$   &  $2.3\pm2.0$ & $3.3\pm2.8$
        & $25.8\pm35$ & $11.3\pm3.9$\\

        \mUnet{} & \(1e\)--\(5\)
        & $4.2\pm2.9$ & $4.5\pm3.0$  &  $2.6\pm2.2$ & $2.1\pm2.0$
        & $67.4\pm75.7$ & $12.4\pm3.8$\\

        \mUnet{} & \(1e\)--\(4\)
        & $2.3\pm1.2$ & $3.7\pm2.6$  &  $1.6\pm1.4$ & $2.0\pm1.8$
        & $12.3\pm9.3$ & $11.8\pm2.5$\\
    
        \mUnet{} & \(1e\)--\(3\)
        & $2.9\pm1.9$ & $1.7\pm0.8$  &  $2.2\pm2.3$ & $1.5\pm0.9$
        & $6.3\pm3.1$ & $9.4\pm2.8$\\

        \mUnet{} & \(1e\)--\(2\)
        & $4.5\pm1.3$ & $3.5\pm1.1$  &  $3.0\pm2.3$ & $1.8\pm0.9$
        & \diverge{$0.1b\pm0.2b$} & \diverge{$15.3\pm1.2$}\\

        \mUnet{} & \(1e\)--\(1\)
        & $5.8\pm1.8$ & $3.0\pm0.8$  &  $5.2\pm1.9$ & $2.3\pm0.6$
        & \diverge{$12b\pm29b$} & \diverge{$15.0\pm1.0$}\\

        \mUnet{} & \(1e0\)
        & $6.8\pm1.5$ & $4.8\pm1.1$  &  $6.6\pm3.0$ & $2.4\pm0.7$
        & \diverge{$17b\pm552b$} & \diverge{$14.9\pm1.0$}\\

        \bottomrule
    \end{tabular}
\end{table}

\paragraph{Accuracy}
In terms of accuracy, adding very small amounts of the LSiM term with \(\lambda=10^{-5}\) to the MSE loss does decrease performance, most likely due to suboptimal gradient signals through the additional steps during back-propagation, as shown in \cref{tab-app: accuracy lsim training}. Similarly, adding too much, such that it predominantly influences the overall loss causes problems. Especially on \dIso{}, this causes models to aggressively diverge after \(30-40\) prediction steps, leading to errors in the range of \(10^9\) (\(b\)) in \cref{tab-app: accuracy lsim training}. As expected, choosing a suitable loss magnitude around \(\lambda=10^{-3}\) substantially reduces errors in terms of LSiM across test sets. However, the added loss term does also improve performance in terms of MSE, as similarly observed in the image domain \parencite{dosovitskiy2016_Generating,johnson2016_Perceptual}.

\begin{figure}[ht]
    \centering
    \includegraphics[width=0.50\textwidth]{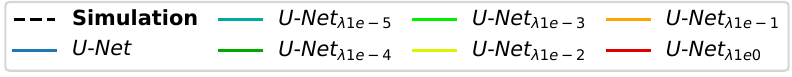}\\
    \includegraphics[width=0.49\textwidth]{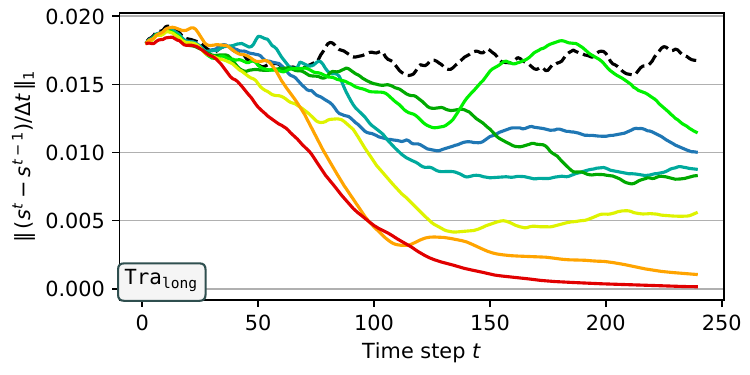}
    \hfill
    \includegraphics[width=0.49\textwidth]{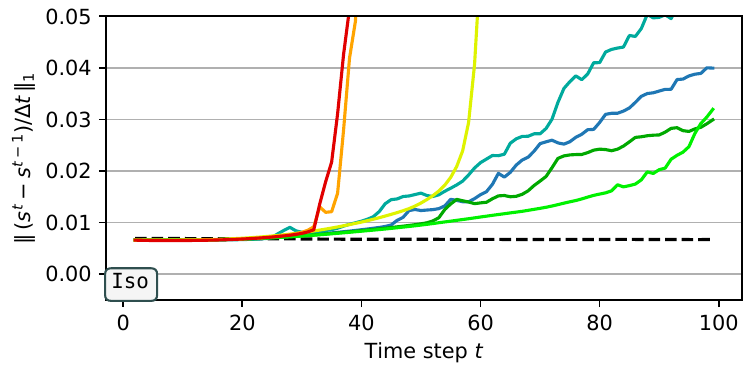}

    \caption{Temporal stability evaluation of \mUnet{} for training with LSiM losses of different strengths \(\lambda\) on \dTraLong{} (left) and \dIso{} (right). Standard deviations are omitted for visual clarity.}
    \label{fig-app: stability lsim training}
\end{figure}

\begin{figure}[ht]
    \centering
    \includegraphics[width=0.50\textwidth]{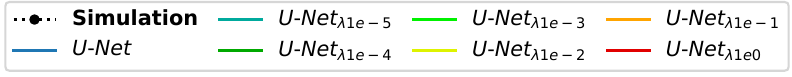}\\
    \begin{adjustbox}{valign=c}
    \includegraphics[width=0.49\textwidth]{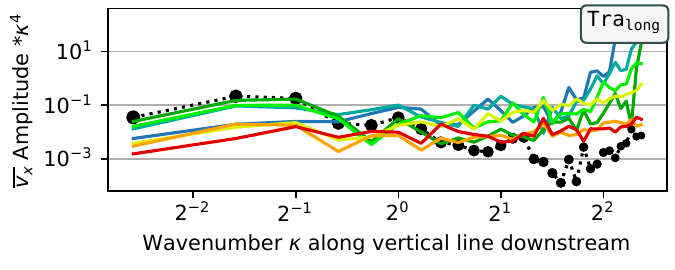}
    \end{adjustbox}
    \hfill
    \begin{adjustbox}{valign=c}
    \includegraphics[width=0.49\textwidth]{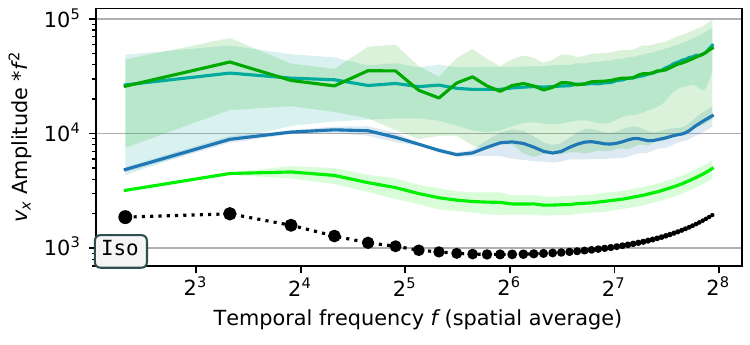}
    \end{adjustbox}

    \caption{Spatial frequency along a vertical line one cylinder diameter downstream on \dTraLong{} (left) and temporal frequency analysis on a sequence from \dIso{} with \( z = 300 \) (right) for the LSiM loss ablation models.}
    \label{fig-app: frequency lsim training}
\end{figure}

\paragraph{Temporal Stability}
In line with the accuracy results, \mUnet{$\lambda$1e-3} exhibits improved temporal stability compared to \mUnet{} as displayed in \cref{fig-app: stability lsim training}. Choosing unsuitable \(\lambda\) causes models to diverge earlier from the reference trajectory when evaluating the difference between predictions steps, for both \dTraLong{} and \dIso{}. When analyzing the frequency behavior of the models trained with LSiM in \cref{fig-app: frequency lsim training} the results are similar: Improved performance across the frequency band can be observed for \mUnet{$\lambda$1e-3}, while smaller values of \(\lambda\) are less potent and can be detrimental or only slightly beneficial compared to \mUnet{}.

\paragraph{Summary}
Training \mUnet{} with LSiM as an additional loss term, can increase accuracy, temporal stability, and frequency behavior across evaluations. However, the resulting models are neither competitive compared to other stabilization techniques discussed above, such as training rollouts or training noise, nor to the proposed diffusion architecture.

\subsection{Ablation on U-Net Modernizations} \label{app: ablation unet modernizations}
As described in \cref{app: unet implementation}, our U-Net implementation follows established diffusion model architectures, that contain a range of modernizations compared to the original approach proposed by \textcite{ronneberger2015_UNet}. Here, we compare to a more traditional U-Net architecture which is known to work well for fluid problems. We adapted the DFP model implementation\footnote{\url{https://github.com/thunil/Deep-Flow-Prediction}} of \parencite{thuerey2020_Deep} for our settings. The architecture features:
\begin{itemize}
    \item batch normalization instead group normalization, and no attention layers in the blocks,
    \item six downsampling blocks consisting of strided convolutions and leaky ReLU layers,
    \item six upsampling blocks consisting of convolution, bilinear upsamling, and ReLU layers,
    \item six feature map levels with spatial sizes of \(64 \times 32\), \(32 \times 16\), \(16 \times 8\), \(8 \times 4\), \(4 \times 2\), and \(2 \times 1\),
    \item an increasing number of channels for deeper features, i.e., \(72\), \(72\), \(144\), \(288\), \(288\), and \(288\).
\end{itemize}
It is trained as a direct one-step predictor (\mDFP{}) in the same way as described in \cref{app: unet implementation}, as well as employing it in the diffusion setup as a backbone architecture (\mDFP{ACDM}). In both cases, we keep all other hyperparameters identical with the corresponding baseline architecture.

\begin{table}[th]
    \caption{Accuracy of the ``modern'' U-Net architecture compared to \mDFP{}.}
    \label{tab-app: accuracy unet modernizations}
    \centering
    \small

    \begin{tabular}{l c c c c c c}
        \toprule
        & \multicolumn{2}{c}{\dTraExt{}{}} & \multicolumn{2}{c}{\dTraInt{}} & \multicolumn{2}{c}{\dIso{}}\\
        \cmidrule(lr){2-5} \cmidrule(lr){6-7}
        & MSE & LSiM & MSE & LSiM & MSE & LSiM\\

        \textbf{Method}
        & (\(10^{-3}\)) & (\(10^{-1}\)) & (\(10^{-3}\)) & (\(10^{-1}\))
        & (\(10^{-2}\)) & (\(10^{-1}\))\\
        \midrule

        \mUnet{}
        & $3.1\pm2.1$ & $3.9\pm2.8$   &  $2.3\pm2.0$ & $3.3\pm2.8$
        & $25.8\pm35$ & $11.3\pm3.9$\\

        \mDFP{}
        & $4.5\pm1.3$ & $3.9\pm0.7$  &  $4.8\pm2.1$ & $3.6\pm1.7$
        & $5.1\pm1.3$ & $5.1\pm2.0$\\

        \midrule

        \mACDM{}
        & $2.3\pm1.4$ & $1.3\pm0.3$  &  $2.7\pm2.1$ & $1.3\pm0.6$
        & $3.7\pm0.8$ & $3.3\pm0.7$\\

        \mDFP{ACDM}
        & \diverge{$NaN$} & \diverge{$NaN$}  &  \diverge{$NaN$} & \diverge{$NaN$}
        & \diverge{$NaN$} & \diverge{$NaN$}\\

        \bottomrule
    \end{tabular}
\end{table}

\paragraph{Accuracy}
\Cref{tab-app: accuracy unet modernizations} shows a comparison of both architectures compared to \mUnet{} and \mACDM{} on our more challenging data sets \dTra{} and \dIso{}. Training \mDFP{ACDM} as a diffusion backbone (with additional time embeddings for the diffusion step \(r\) as discussed in \cref{app: acdm implementation}) failed to generalize beyond the first few prediction time steps across test sets in our experiments. This highlights the general usefulness of the recently introduced modernizations to the U-Net architecture. There is a noticeable drop in accuracy on \dTra{} for \mDFP{} compared to \mUnet{}, but it performs clearly better than \mUnet{} on \dIso{}, however still lacking compared to \mACDM{}. This unexpected trend in accuracy is mainly caused by the different rollout behavior of these architectures discussed in the following.

\begin{figure}[ht]
    \centering
    \includegraphics[width=0.50\textwidth]{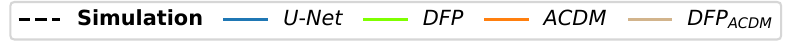}\\
    \includegraphics[width=0.49\textwidth]{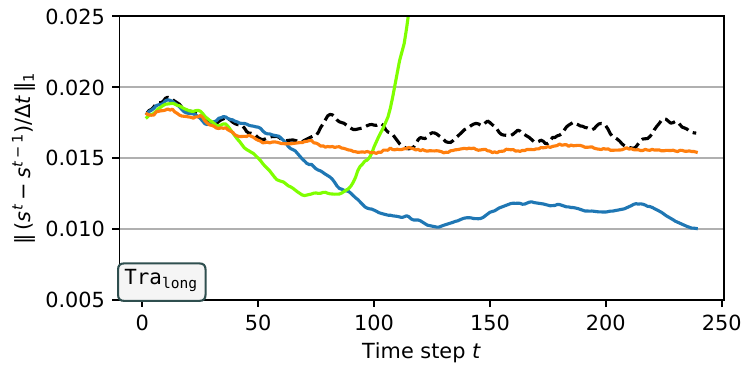}
    \hfill
    \includegraphics[width=0.49\textwidth]{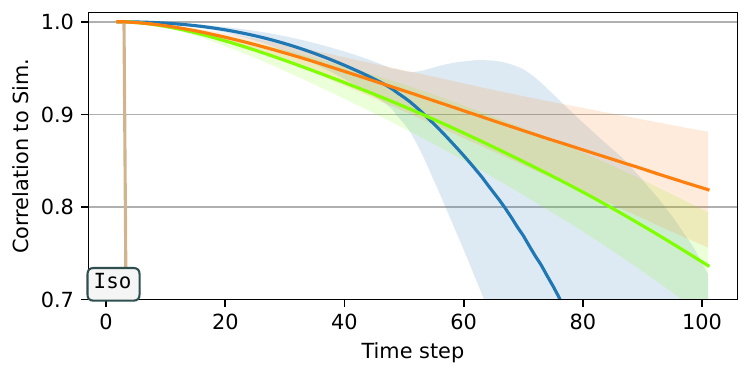}

    \caption{Temporal stability evaluation via error to previous time step on \dTraLong{} (left) and the correlation to the ground truth on \dIso{} (right) for the U-Net modernization ablation.}
    \label{fig-app: stability unet modernizations}
\end{figure}

\paragraph{Temporal Stability}
\Cref{fig-app: stability unet modernizations} shows temporal stability evaluations of the variants on \dTraLong{} and \dIso{}, that illustrate the different rollout behavior of \mDFP{} compared to \mUnet{} depending on the data set. On \dTraLong{} on the top, \mDFP{} diverges earlier and more substantially compared to \mUnet{}, when measured via the difference to the previously predicted time step. However, the rollout behavior is different on \dIso{}, as illustrated via the Pearson correlation coefficient to the ground truth trajectory on the bottom in \cref{fig-app: stability unet modernizations}. The simpler \mDFP{} model decorrelates more quickly for the first 50 steps, while keeping a relatively constant decorrelation rate. \mUnet{} is initially more in line with the reference, however it sharply decreases after about 50 steps, meaning errors accumulate more quickly after an initial phase of higher stability. 

\paragraph{Summary}
We found the recently proposed architecture modernizations to U-Nets to be an important factor, when employing them as backbones in a diffusion-based setup. For some direct prediction cases, the modernizations can delay diverging behavior due to unrolling during inference to some degree. On other data, using no modernizations can be beneficial for longer rollouts in direct prediction setting, but this comes at the costs of less initial accuracy, and lacking capacities as a diffusion backbone.

\subsection{Ablations on PDE-Refiner} \label{app: ablation refiner}

\begin{figure}[p]
    \centering
    \includegraphics[width=0.88\textwidth]{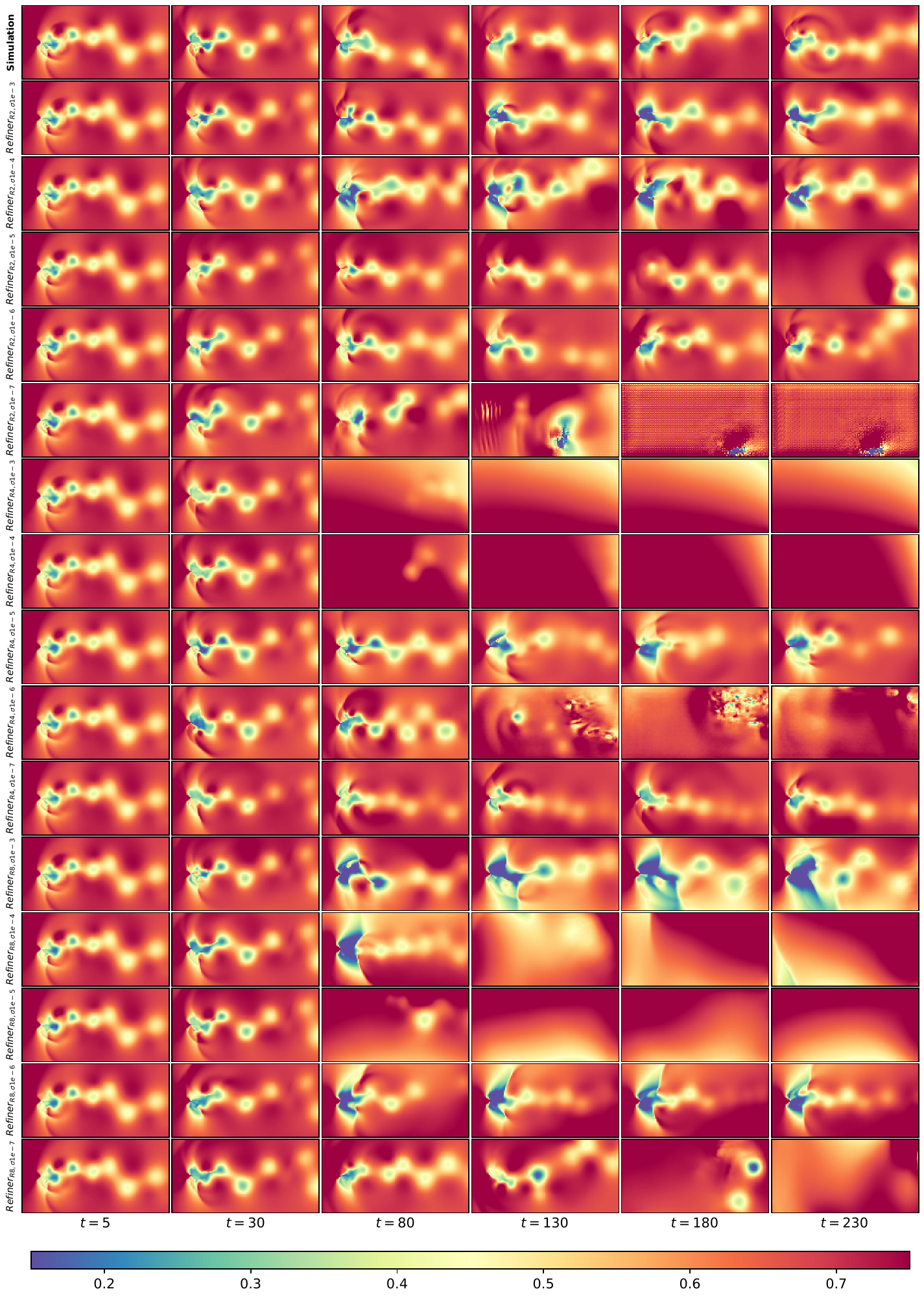}

    \caption{Comparison to PDE-Refiner (see \cref{app: ablation refiner}): Pressure predictions from \dTraLong{}.}
    \label{fig-app: example tra refiner}
\end{figure}

\begin{figure}[p]
    \centering
    \includegraphics[width=0.88\textwidth]{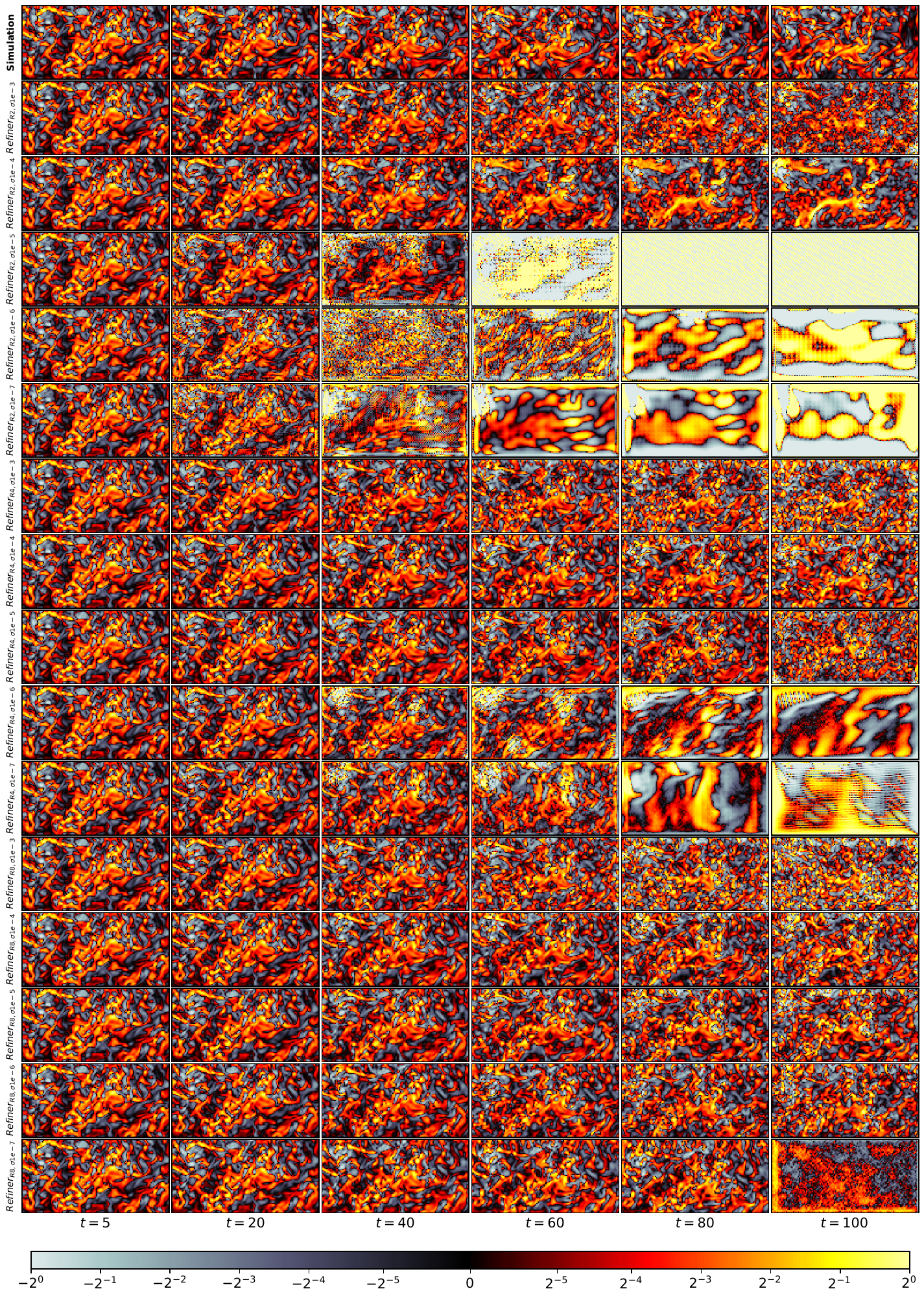}

    \caption{Comparison to PDE-Refiner (see \cref{app: ablation refiner}): Vorticity predictions from \dIso{}.}
    \label{fig-app: example iso refiner}
\end{figure}

Here, we investigate PDE-Refiner in more detail, especially with respect to the number of refinement steps \(R\) and the minimum noise variance \(\sigma\), its key hyperparameters. We sweep over combinations of \(R \in \{2,4,8\}\) and \(\sigma \in \{10^{-7}, 10^{-6}, 10^{-5}, 10^{-4}, 10^{-3}\}\), and report accuracy, temporal stability, and posterior sampling results. Due to computational constraints for this large sweep, only one model per combination is trained. Five samples from each model are considered, as above. We denote models trained with e.g., \(R=2\) and \(\sigma = 10^{-3}\) by \mRefiner{R2,$\sigma$1e-3} in the following. Prediction examples can be found in \cref{fig-app: example tra refiner,fig-app: example iso refiner} in \cref{app: prediction examples ablation}.

\begin{table}[th]
    \caption{Accuracy comparison for PDE-Refiner using different refinement steps \(R\) and noise variances \(\sigma\).}
    \label{tab-app: accuracy refiner}
    \centering
    \small

    \begin{tabular}{l c c c c c c c c}
        \toprule
        & & & \multicolumn{2}{c}{\dTraExt{}{}} & \multicolumn{2}{c}{\dTraInt{}} & \multicolumn{2}{c}{\dIso{}}\\
        \cmidrule(lr){4-7} \cmidrule(lr){8-9}
        & & & MSE & LSiM & MSE & LSiM & MSE & LSiM\\

        \textbf{Method} & \(R\) & \(\sigma\)
        & (\(10^{-3}\)) & (\(10^{-1}\)) & (\(10^{-3}\)) & (\(10^{-1}\))
        & (\(10^{-2}\)) & (\(10^{-1}\))\\
        \midrule

        \mACDM{} & \miss{} & \miss{}
        & $2.3\pm1.4$ & $1.3\pm0.3$  &  $2.7\pm2.1$ & $1.3\pm0.6$
        & $3.7\pm0.8$ & $3.3\pm0.7$\\

        \mUnet{} & \miss{} & \miss{}
        & $3.1\pm2.1$ & $3.9\pm2.8$   &  $2.3\pm2.0$ & $3.3\pm2.8$
        & $25.8\pm35$ & $11.3\pm3.9$\\
        \midrule

        \mRefiner{} & \(2\) &\(1e\)--\(3\)
        & $3.3\pm1.3$ & $1.4\pm0.3$   &  $3.9\pm1.6$ & $1.4\pm0.3$
        & $6.1\pm1.9$ & $7.2\pm1.6$\\

        \mRefiner{} & \(2\) &\(1e\)--\(4\)
        & $12.7\pm2.9$ & $4.2\pm0.5$   &  $10.1\pm1.5$ & $2.4\pm0.3$
        & \diverge{$0.1m\pm0.3m$} & \diverge{$12.5\pm5.2$}\\

        \mRefiner{} & \(2\) &\(1e\)--\(5\)
        & $4.8\pm1.4$ & $2.6\pm0.3$   &  $4.0\pm3.1$ & $2.1\pm0.5$
        & \diverge{$3.3e30$} & \diverge{$15.2\pm0.9$}\\

        \mRefiner{} & \(2\) &\(1e\)--\(6\)
        & $5.0\pm1.9$ & $2.0\pm0.3$   &  $3.6\pm2.6$ & $1.9\pm0.4$
        & \diverge{$0.1m\pm0.2m$} & \diverge{$16.1\pm1.0$}\\

        \mRefiner{} & \(2\) &\(1e\)--\(7\)
        & $13.6\pm9.9$ & $6.1\pm4.0$   &  $54.6\pm68.7$ & $6.7\pm5.0$
        & \diverge{$22k\pm13k$} & \diverge{$14.9\pm0.9$}\\
        \midrule

        \mRefiner{} & \(4\) &\(1e\)--\(3\)
        & $5.3\pm0.8$ & $3.2\pm0.4$   &  $6.0\pm1.2$ & $2.6\pm0.4$
        & $5.1\pm1.8$ & $4.7\pm0.8$\\

        \mRefiner{} & \(4\) &\(1e\)--\(4\)
        & $3.4\pm2.0$ & $1.9\pm0.3$   &  $5.7\pm2.4$ & $1.9\pm0.5$
        & $7.0\pm3.1$ & $5.0\pm1.0$\\

        \mRefiner{} & \(4\) &\(1e\)--\(5\)
        & $7.0\pm1.7$ & $2.7\pm0.4$   &  $3.1\pm0.8$ & $1.7\pm0.2$
        & $4.9\pm2.0$ & $7.6\pm2.1$\\

        \mRefiner{} & \(4\) &\(1e\)--\(6\)
        & $3.5\pm1.1$ & $2.1\pm0.5$   &  $8.8\pm0.9$ & $4.3\pm2.1$
        & $66.1\pm38.4$ & $11.7\pm0.7$\\

        \mRefiner{} & \(4\) &\(1e\)--\(7\)
        & $5.4\pm1.0$ & $3.1\pm0.2$   &  $8.3\pm2.2$ & $2.7\pm0.2$
        & \diverge{$1.9e18$} & \diverge{$14.8\pm1.0$}\\
        \midrule

        \mRefiner{} & \(8\) &\(1e\)--\(3\)
        & $7.1\pm1.5$ & $3.5\pm0.4$   &  $4.4\pm1.8$ & $2.7\pm0.4$
        & $5.5\pm1.3$ & $6.9\pm1.0$\\

        \mRefiner{} & \(8\) &\(1e\)--\(4\)
        & $13.8\pm2.3$ & $5.0\pm0.5$   &  $8.6\pm4.2$ & $2.4\pm0.7$
        & $5.1\pm1.3$ & $5.9\pm1.1$\\

        \mRefiner{} & \(8\) &\(1e\)--\(5\)
        & $6.3\pm1.1$ & $3.5\pm0.4$   &  $6.0\pm1.8$ & $2.4\pm0.6$
        & $4.7\pm0.7$ & $5.4\pm1.2$\\

        \mRefiner{} & \(8\) &\(1e\)--\(6\)
        & $3.1\pm1.3$ & $2.2\pm0.2$   &  $6.4\pm2.1$ & $2.0\pm0.4$
        & \diverge{$0.1k\pm0.3k$} & \diverge{$6.1\pm4.3$}\\

        \mRefiner{} & \(8\) &\(1e\)--\(7\)
        & $4.3\pm1.4$ & $2.1\pm0.3$   &  $3.3\pm1.2$ & $1.6\pm0.3$
        & $88\pm70$ & $6.2\pm1.9$\\

        \bottomrule
    \end{tabular}
\end{table}

\paragraph{Accuracy}
\Cref{tab-app: accuracy refiner} evaluates the accuracy of these PDE-Refiner variants compared to \mACDM{} and \mUnet{} on our data sets \dTra{} and \dIso{}. Overall, the performance of \mRefiner{} across data sets, number of refinement steps \(R\), and noise variances \(\sigma\) is highly unpredictable. There is neither a clear accuracy trend over few or many refinement steps, nor high or low noise variance. Furthermore, a high accuracy on \dTra{} is not directly correlated with a high accuracy on \dIso{} either. As \mRefiner{} essentially improves upon one-step predictions of \mUnet{} via additional refinement steps, the results of a direct comparison are interesting: On \dTra{}, while \mRefiner{} consistently outperforms \mUnet{} in terms of LSiM, it just as consistently remains worse in terms of the MSE across hyperparameter combinations. We hypothesize that these results are linked to the fundamentally different spectral behavior of \mRefiner{} described by \textcite{lippe2023_PDERefiner}, but further research is required in this direction. On \dIso{}, \mRefiner{} either improves upon \mUnet{} or substantially diverges (marked in grey in \cref{tab-app: accuracy refiner}), especially for small \(\sigma\). Overall, PDE-Refiner is less effective than the stabilization techniques discussed in \cref{app: ablation training rollout,app: ablation training noise} in terms of accuracy improvements, and thus consistently falls short with respect to \mACDM{} across the test sets and hyperparameter combinations considered here.

\begin{figure}[ht]
    \centering
    \includegraphics[width=0.99\textwidth]{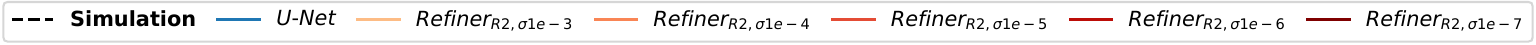}\\
    \includegraphics[width=0.48\textwidth]{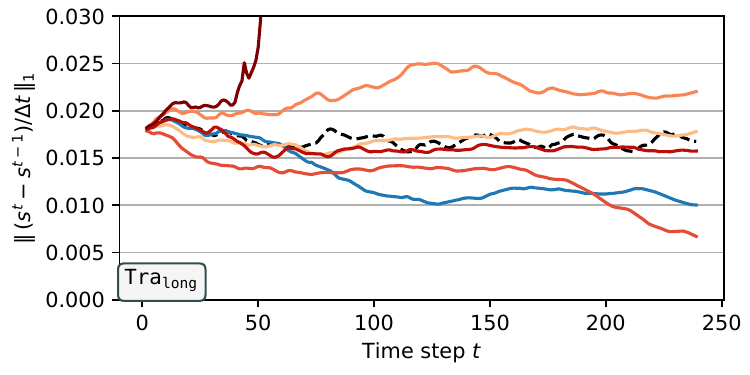}
    \hfill
    \includegraphics[width=0.48\textwidth]{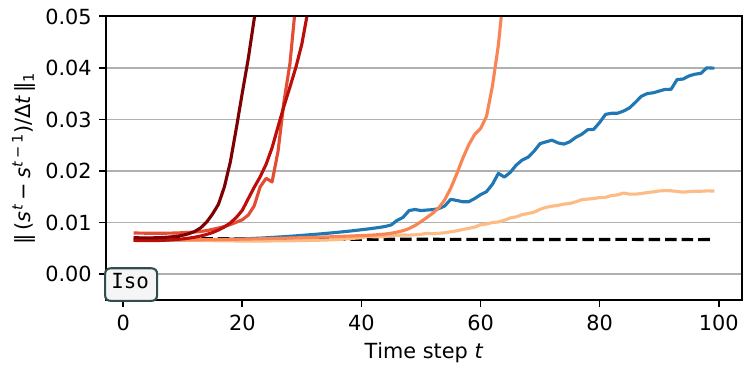}

    \vspace{0.1cm}

    \includegraphics[width=0.99\textwidth]{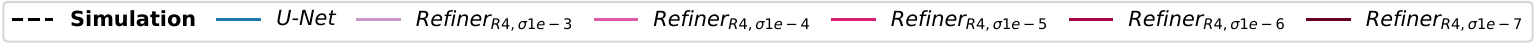}\\
    \includegraphics[width=0.48\textwidth]{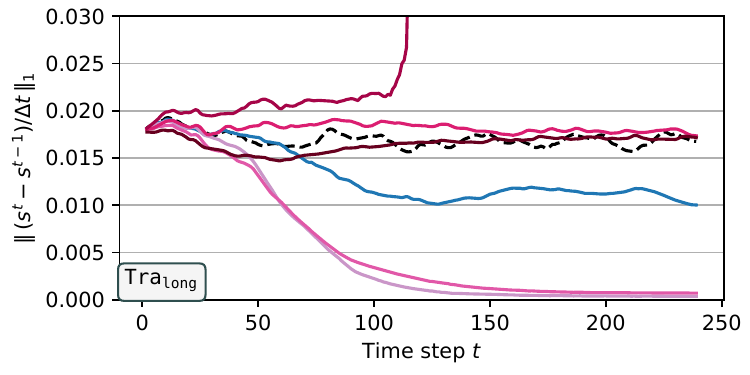}
    \hfill
    \includegraphics[width=0.48\textwidth]{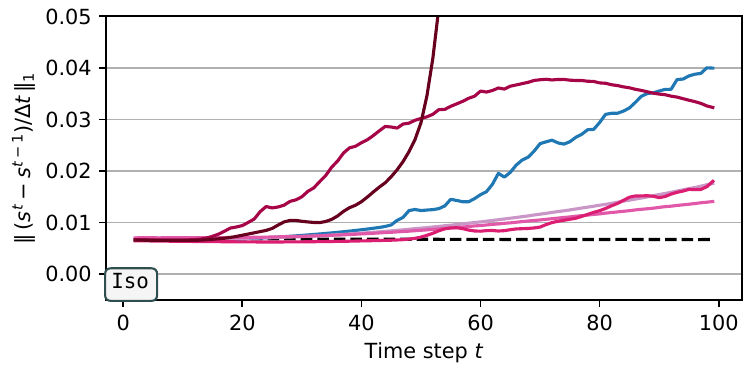}

    \vspace{0.1cm}

    \includegraphics[width=0.99\textwidth]{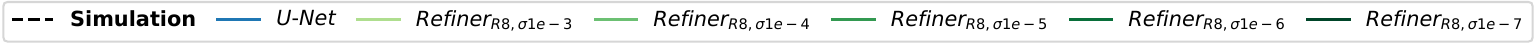}\\
    \includegraphics[width=0.48\textwidth]{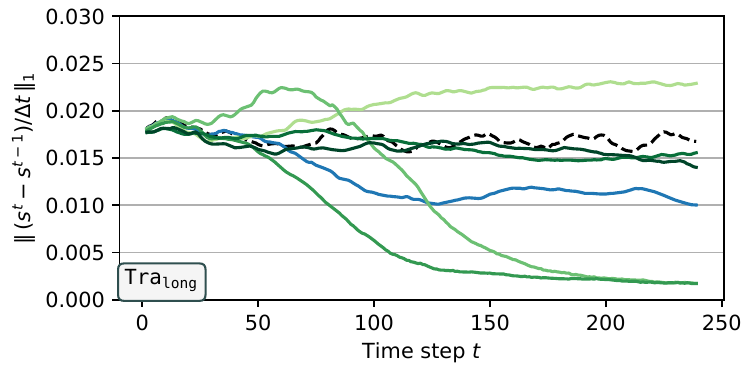}
    \hfill
    \includegraphics[width=0.48\textwidth]{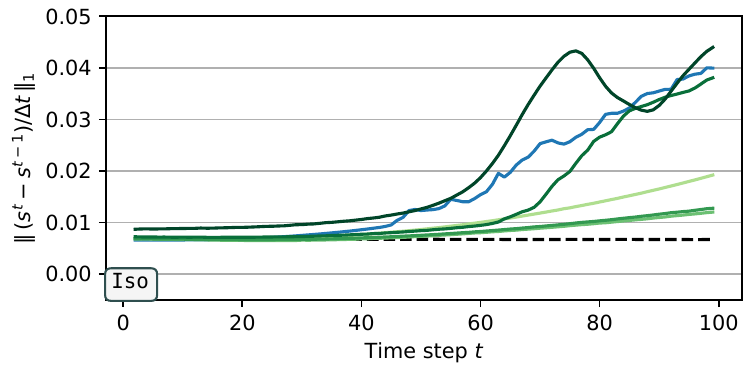}

    \caption{Temporal stability evaluation via error to previous time step on \dTraLong{} (left) and on \dIso{} (right) for PDE-Refiner with different hyperparameter combinations of refinement steps \(R\) and noise variances \(\sigma\). Standard deviations are omitted for visual clarity. The temporally most stable \mRefiner{} configuration is highly inconsistent, and for a given \(R\) depends on the data set and noise variance.}
    \label{fig-app: stability refiner}
\end{figure}

\paragraph{Temporal Stability}
To investigate the temporal stability of \mRefiner{}, we analyze the difference to the previous time step in \cref{fig-app: stability refiner}. First, it is shown that there are combinations of \(R\) and \(\sigma\) that substantially improve the rollout stability of \mRefiner{} compared to \mUnet{}, confirming the results from \textcite{lippe2023_PDERefiner}. However, as observed in terms of accuracy above, there is no consistent trend across hyperparameters and data sets. Especially, finding a suitable minimum noise variance \(\sigma\) depends on both, data set and number of refinement steps \(R\): While \(\sigma=10^{-6}\) works best on \dTraLong{} for \(R=2\), \(\sigma=10^{-7}\) is ideal for \(R=8\). On \dIso{}, \(R=2\) only works with \(\sigma=10^{-3}\), \(R=4\) requires \(\sigma=10^{-4}\), and \(R=8\) is most stable with \(\sigma=10^{-5}\). This unpredictable behavior with respect to important hyperparameters makes PDE-Refiner resource-intensive and difficult to employ in practice. The best \mRefiner{} variants on \dIso{}, while more stable compared to \mUnet{}, are nevertheless showing signs of instabilities around \(t=70\). This means the refinement increases stability, but still falls short compared to the other stabilization techniques discussed in \cref{app: ablation training rollout,app: ablation training noise}.

\begin{figure}[ht]
    \centering
    \includegraphics[width=0.49\textwidth]{plots_tra/data_posterior_longer_pres_2_acdm}
    \hfill
    \includegraphics[width=0.49\textwidth]{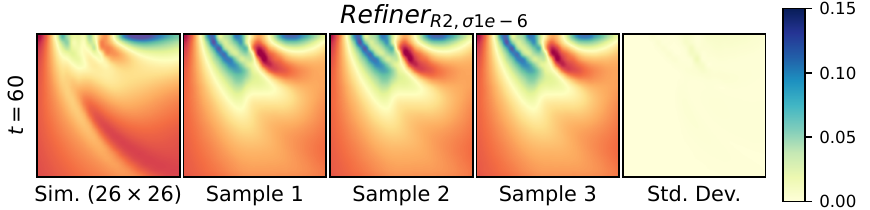}\\

    \vspace{0.1cm}
    \includegraphics[width=0.49\textwidth]{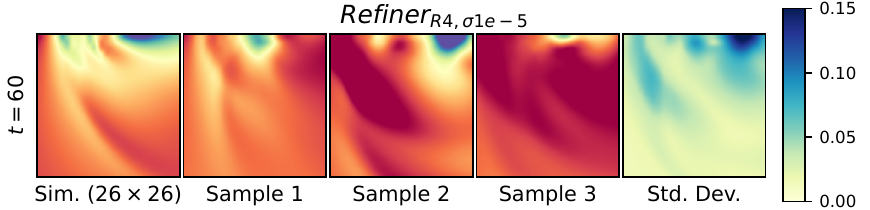}
    \hfill
    \includegraphics[width=0.49\textwidth]{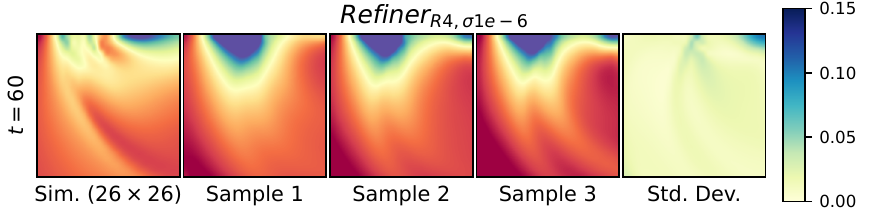}\\

    \vspace{0.1cm}
    \includegraphics[width=0.49\textwidth]{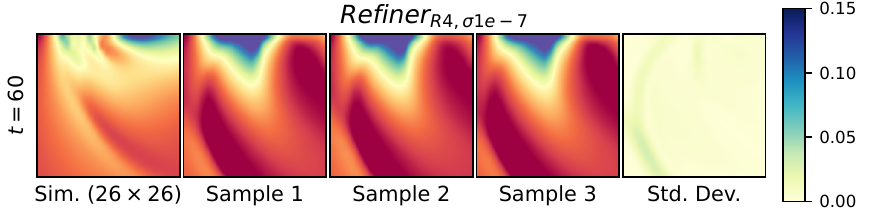}
    \hfill
    \includegraphics[width=0.49\textwidth]{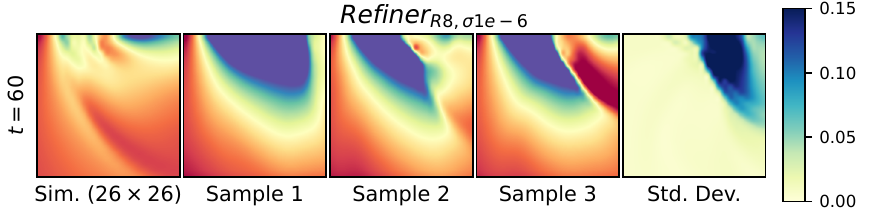}

    \caption{Posterior samples on \dTraLong{} from \mACDM{} (top left) compared to various PDE-Refiner ablation models with different refinement steps \(R\) and noise variances \(\sigma\) (other). \mRefiner{} lacks sample diversity and quality compared to \mACDM{} across values for \(R\) and \(\sigma\).}
    \label{fig-app: posterior refiner}
\end{figure}

\paragraph{Posterior Sampling}
As PDE-Refiner relies on deterministic predictions combined with probabilistic refinements, achieving a broad and diverse posterior distribution is difficult. In \cref{fig-app: posterior refiner}, we visualize posterior samples for \dTraLong{} from \mACDM{} and \mRefiner{} with \(R \in \{2,4,8\}\) and \(\sigma \in \{10^{-5}, 10^{-6}, 10^{-7}\}\). While \mACDM{} creates a broad range of samples as discussed above, \mRefiner{R2,$\sigma$1e-6}, \mRefiner{R4,$\sigma$1e-6}, and \mRefiner{R4,$\sigma$1e-7} do not create any noticeable variance. While additional refinement steps slightly improve the spread across samples, even \mRefiner{R8,$\sigma$1e-6} can only create minor differences with very similar spatial structures. Note that the \mRefiner{} models are in general unable to create the detailed shockwaves below the cylinder that are found in the simulation and the \mACDM{} samples. In addition, unphysical predictions after longer rollouts can be observed across refinement steps and noise variances in the visualizations in \cref{fig-app: example tra refiner}. Using very larger values of \(\sigma\) should theoretically allow \mRefiner{} to focus on a larger range of frequencies. However, this increased the stability issues further and did not substantially improve the quality or diversity of posterior samples over \cref{fig-app: posterior refiner}.

\paragraph{Summary}
While the stability benefits of a well-tuned setup with PDE-Refiner compared to a simple one-step prediction with \mUnet{} are highly desirable and can be achieved with less inference overhead compared to \mACDM{}, the method has several disadvantages: We found the setup to be very sensitive regarding changes to refinement steps, data set, or noise variance. This means, a large amount of computational resources are required for parameter tuning, which is crucial to obtain good results. Suboptimal combinations of refinement steps and noise variance show substantially degraded performance compared to \mUnet{} in our experiments, and even tuned setup did exhibit instabilities across training runs and model samples. Furthermore, \mRefiner{} is less accuracy, and has limits in terms of the posterior sampling compared to a more direct application of diffusion models in \mACDM{}.

\section{Prediction Examples} \label{app: prediction examples}

\begin{figure}[p]
    \centering
    \includegraphics[width=0.99\textwidth]{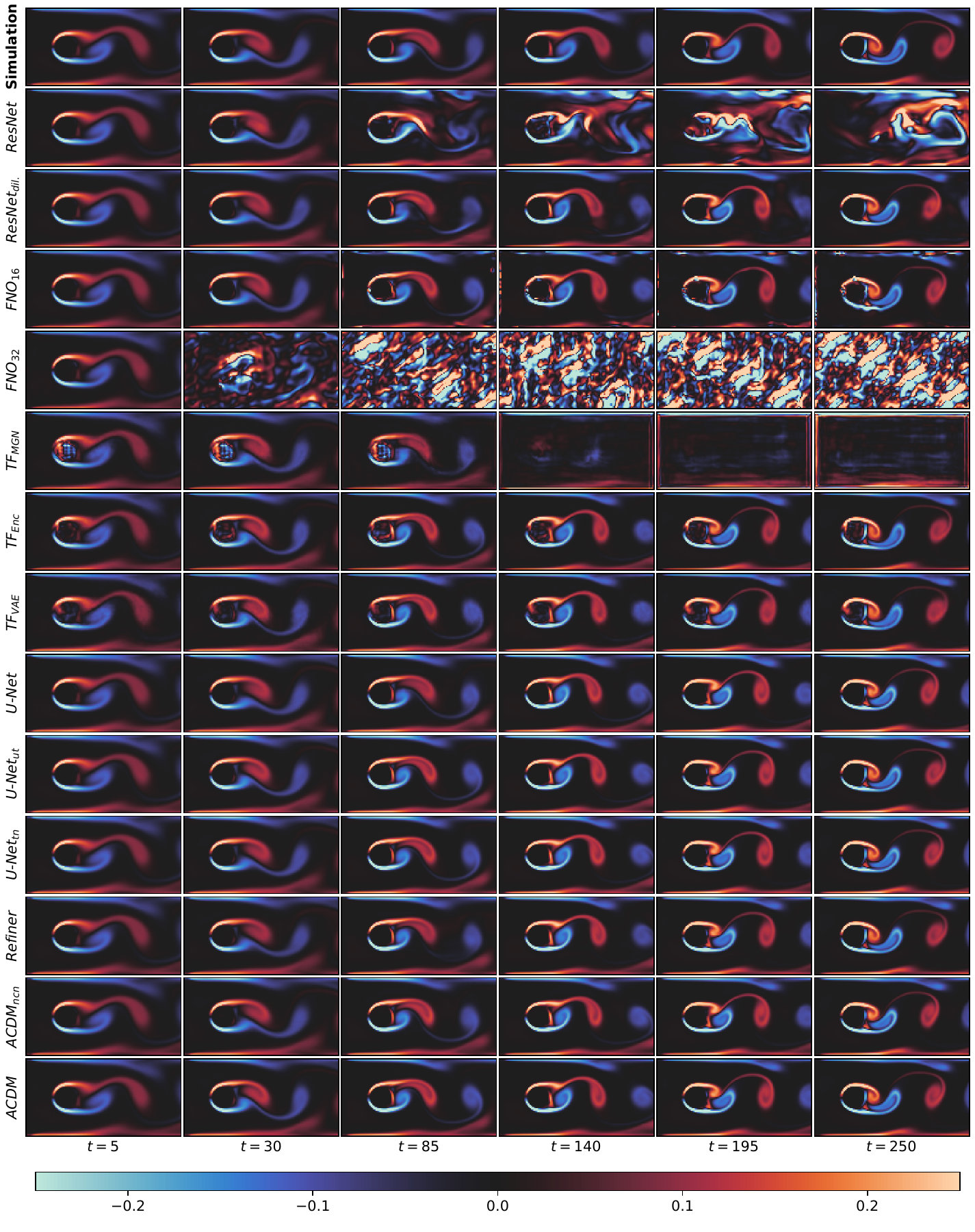}
    
    \caption{Vorticity predictions for the \dIncVar{} sequence.}
    \label{fig-app: example inc}
\end{figure}

\begin{figure}[p]
    \centering
    \includegraphics[width=0.99\textwidth]{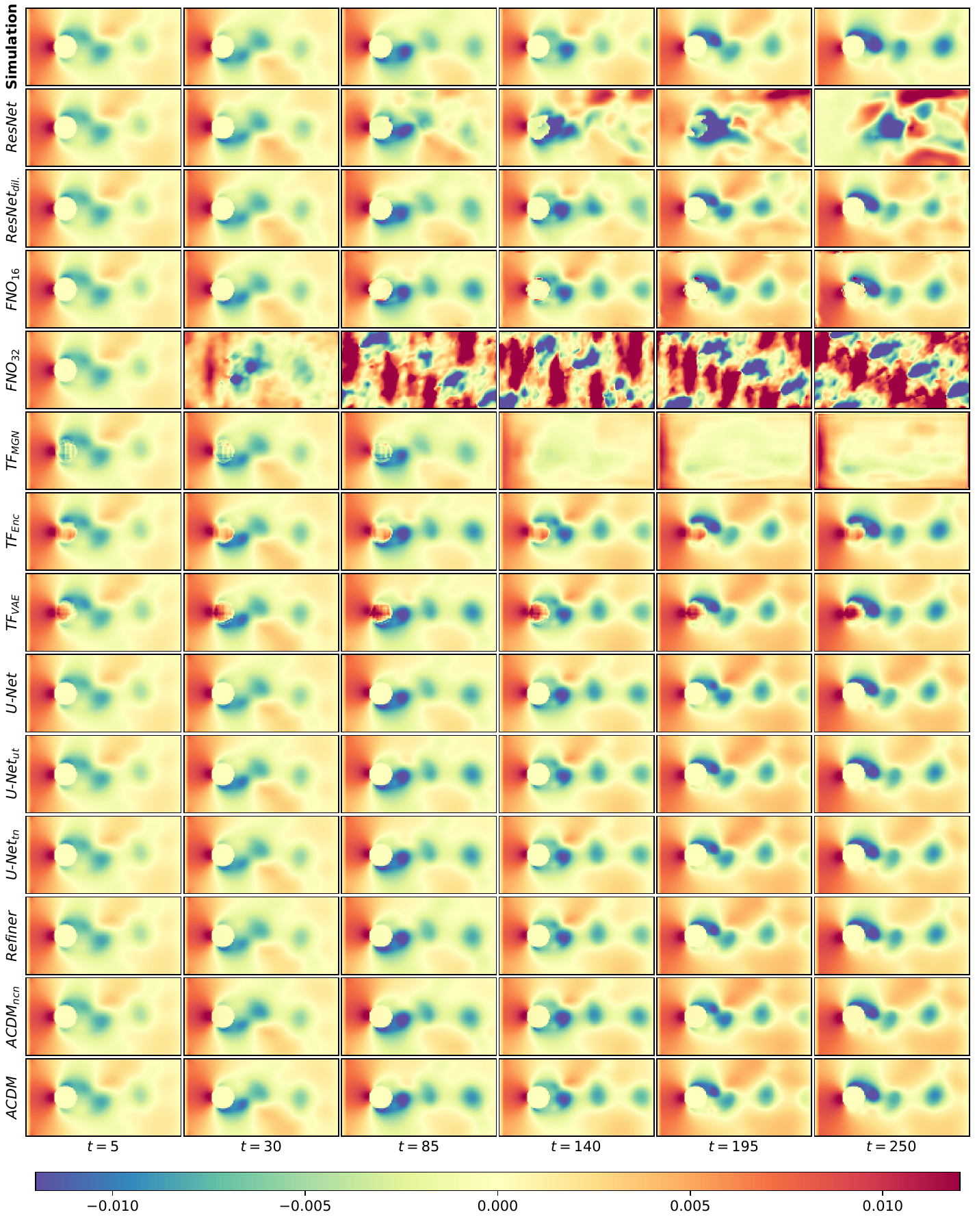}

    \caption{Pressure predictions for the \dIncVar{} sequence.}
    \label{fig-app: example inc 2}
\end{figure}

\begin{figure}[p]
    \centering
    \includegraphics[width=0.99\textwidth]{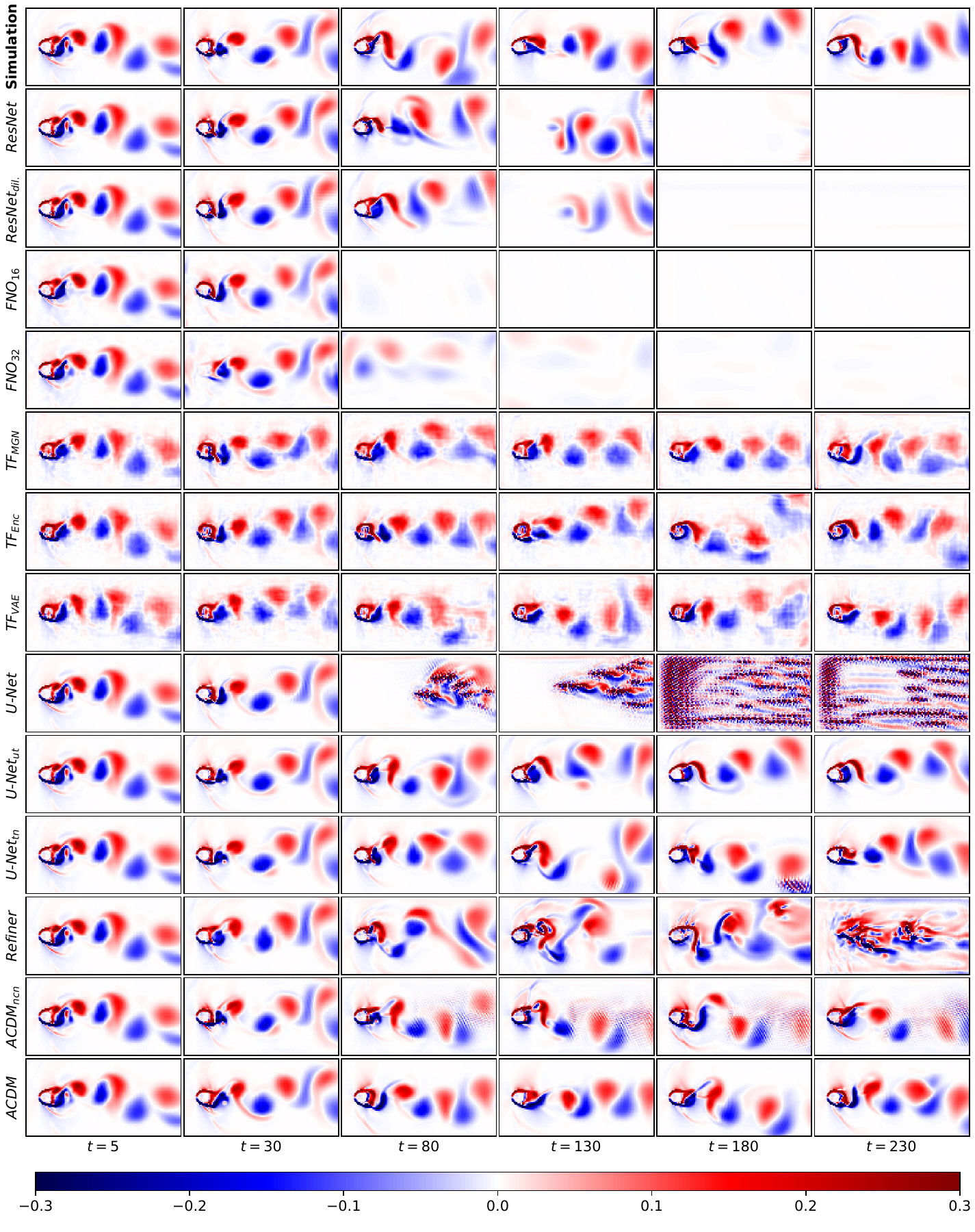}

    \caption{Vorticity predictions for an example sequence from \dTraLong{} with \( \mathit{Ma} = 0.64 \).}
    \label{fig-app: example tra}
\end{figure}

\begin{figure}[p]
    \centering
    \includegraphics[width=0.99\textwidth]{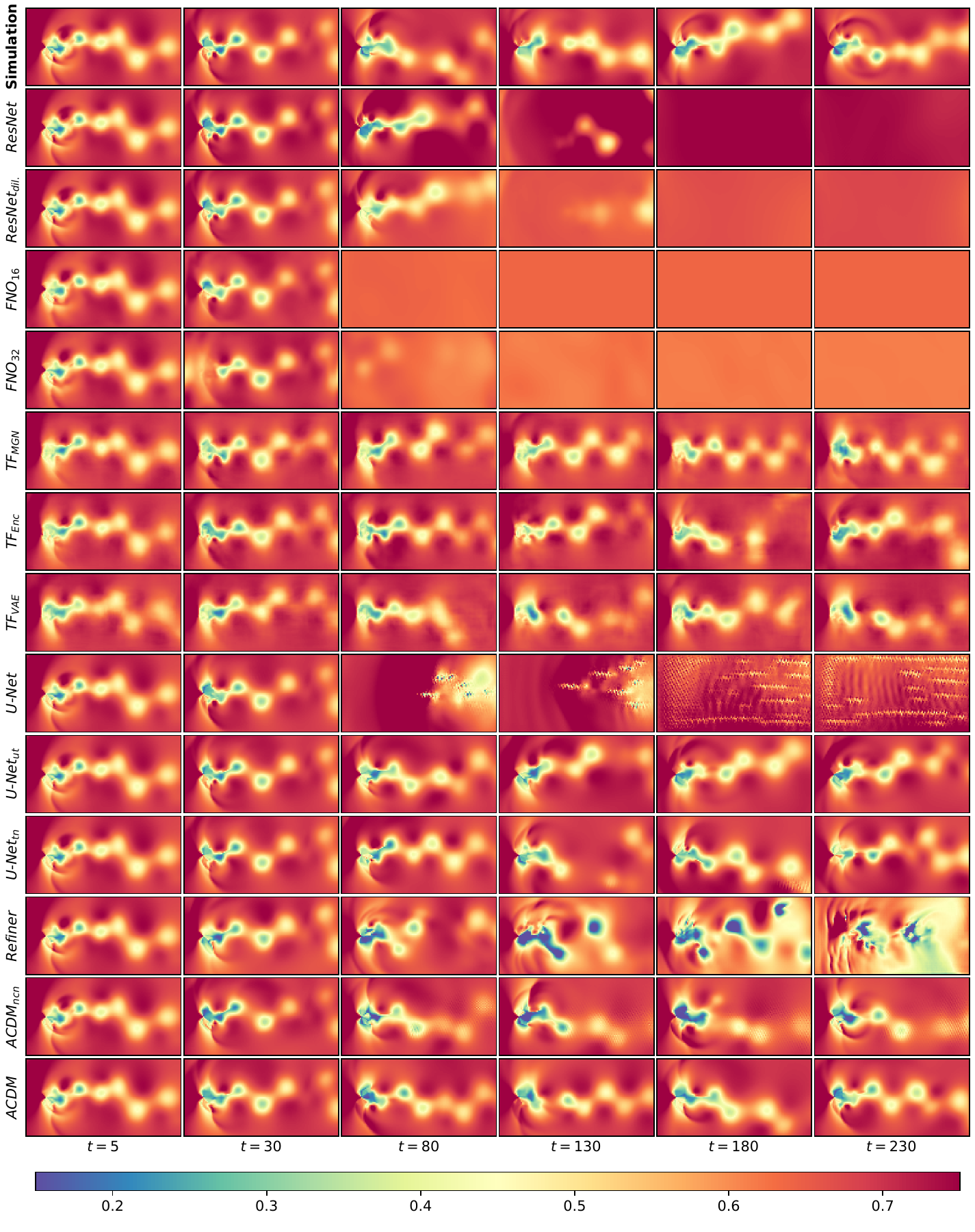}

    \caption{Pressure predictions for an example sequence from \dTraLong{} with \( \mathit{Ma} = 0.64 \).}
    \label{fig-app: example tra 2}
\end{figure}

\begin{figure}[p]
    \centering
    \includegraphics[width=0.99\textwidth]{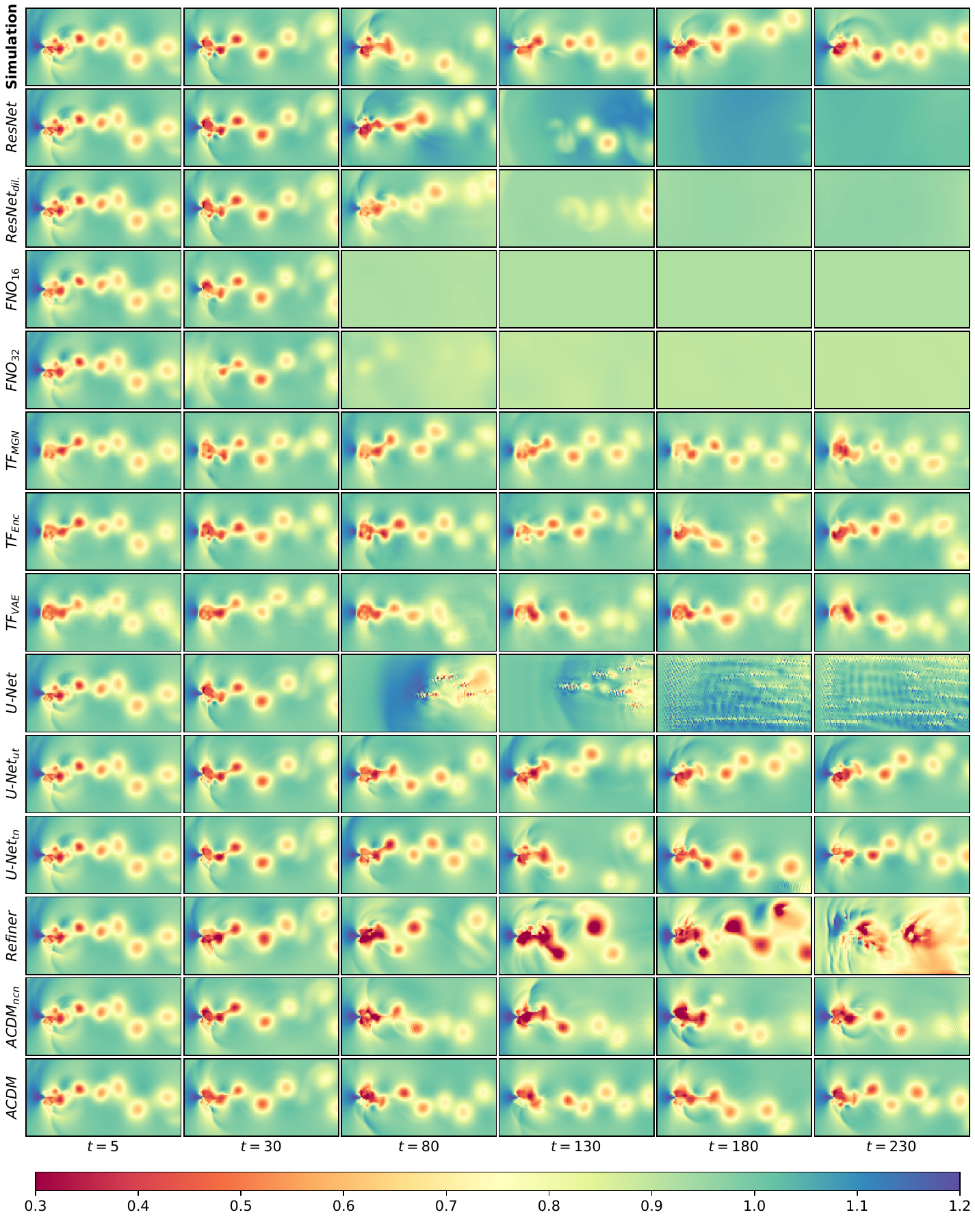}

    \caption{Density predictions for an example sequence from \dTraLong{} with \( \mathit{Ma} = 0.64 \).}
    \label{fig-app: example tra 3}
\end{figure}

\begin{figure}[p]
    \centering
    \includegraphics[width=0.99\textwidth]{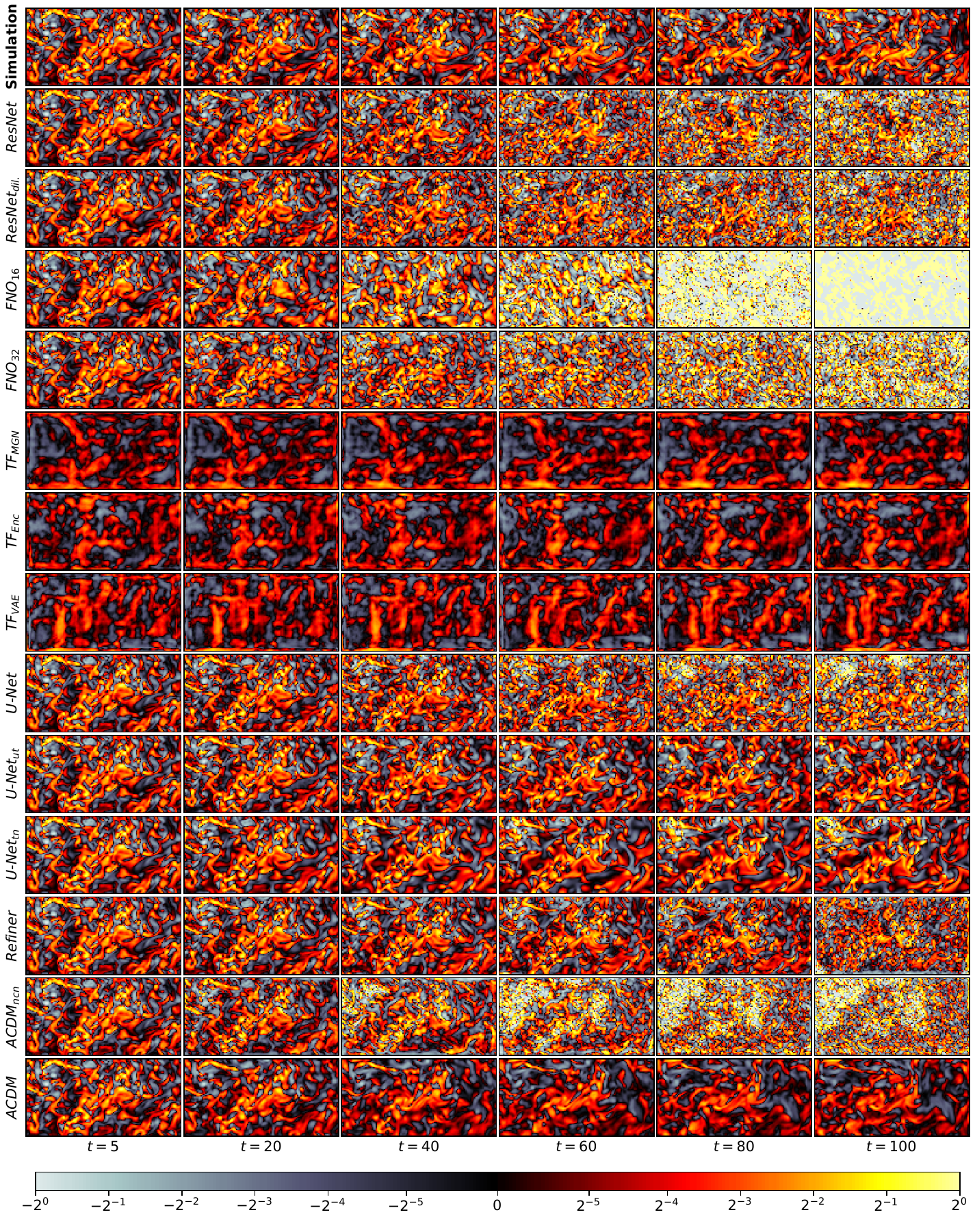}

    \caption{Vorticity predictions (only z-component) for an example sequence from \dIso{} with \( z = 280 \).}
    \label{fig-app: example iso}
\end{figure}

\begin{figure}[p]
    \centering
    \includegraphics[width=0.99\textwidth]{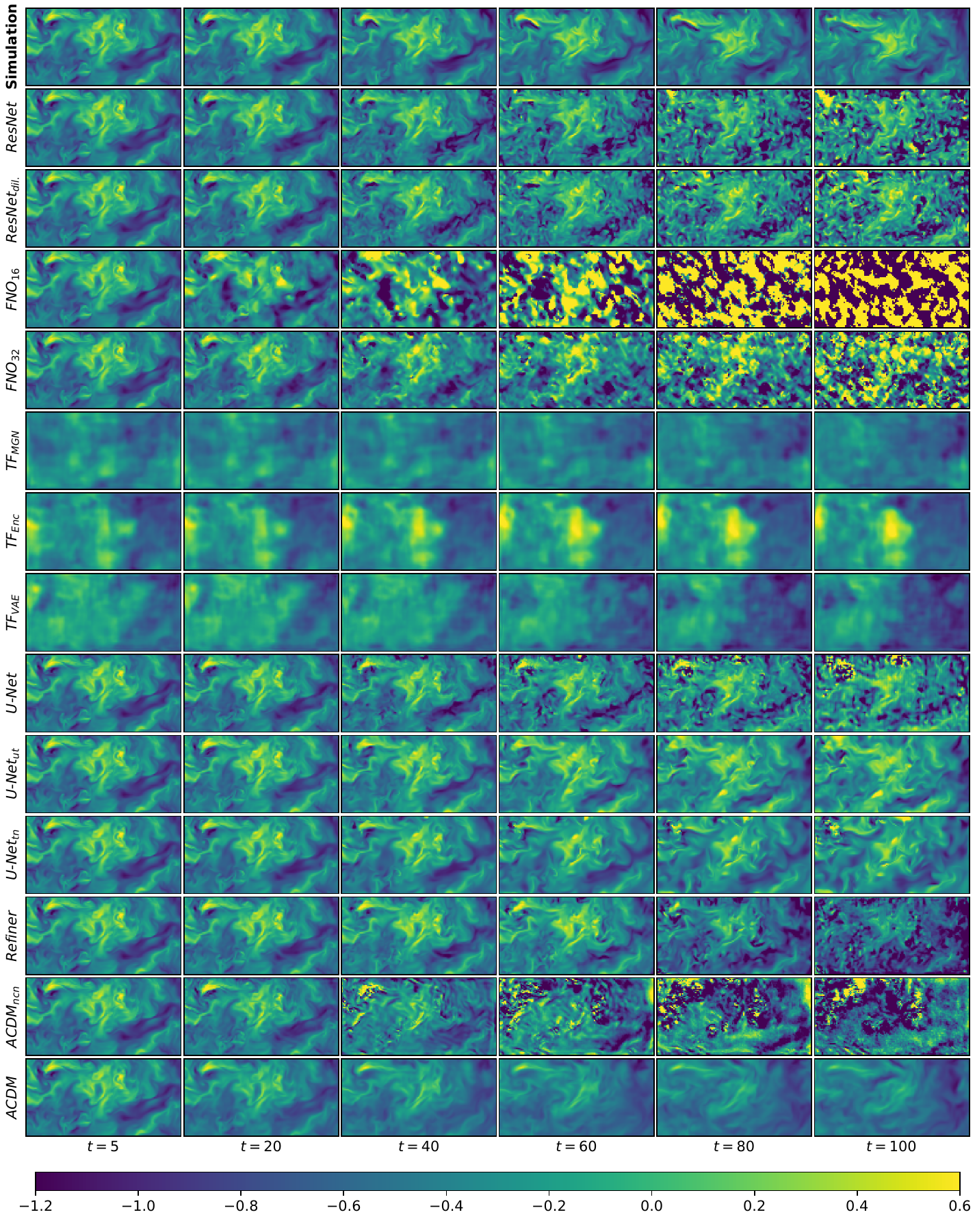}

    \caption{Z-velocity predictions for an example sequence from \dIso{} with \( z = 280 \).}
    \label{fig-app: example iso 2}
\end{figure}

\begin{figure}[p]
    \centering
    \includegraphics[width=0.99\textwidth]{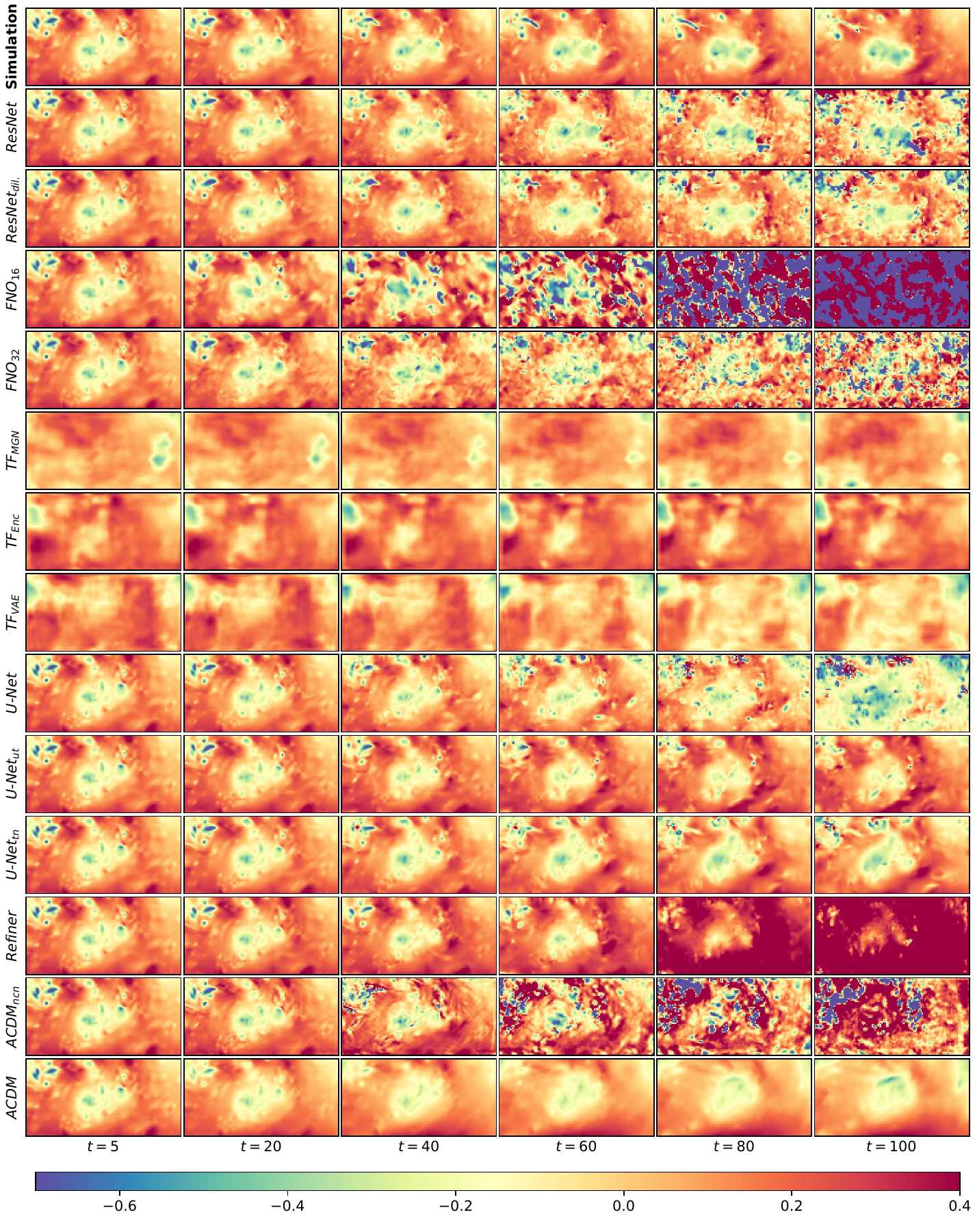}

    \caption{Pressure predictions for an example sequence from \dIso{} with \( z = 280 \).}
    \label{fig-app: example iso 3}
\end{figure}

Over the following pages, prediction examples from all analyzed methods are displayed. Shown are the different fields contained in an exemplary test sequence from each experiment. \Cref{fig-app: example inc,fig-app: example inc 2} feature the \dIncVar{} case, \cref{fig-app: example tra,fig-app: example tra 2,fig-app: example tra 3} contain an example from \dTraLong{} with \( \mathit{Ma} = 0.64 \), and \cref{fig-app: example iso,fig-app: example iso 2,fig-app: example iso 3} display a sequence from \dIso{} with \( z = 280 \). Videos of model predictions for some example sequences from each data set are provided, as they can visualize several aspects like temporal stability, temporal coherence, and visual quality better than still images. We also include videos of posterior samples from the probabilistic architectures, and a temporal coherence analysis of \mACDM{}. All videos can be found alongside this work at \url{https://ge.in.tum.de/publications/2023-acdm-kohl/}.

\section{Ablation Study Prediction Examples} \label{app: prediction examples ablation}

Below the prediction examples for the model architectures, we show prediction examples from different ablation study models provided in \cref{app: ablation diffusion steps,app: ablation training rollout,app: ablation training noise,app: ablation refiner}. Shown are the pressure field from \dTraLong{} with \( \mathit{Ma} = 0.64 \), as well as a vorticity sequence from \dIso{} with \( z = 280 \).

\clearpage

\end{document}